\documentclass[12pt]{article}
\usepackage[bottom]{footmisc}
\usepackage{algorithm} 
\usepackage{enumitem}
\usepackage{algpseudocode}
\AtBeginDocument{\renewcommand{\State}{}}
\usepackage[english]{babel}
\usepackage{blindtext}
\usepackage[utf8]{inputenc}
\usepackage{pgfplots}
\usepackage{amsmath}
\usepackage{amssymb} 
\usepackage{listings}
\usepackage{xcolor}
\usepackage{fancyhdr}
\usepackage{stackengine}
\usepackage{graphicx}
\usepackage{tabularx,booktabs,ragged2e}
\usepackage{subfig}
\usepackage{tikz}
\usepackage{pgf}
\usepackage{inputenc}
\usepackage{hyperref}
\usepackage[T1]{fontenc}
\usepackage[scaled=0.85]{beramono}
\usepackage{microtype}

\definecolor{unige}{HTML}{d80669}
\definecolor{code_background}{HTML}{F0F0F0}
\definecolor{code_strings}{HTML}{845c6d}
\definecolor{code_output}{HTML}{7897AB}

\hypersetup{
    colorlinks=true,
    linkcolor=unige,
    urlcolor=black,
}

\lstdefinestyle{mystyle}{
    backgroundcolor=\color{code_background},   
    commentstyle=\color{gray},
    keywordstyle=\color{unige},
    numberstyle=\tiny\color{gray},
    stringstyle=\color{gray},
    basicstyle=\ttfamily\footnotesize,
    breakatwhitespace=false,         
    breaklines=true,                 
    captionpos=b,                    
    keepspaces=true,                 
    numbers=left,                    
    numbersep=5pt,                  
    showspaces=false,                
    showstringspaces=false,
    showtabs=false,                  
    tabsize=2
}

\begin{document}

\newcommand{\Exjobbsnummer}[1]{
\begin{tikzpicture}[overlay, remember picture]
\path (current page.north east) ++(-1,-1) node[below left] {{\small #1}};
\end{tikzpicture}
}

\newcommand{\Examensjobbspoang}[1]{
\begin{tikzpicture}[overlay, remember picture]
\path (current page.north east) ++(-1,-1.5) node[below left] {{\normalsize \scshape Examensarbete #1 HP}};
\end{tikzpicture}
}

\newcommand{\datum}[1]{
\begin{tikzpicture}[overlay, remember picture]
\path (current page.north east) ++(-1,-2.0) node[below left] {{\normalsize #1}};
\end{tikzpicture}}

\newcommand{\storlitentitel}[2]{
\center
\rule[0.2cm]{13cm}{0.1cm}
{ \huge \bfseries #1}\\[0.4cm] 
{\Large \slshape #2}\\[0.4cm]
\rule[0.2cm]{13cm}{0.1cm}\\[3cm]

}

\newcommand{\Namn}[2]{
\begin{minipage}{0.4\textwidth}
\normalsize
\centering
#1 \textsc{#2}\\
\end{minipage}\\
}

\newcommand{\LoggaSwe}{
\includegraphics[scale=.3]{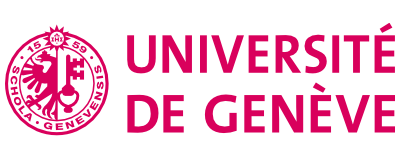}\\[1.5cm]
\textsc{\large Master of Computer Science, Faculty of Sciences}\\[0.7cm]
}

\begin{titlepage}

\center

\datum{June 2023}

\LoggaSwe

\storlitentitel{Master Thesis}{Custom Named Entity Recognition and Topic Classification for Global Health Publications}

\vspace{1cm}
\Large Genis Skura
\\
\vspace{3cm}
\large Supervisor: Dr. Jean-Luc Falcone\\
\large Co-Supervisor: Prof. Antoine Geissbühler
\vfill
\end{titlepage}
\pagestyle{fancy}
\fancyhf{}
\rhead{\small University of Geneva}
\lhead{\small Genis Skura}
\rfoot{\thepage}

\section*{Acknowledgments}
\State As this two-year spanning chapter of my academic voyage draws to a close, it has been a testament to the power of support from the University of Geneva since my acceptance as an international student. These two years have unfolded as a mosaic of experiences, challenges, academical and personal triumphs.

\State I extend my sincere appreciation to the mentors of this project, Dr. Jean-Luc Falcone, whose guidance has illuminated this academic assignment and whose trust allowed me to pursue research on a topic of my proposal, and Prof. Antoine Geissbühler, whose expertise facilitated my adaptation to a new domain such as global digital health. Your mentorship has not only made this possible but has also enriched my personal and professional growth.

\State To friends, colleagues, peers, and well-wishers, your contributions have made this a memorable chapter. Lastly, To my cherished parents, Alisa and Mati. Your unending encouragement and sacrifices have paved the path for this journey and your belief in my potential has been the driving force behind every step I've taken and will take.

\newpage
{
  \hypersetup{linkcolor=black}
  \tableofcontents
}

\newpage
\null
\thispagestyle{empty}

\newpage

\section{Introduction}
\State This document presents a comprehensive account of my practical and theoretical work over the past year, serving as a written report. A work dedicated to the field of natural language processing (\textit{NLP}) for the global health domain, one domain of a multi-disciplinary nature that is found at the intersection of various fields such as: healthcare, global politics and decision-making, social sciences, economics, etc. \\ 
After having conceptualized the nature of this domain, one can imagine the large quantity of literature frequently published by different professionals and numerous organizations located all around the world with the goal of tackling each global health challenge. One of the crucial objectives of this published literature is to enhance inter-connectivity and foster mutual understanding among all entities that work on the domain.

\vspace{0.2cm}

\State As a principle, mutual understanding has one core concept in it's meaning, human language. Broadly, natural language processing is defined as a set of methods that makes human language accessible to computers. It is situated in the neighbourhood of computational linguistics, machine learning, and artificial intelligence. This neighbourhood offers a variety of tasks for an emerging interdisciplinary field such as the chosen one, supported by tools laying in the thin line between text mining and natural language processing [\hyperlink{eisenstein_intro}{1}].

\State From a large variety of tasks and applications of NLP, we will be focused on ones that analyze linguistic structures based on form (morphologically), meaning (semantically), sentence structure (syntactically), and word usage (lexically). This project can be categorized as a contribution to NLP applications, particularly in the fields of text mining, information retrieval, and document categorization, all of which are essential for healthcare and knowledge systems worldwide [\hyperlink{nlp_applications}{2}].

\vspace{0.2cm}

\State Specifically, this thesis delves into the realm of information retrieval, employing \textit{word2vec} for tag similarity discovery, studying the impact of corpus size on the process. Document categorization involves a taxonomic classification of tags, assigning multiple tags to sentences using a multi-label AI model. Last but not least, Named Entity Recognition takes center stage, extracting essential entities from publications.

\hypertarget{motivy}{\subsection{Motivation \& Industry Need}}
\State In this next chapter, I would like to identify my motivation behind the choosing of this topic, a motivation fueled by two main reasons, a personal one and an academical one. Reasons that drove me to the proposal of this topic to my supervisors.

\vspace{0.3cm}

\State Personally, since the first day I became part of this domain through an internship at \href{https://gdhub.org/}{\textcolor{unige}{\textit{\underline{Geneva Digital Health Hub}}}}, I could observe a common theme in it's knowledge base and the ones of it's global partners such as WHO \footnote{\ World Health Organization}, incomplete annotations going through a process of human annotation. \\
This remark of data quality is especially delicate in cases where we are dealing with a graph knowledge base, such as was my case with \textit{neo4j}, where accurate annotations are vital to ensuring efficient retrieval, data representation, and contextual analysis. This isn't an isolated phenomena of my experience as a developer, but rather a general industry need that tends to be more delicate in the global health domains and healthcare systems. \\
Acknowledging the expertise and authority of domain experts in human annotation, it also becomes imperative to free their time from data preparation, allowing them to concentrate more on providing care and facing global challenges [\hyperlink{liberate}{3}].  

\vspace{0.3cm}

\State Having expressed that, I would like to mention the second main part behind my motivation. A part born of a pure academic interest to dive deeper into the field of NLP after having taken both courses of our masters related to it, courses given by Dr. Tanja Samardžić. Explicitly, formulating and analyzing a research hypothesis, comparing state of the art to my custom methods, and exploring the literature for each one of the above mentioned topics.

\vspace{0.2cm}

\State To recap, the combination between a systematic theoretical review of the project's NLP topics and the development of a potential finished annotation pipeline able to be deployed in a real-world industrial environment, served as catalysts behind my choice. \\
Let us move on to the next final introductory chapter where a general organization of the thesis will be provided.

\newpage

\subsection{Organization of the Thesis}
\State Being aware that the thesis is made up of different disjointed sub-tasks of NLP in theory, but with the same goal in mind, I decided to provide a simple schema of it's organization. A schema that will briefly present the main tasks that made up the core of this project. 

\begin{center}
\includegraphics[scale = 0.55]{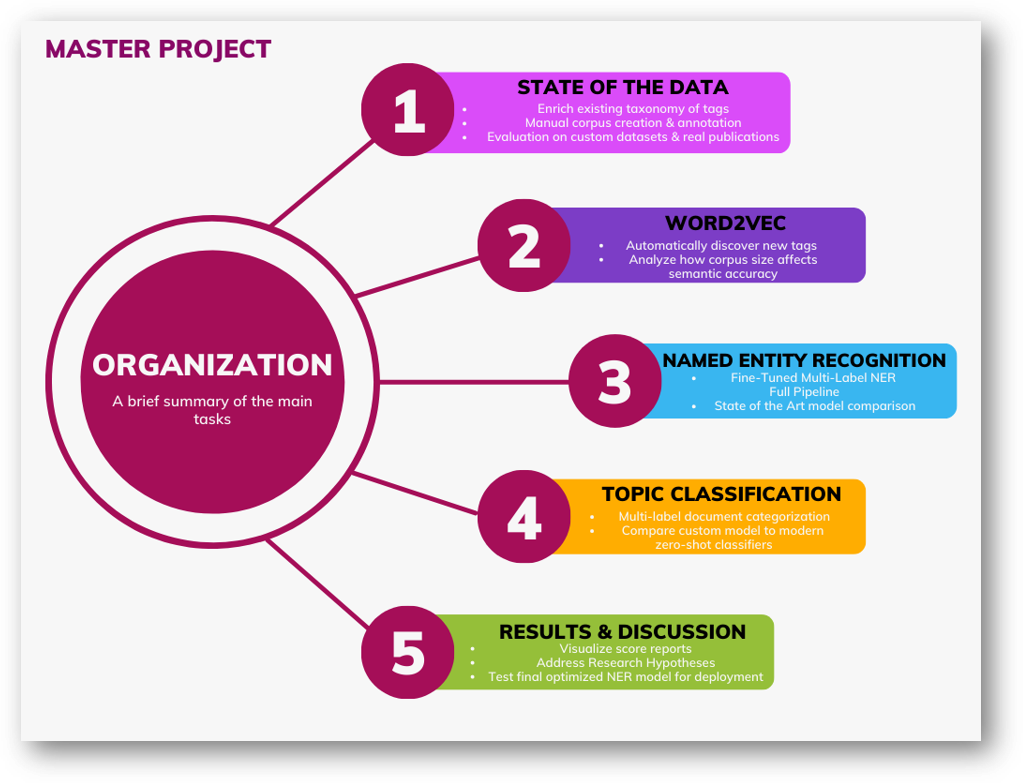}
\textit{Circular Diagram of the Workflow} \footnote{\ Created through a template in \href{https://www.canva.com/}{\textcolor{unige}{\textit{\underline{canva}}}}}
\end{center}

\State This visual representation should be viewed as an accurate order of how the project progressed from one stage to the other. Although each chapter was born with it's own formulized research objective and their own results to be discussed, they all tended to the same nature of data, that which is found in global digital health publications.

\newpage

\section{word2vec Tag Discovery}
\State We will begin with the task of utilizing word embeddings as a system of tag discovery. This task should be viewed as the prologue of the project's workflow, one that studies one of the integral parts of NLP, embedding words to a low-dimensional vector representation that can be mathematically manipulated to our task's desires.

\subsection{Word Embeddings}
\State To present it simply, word embeddings map multiple thousands of words in a mathematical space that positions them corresponding to their semantic properties. Subsequently, the mathematical similarity of these positioned continuous vectors is conjectured to the real semantic similarity that words share with each other in human language [\hyperlink{word_embeddings}{4}]. 


\vspace{0.2cm}

\State For our case, a sub-technique of word embeddings was chosen, neural embeddings. Given any given set of words $\mathcal{W}$ in a provided corpus, neural embeddings achieve a dense low-dimension vector representation in a continuous space through a neural model [\hyperlink{pred_embeddings}{5}].

\begin{center}
    $\mathcal{W} \xrightarrow[]{Neural Network} \mathbb{R}^d$
\end{center}
To map a set of words $\mathcal{W}$ to their respective vector space $\in \mathbb{R}^d$ through a pre-defined neural network. To briefly understand how this is achievable, it is important to delve into the mathematical formalization of the process, which enabled the training of a neural network to discover its optimized parameters. \\
Let the vector $u_w$ be the embedding of word $w$ and $v_c$ the embedding for a given context $c$ \footnote{\  Context $c = \{w_{i-N}, \ldots, w_{i-1}, w, w_{i+1}, \ldots, w_{i+N}\}$ represents the words surrounding a given word $w$. $N$ is a hyperparameter.}. The inner product $u_w \cdot v_c$ represents the compatibility between the word and the context. By representing this inner product into an approximation of the log-likelihood of a corpus, it creates the possibility two estimate both parameters by backpropagation [\hyperlink{dae_embeddings}{6}].

\begin{center}
    $log$ - $likelihood = \log (\sigma(u_w \cdot v_c))$ \footnote{\ $\sigma \Rightarrow$ Sigmoid function which maps the final result to a probability value between $0$ and $1$.} \ (1)
\end{center}

\State Therefore, through this estimation of the likelihood of the context word when given the chosen word, the model measures how well they align in the vector space. The parameters are adjusted as traditional in a neural network, with an optimization process that finds the ideal parameters which minimize a given loss function.

\State The power of neural networks to learn complex patterns and capture non-linear relationships in the data reflected in their accuracy to also capture the complex linguistic nuances in our languages. Moreover, neural methods turned out more adaptable than traditional ones when considering that these nuances and relationship between linguistic elements differs for every language.\\ 
In addition to that, the raw superiority in computational power and efficiency to process large amounts of text corpora made them even more of a suitable choice. As a consequence, neural network based representations have almost fully replaced the traditional conventional count-based methods [\hyperlink{pred_embeddings}{5}].

\State After having pointed out the reasons behind choosing neural embeddings as a technique, I'd like to specify the chosen framework in itself, \textit{word2vec}. Since it's initial presentation in the influential paper of Mikolov et al. in 2013 [\hyperlink{word2vec}{7}], \textit{word2vec} popularized word embeddings and consequently, language modeling tasks [\hyperlink{pred_embeddings}{5}]. It is also classified as a predictive model given that it was trained with a loss function based on predictive language modeling objectives, such as:

\begin{center}
    $P(w_c | w) = \frac{exp(u_w \cdot v_c)}{\sum^N_{j = 1} exp(u_w \cdot v_j)}$ \ \hypertarget{lang_model}{(2)}
\end{center}
Predicting a context word $w_c$ when given a target word $w$. To ultimately define a general loss function calculated for each word in the original set $\mathcal{W}$:\

\begin{center}
    $l = - \sum^{\mathcal{W}}_1 \sum_{w_c \in c} log(P(w_c | w))$ \ (3)
\end{center}
We can notice that the loss function $l$ corresponds to the cross-entropy for all context words across all target words in the corpora [\hyperlink{word2vec_jura}{8}]. Therefore, theoretically, a minimization of this loss will lead to a better language model.

\State In the original presentation, it's accuracy as a vector representation was measured in a word similarity task based on syntactic and semantic similarity [\hyperlink{word2vec}{7}]. For example, in the context of countries and capitals:

\begin{center}
    "\textit{Paris}" - "\textit{France}" + "\textit{Italy}" $\approx$ "\textit{Rome}" \footnote{\ True equation in the vector space}
\end{center}

\newpage

\State This exact discovery of real semantic relationships between words in addition to the speed of training on considerable corpus lengths led me to choose \textit{word2vec} as the framework for my task. Concretely, my approach that aims to automatically discover novel global health terms by querying existing labels for similarity within the vector space of a custom corpus.

\State This last statement led to the birth of the research question related to this framework: \textit{
"What is the influence of corpus size and limited specialized corpora on the performance of Word2Vec in capturing word similarity within the domain of global health?"} 

\State However, before specifying the model and its hyperparameters, we need to define our existing set of labels and the text corpora that will be compared with the question in mind.

\subsection{Data Preparation}
\State Before building the word2vec model in itself, a data preparation process was made. This should be a given fact knowing that the model needs a predefined vocabulary in which it will be trained and learn to map it to the corresponding continuous vector space. The vocabulary is build from the given raw text corpus while our custom tag discovery process is achieved by querying each element of another predefined set of tags. \\
Therefore, in the next two sub-chapters, we will proceed to specify these two segments required before training the model.

\subsubsection{Initial Tags}
\State The preliminary set of tags was extracted from the knowledge base of \\ \textit{Implementome}, a global digital health knowledge management platform released by \textit{gdhub} \footnote{\ Geneva Digital Health Hub}, a platform in which I was part of the developing team.

\State Formally, these tags were named concept tags and added as an optional property to the global digital health publications stored in the database. Therefore, each publication was manually being annotated with one or more of these tags by a given global health expert part of the team. This detail motivated me to incorporate the automatic multi-label topic classification study into this project, with the goal of increasing data quality.

\State Another important factor that went into my consideration was the fact that the knowledge base was structured in graphs, through \href{https://neo4j.com/}{\textcolor{unige}{\textit{\underline{neo4j}}}} as a graph database management system. A correct high quality data representation is vital in graph-based management as it heavily influences the effectiveness of processing and information retrieval and leads to better long-term scalability [\hyperlink{neo4j_imp}{9}]. 

\vspace{0.1cm}

\State Having said that, I would like to present them with a visualization of their initial form in the database.

\begin{center}
    \includegraphics[scale = 0.7]{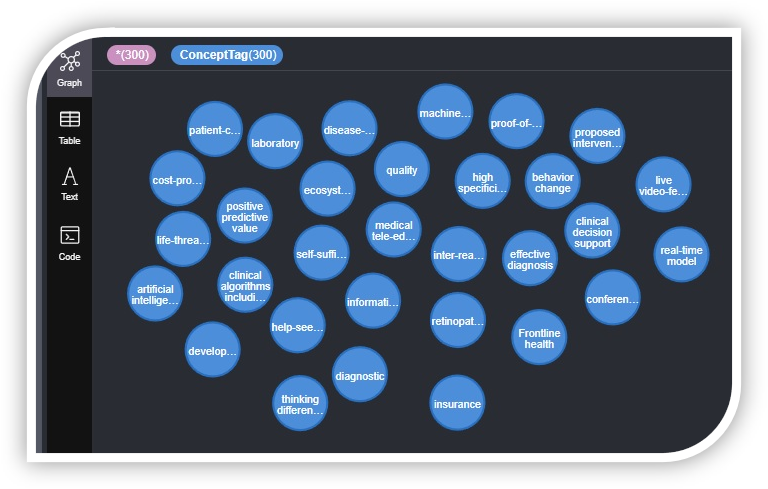}\\
    \textit{A sample of the initial unfiltered tags - Implementome's Database - neo4j}
\end{center}

\vspace{0.2cm}

\State After a first observation, we can notice that a few challenges related to the structuring of the tags present themselves. Given that they are a combination of WHO's official Digital Health Interventions Classification [\hyperlink{who_dhis}{10}] and other user-input tags related to the field without following any specific format, they highlight a disorganized structure.

\State Therefore, a filtering process needed to be performed through the initial set of $300$ concept tags. A filtered version which would serve as the basis behind the topic classification part of the project.

\begin{center}
    \underline{\textit{Filtering Process:}}
\end{center}

\State For the final labels that will pass to the topic classification pipeline, I opted out for a final clear distinct set of tags that represents somewhat of a taxonomy, albeit not official. Knowing that the publications in this domain encompass a wide range of other domains --with tags ranging from from specific parasitic diseases to financing options, political agendas, modern technologies, and beyond--, the diverse nature of tags poses a significant challenge and underscores the crucial need for a filtering process. 

\State Apart from a cleaning up process that removed incorrect or abundant data, my general filtering process was based on the following principles:

\begin{itemize}
    \item \textit{Detecting Synonyms} $\Rightarrow$ Instances of multiple tags representing the same concept were combined into one. \\ 
    For example: \textit{outbreak} $+$ \textit{epidemic} $+$ \textit{pandemic} $+$ \textit{virus} $\mapsto$ \textit{epidemic}

    \item \textit{Removing Named Entities} $\Rightarrow$ Initially, named entities (e.g. \textit{UNAID}, \textit{Switzerland}, \textit{LMICs}) were incorporated into the concept tags. They were filtered out given that their detection will be included in a pipeline of its own and I wanted to drive the tags towards a taxonomy of global health document categorization topics.

    \item \textit{Disease Instances} $\Rightarrow$ Recognition of disease instances (e.g. \textit{hypertension}, \textit{pneumonia}, \textit{leishmaniasis}) in a publication will be a custom pipe part of the named entity recognition model, therefore the task is shifted from topic classification and such instances were removed.

    \item \textit{Refining WHO's Classification} $\Rightarrow$ Some instances of DHI's classification are viewed to be a bit redundant and hence can result in an unnecessary overlapping during classification. Such thought instances were also consolidated into a single representation.\\
    For example: \textit{health information} $+$ \textit{health data analytics} $+$ \textit{electronic health records} $\mapsto$ \textit{electronic health records}

    \item \textit{Merging unigrams to bigrams} $\Rightarrow$ Multiple unigram tags were merged into bigrams representing their hierarchical parent in terms of domain. This grouping improves the overall clarity of the concept tags, allowing a more inclusive classification of publications.\\
    For example: \textit{pressure} $+$ \textit{anxiety} $+$ \textit{stress} $+$ \textit{ptsd} $\mapsto$ \textit{mental\_health}
\end{itemize}

\State This filtering process halved the total number of concept tags into a more balanced representation between granularity and coarseness. Specifically, you would no longer find excessively specific tags such as a rare disease (e.g. "\textit{echinococcosis}") or overly general ones such as "\textit{ICT}" or "\textit{quality}". Having said that, let us visualize a sample of the concept tags after the filtering process through the sunburst chart below:

\vspace{0.2cm}

\begin{center}
    \hypertarget{tag_set}{\includegraphics[scale = 0.8]{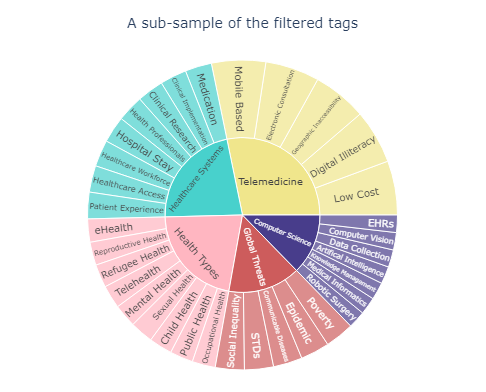}}
\end{center}

\State The graph was built with the expectation to illustrate an improvement of the tags in terms of semantics and distinction between categories. That was achieved, as evidenced by the fact that the tags currently follow a small but concise taxonomy, a fact that presents a better classification. An improved classification through this filtering process leads to better conceptualization, reasoning, language, data analysis, and finally, a more coherent document categorization process [\hyperlink{classify}{11}].\\
Nevertheless, a few challenges arise by the diversity between the parent tags in terms of their respective domains. Specifically, multiple labels of various specific sub-domains introduces complexities in the task of topic classification and the process of constructing corpora. We will proceed with the latter in the next subsection.

\subsubsection{Constructing Corpora}
\State Before beginning to answer the research question regarding the impact of corpus size on performance, we must define the corpora to be compared. The main key consideration to keep in mind is that each corpus will be constructed with publications related to our filtered set of tags. Therefore, we will be dealing with specialized corpora. \\
The most important factor when collecting text data for a specialized corpus is representation, one that should be suitable for the purpose of the research. Douglas Biber defined representativeness as 'the extent to which a sample includes the full range of variability in a population' in 1993 [\hyperlink{represent}{12}]. 

\State With that in mind, a data collection process was made to obtain specialized corpora of upmost inclusiveness towards the variety of our concept tags. Each corpus differs from the other in terms of its total vocabulary size in order to reflect the study behind our research question. The sampling process to collect publications, from which the combined texts would constitute each specialized corpus. As a result of this sampling process, the corpora are presented below in descending order of size.

\begin{itemize}
    \item \textit{Implementome Publications} $\Rightarrow$ The initial corpus constructed on the text of publications already stored in the database of \textit{Implementome}. These $33$ publications ranged from short research papers to voluminous WHO yearly guidelines. Their lack of classification in terms of concept tags also motivated me to start this project's journey in the first place.

    \item \textit{100 Publications} $\Rightarrow$ A selection process of publications in PubMed found by querying each concept tag formed this set of $100$. The selection of each publication aimed to maximize representativeness in relation to our filtered tags and prioritizing focus on real-world implementation. The latter was observed by an abundance of named entities that would be crucial for our named entity recognition task, given that this set would also be part of its evaluation. \\
    Publication Example: \textit{Study of the impact of a telemedicine service in improving pre-hospital care and referrals to a tertiary care university hospital in Nepal} [\hyperlink{neppy}{13}]

    \item \textit{100 Books} $\Rightarrow$ Following the same principles as the corpus above, another set was created with $100$ books instead of publications. This interchange into books was made with the simple goal of amplifying the corpus size and the general context, in order to compare the two.\\
    Book Example: \textit{Sexually Transmitted Infections: Adopting a Sexual Health Paradigm (2021)} [\hyperlink{booky}{14}]

    \item \textit{Pre-trained PubMed and MIMIC-III} $\Rightarrow$ \href{https://github.com/ncbi-nlp/BioSentVec#text-corpora}{\textcolor{unige}{\textit{\underline{BioWordVec}}}}, pre-trained embeddings trained on a massive corpora constituted of millions of documents found in PubMed and MIMIC-III were finally added [\hyperlink{wow}{15}]. The reason behind this addition was to compare our small specialized corpora to the state of the art of biomedical word embeddings.
    
\end{itemize}

\State To avoid introducing bias with respect to the context and to solely focus on studying the increase in size, each subsequent custom-created corpus was built by incorporating the preceding one, beginning with the Implementome corpus.

\vspace{0.5cm}

\begin{center}
    \centering
    \begin{tabular}{|c|c|c|c|m{3cm}|} 
        \hline
        Corpus & Corpus Size & Vocab Size V & $\mu_s$\footnotemark & Source  \\[2ex]\hline
        Implementome & 305474 & 6032 & 8.5 & In-house \\\hline
        100 Publications & 910324 & 15322 & 9.04 & PubMed \\\hline
        100 Books & 8924342 & 76219 & 11.12 & PubMed, Google Scholar, Universities \\\hline
        \href{https://github.com/ncbi-nlp/BioSentVec#text-corpora}{\textcolor{unige}{\textit{\underline{BioWordVec}}}} & 4B & 1B & - & PubMed, MIMIC-III \\\hline
    \end{tabular}\\[0.4cm]
    \textit{Corpus comparison in terms of total corpus size, vocabulary size and mean sentence length $mu_s$ (measured by averaging number of word tokens)}
\end{center}
 
\footnotetext{\ Mean Sentence Length}

\vspace{1cm}

\State \textit{Disclaimer:} To comply with copyright regulations, it is important to note that all downloaded materials for the creation of these corpora were limited to public domain books and publications, or through authorized access provided by the University of Geneva. 

\begin{center}
    \includegraphics[scale = 0.5]{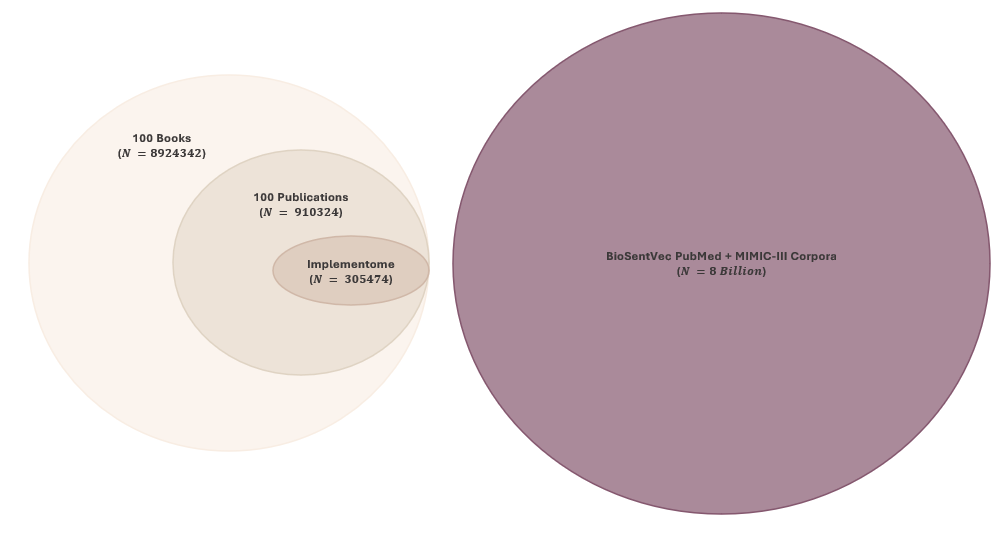}\\\
    \textit{Corpus Comparison - Overlap Visualization}
\end{center}

\vspace{0.2cm}

\State Through observing the Venn diagrams above, one can gain a clearer understanding of the variations in corpus size and the degree of overlap between them. Hence, the word embeddings derived from each corpus, varying in size from a small specialized one to a large raw text one, will make up our comparative analysis. \\
In the upcoming chapter, we will thoroughly describe the word2vec model that will produce these word embeddings. \\

\newpage

\subsection{word2vec Pipeline}
\State Both original model architectures presented in the word2vec paper [\hyperlink{word2vec}{7}] needed to be studied before proceeding with actual implementation. Continuous Bag-of-Words (CBOW) and Continuous Skip-gram are the pair architectures designed to efficiently learn distributed representations of words with minimal computational complexity while maximizing the probability in equation \hyperlink{lang_model}{(2)}, as an indication of their language modelling ability. \\
Although very similar in their common objective and training complexity with regard to total vocabulary size $N$, the two differ in terms of network design:

\begin{center}
    \includegraphics[scale = 0.5]{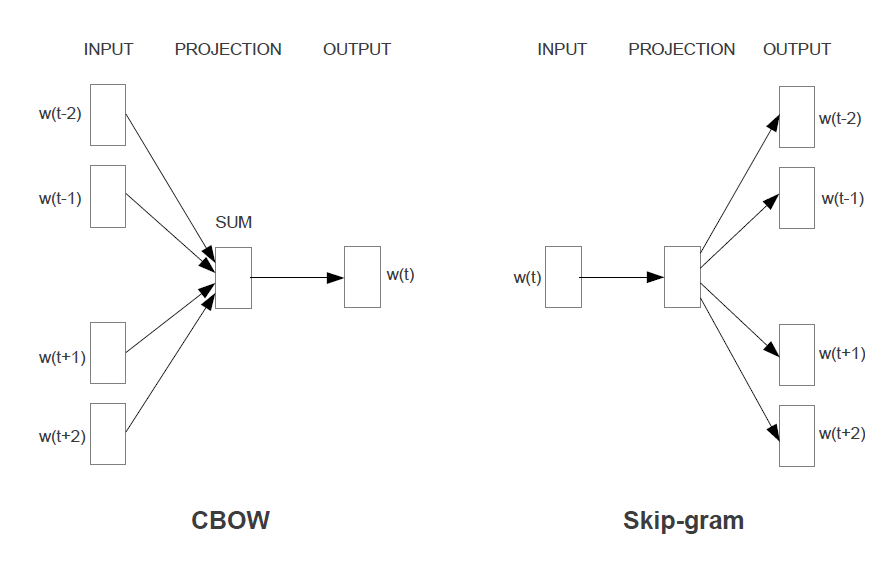}\\
    \textit{The CBOW architecture predicts the current word based on the context, and the Skip-gram predicts surrounding words given the current word.} [\hyperlink{word2vec}{7}]
\end{center}

\State As observed, CBOW sums up and averages the context words $w_c$ within a custom window to predict the target word $w$, hence maximizing: $P(w | w_c)$. \\
Reversely, Skip-gram is given the target word $w$, to predict the context words $w_c$ around it. Its objective is to maximize $P(w_c | w)$ instead.

\State Skip-gram was chosen as the model for our tag discovery task given that it excels in capturing semantic relationships, while CBOW excels in syntactic relationships [\hyperlink{skippy}{16}]. This is simply because words that occur in a similar context tend to have a similar meaning and we also want to study the effect of variations in context presented by our corpora.

\subsubsection{Text Pre-Processing}
\State After having chosen Skip-gram as the architecture, the implementation process started. \href{https://pypi.org/project/gensim/}{\textcolor{unige}{\textit{\underline{Gensim}}}} was chosen as the working framework, being the most popular Python library for unsupervised topic modelling, document indexing, and similarity retrieval with large corpora. Moreover, it offers a specialized word2vec training function that requires a list of sentences constituting the entire corpus in addition to the model's hyperparameters. Prior to specifying those, we need to address the sequence of pre-processing steps involved in transforming a folder of PDFs into a curated train list of sentences constituting the corpora.

\begin{center}
    \underline{\textit{PDF Parsing:}}
\end{center}

\State \href{https://pypdf2.readthedocs.io/en/3.0.0/}{\textcolor{unige}{\textit{\underline{PyPDF2}}}} is the open-source library used to read each PDF file in order to convert it to a text file. It extracts the main data from documentations into a text format, therefore skipping any tables or graphics that the document may contain. The extracted text from all publications constituting a corpus was saved in a single text file, creating the raw text corpus.

\vspace{0.2cm}

\begin{center}
    \underline{\textit{Text Normalization:}}
\end{center}

\State As any other raw text corpus, the produced corpus text file needed to go through a normalization process before any natural language processing is made [\hyperlink{norm_me_sire}{17}]. Specifically, the normalization function included: separation of the text into sentences, lowercasing, removal of unnecessary numerical digits or special characters, punctuation removal, and a final tokenization process to break each sentence into a sequence of discrete word tokens. \\

\State Other cleaning functions were built using regular expressions to detect certain patterns, such as the one below used to detect and remove links in a string:

\begin{center}
\begin{lstlisting}[language=Python]
import re

#### For any given string input, detect and remove links in it
def remove_links(original_sentence):
  new_sentence = re.sub(r'https?:\/\/.*[\r\n]*', '', original_sentence)
  return new_sentence
\end{lstlisting}
\end{center}

\State To ensure the quality of the training data, an additional criterion was introduced to filter out sentences without a substantial contribution to the training. Only lists of sentences that surpassed a predefined threshold of word count were retained, thus eliminating shorter sentences that might represent parts of publications such as: titles, references, or headlines. 

\vspace{0.2cm}

\begin{center}
    \underline{\textit{Automatic Phrase Detection:}}
\end{center}

\State After having normalized and tokenized the raw text corpus in order to obtain a refined set of sentences and consequently a clearer context $w_c$ for each word $w$, a final pre-processing step was implemented. Specifically, a phrase detector to combine unigram tokens to bigrams or trigrams, given that a multitude of our concept tags are of the same nature. \\

\State The main objective of it is to learn frequent co-occurrences of pairs of words and merge them into one unique token, embedded in one unique vector [\hyperlink{mikolov2}{18}]. Its implementation is offered in \href{https://radimrehurek.com/gensim/models/phrases.html}{\textcolor{unige}{\textit{\underline{gensim}}}} however I modified it by choosing Normalized (Pointwise) Mutual Information (NPMI) as the scoring function between words instead. This choice was based on its demonstrated better performance [\hyperlink{npmi}{19}], in addition to the improved results it provided in my experiments. Notably, I found that most of the automatic merges made sense semantically and resembled a potential bigram (or trigram) tag.

\vspace{0.3cm}

\begin{center}
    $\text{NPMI}(w_1, w_2) = \frac{\log \left( \frac{P(w_1, w_2)}{P(w_1) \cdot P(w_2)} \right)}{-\log(P(w_1, w_2))}$ \ (4)\\
\end{center}

\State This concept of scoring measures the co-occurrence of two words in comparison to their individual occurrence probabilities $P(w_1)$ and $P(w_2)$ in addition to the joint probability of observing the two words together $P(w_1, w_2)$. \\
The negative logarithm of the joint probability in the denominator normalizes the final score to the range $[0, 1]$ in order to obtain a more readable score and clearer threshold.

\begin{center}
    \includegraphics[scale = 0.6]{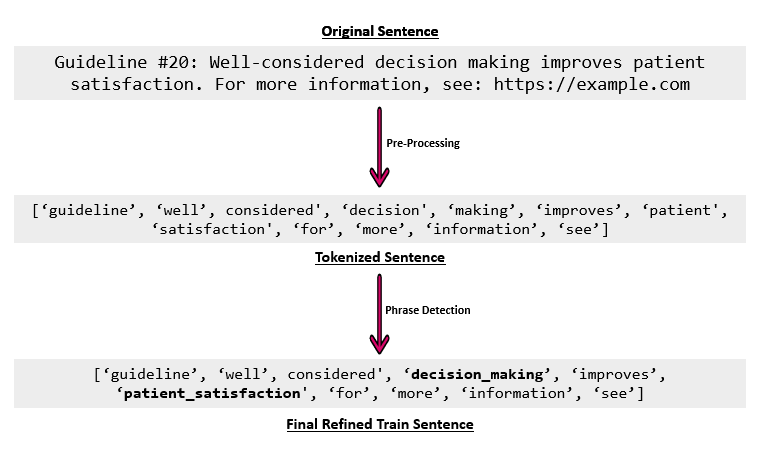}\\
    \textit{Pre-processing Pipeline Example}
\end{center}

\State Once each raw corpus text file was curated into a refined list of sentences ready to be fed to the word2vec model, training was ready to start.

\vspace{0.5cm}
 
\subsubsection{Model Training \& Hyperparameters}
\State word2vec model training could proceed after having the training data in a refined format of tokenized sentences. After initializing a model with a specific configuration of hyper-parameters, the training process starts to construct the vocabulary $V$, assigning a unique ID to each unique word token. Upon finish, word embeddings are generated for the corpus in the form of continuous vector representations.   

\vspace{0.1cm}

\begin{center}
\begin{lstlisting}[language=Python]
from gensim.models import Word2Vec

print('Word2Vec model training started... (default corpus and bigram one)')

##### Model Definition - Configuration & Vocabulary & Training
model = Word2Vec(vector_size = 500, window = 8, min_count = 3, workers = 4, sg = 1, hs = 0, negative = 10, compute_loss = True, epochs = 25)
model.build_vocab(train_set)
model.train(train_set, total_examples = model.corpus_count, epochs = model.epochs)
\end{lstlisting}
\end{center}

\State The configuration of hyperparameters stands as the most crucial task to achieve the desired performance and adaptation of the model to tag discovery. The combination visualized in the code sample above, after thorough experimentation and theoretical study, represents my final choice for this particular task. \\
Despite the importance, there is a lack of documentation or research material that describes the parameters, given the numerous modifications made to the original model over the years [\hyperlink{hyperpamparam}{20}]. Having said that, let us give a brief explanation of the main hyperparameters in the configuration our model.

\begin{itemize}
    \item \textit{Vector Size} $\Rightarrow$ This represents the dimensionality of the word embedding vectors and the value of it directly influences the model's ability to capture semantic relationships. Therefore, increasing the dimensions of the word embedding allows for a more extensive representation of the underlying semantic relationships while also leading to an increased computational complexity. \\
    Having said that, a large value of $500$ was chosen for our task to obtain more detailed representation of words without being hindered by computational complexity due to the relatively small size of our specialized corpora.

    \item \textit{Context Window Size} $\Rightarrow$ The number of words to consider as context around the target word $w$, where a value of $>1$ puts you in a multi-word context setting [\hyperlink{hyperpamparam}{20}]. A window size of $8$ was decided on as the balance between having enough local context that allows the discovery of local linguistic relationships between words, while also ensuring that the general context of a sentence is also taken into account.

    \item \textit{Skip-gram with Negative Sampling} $\Rightarrow$ The combination of parameters $sg = 1$ and $negative = 10$ \footnote{\ Number of negative samples $k$} provides you with one of the most popular word2vec setups to produce quality representations [\hyperlink{melvins}{21}]. \\
    Also known as Noise Contrastive Estimation (NCE), it challenges the model in learning to differentiate true real positive word pairs \footnote{\ A positive word pair $(w_I, w_O)$ in Skip-gram is the duo of the input word $w_I$ and the target context words $w_O$ around it.} from \textit{negative} words that are randomly sampled from the training set, similar to a noise distribution $P_n(w)$ [\hyperlink{mikolov2}{18}]. \\
    
    \vspace{0.5cm}
    
    Accordingly, its objective defined as:

    \begin{center}
        $log \ \sigma(v'_{w_O} {^\mathsf{T}}v'_{w_I}) + \sum^k_{i = 1} \mathbb{E}_{w_i \sim P_n(w)} [log \ \sigma(-v'_{w_i} {^\mathsf{T}}v'_{w_I})]$ \ (5) [\hyperlink{mikolov2}{18}]
    \end{center}

    The first term opts to maximize the probability of positive pairs $(w_I, w_O)$, consequently the similarity between their vectors $v'{w_O}$ and $v'{w_I}$, while minimizing the similarity between the negative words $w_i$ and the input word $w_I$. \\
    The presence of negative sampling can be noticed in the second term of the equation, where $k$ negative words $(w_i)$ are sampled from the noise distribution $P_n(w)$. \\
    After experimenting with suggested $k \in [5, 20]$ values for small datasets [\hyperlink{mikolov2}{18}], $k = 10$ was selected as final. It should also be noted that this approach provides a speedup in total training time, as the model needs to compare positive pairs with only the probabilities of negative samples rather those of every word in the vocabulary $V$.

    \item \textit{Number of epochs $e$} $\Rightarrow$ Even though large established corpora word2vec models are trained with a low number of epochs ($e \leq 10$) due to computational reasons, the small size of our custom corpora allowed a higher value of $e = 25$. The other reason behind my choice was that I prioritized convergence and uncovering tag similarities within my specialized corpora, rather than aiming for a generalized model that encompasses the entire global health literature. \\

    \textit{Note:} The other parameters aren't of essence to the structure of word2vec models in itself but the training process instead, such as the $workers$ parameter specifying the number of CPU cores to be used during training.
\end{itemize}

\vspace{0.1cm}

\newpage

\subsubsection{Tag Discovery}
\State The main goal of word2vec, as outlined in Mikolov's seminal work [\hyperlink{word2vec}{7}], is to map similar words to similar continuous representation vectors. Therefore, our theoretical intention to discover new tags by examining the vectors that are most similar to the vector of the initially queried concept tag could proceed, upon reaching the end of training. \\
Each element part of the constructed vocabulary $V$ of words was mapped to vectors of dimension $500$ for each corpus. Cosine similarity is used to measure the cosine angle between vectors and provide a measure of similarity, that ranges from $-1$ to $1$:

\begin{center}
    $cosine(v_{w_1}, v_{w_2}) = \frac{\sum_i v_{w_1}(i) \cdot v_{w_2}(i)}{\sqrt{\sum_i v_{w_1}(i)^2} \cdot \sqrt{\sum_i v_{w_2}(i)^2}}$ \ (6)\\[0.2cm]
    $-1 \leq cosine(v_{w_1}, v_{w_2}) \leq 1$
\end{center}
\State In implementation, the tag discovery process using cosine similarity is as follows:

\begin{center}
\hypertarget{tagitbaby}{}
\begin{lstlisting}[language=Python, escapeinside={(*@}{@*)}]
### Store the model's vectors in a variable
output_vectors = (*@\textcolor{unige}{model}@*).wv

### Tag to be queried
initial_tag = (*@\textcolor{teal}{'artificial\_intelligence'}@*)

### Output: (Tag, Cosine Similarity)
output_vectors.(*@\textcolor{unige}{most\_similar}@*)(initial_tag, (*@\textcolor{unige}{topn}@*) = 10)  
"""
(*@\textcolor{code_output}{[('machine\_learning', 0.4152183532714844),\\ 
 ('deep\_learning', 0.38648661971092224),\\
 ('computer\_vision', 0.3804324269294739),\\
 ('analytics', 0.37518802285194397),\\
 ('neural\_networks', 0.37186920642852783),\\
 ('precision\_medicine', 0.367200642824173),\\
 ('big\_data', 0.35692861676216125),\\
 ('lecture\_notes', 0.35189390182495117),\\
 ('data\_mining', 0.3500444293022156),\\
 ('data\_analytics', 0.3479711711406708)]\\}@*)
 """
\end{lstlisting}
\textit{Querying for Tag Similarity - gensim in Python}\\
\end{center}

\State As observed from the results of querying the \textit{artificial\_intelligence} tag, the word embedding process has been successful given that the majority of top similar vectors correspond to similar words. Nevertheless, the results are not perfect (e.g. \textit{lecture\_notes}) and a human curation process is needed.\\
All tags were queried using this method for each embeddings of the 4 corpora, the results of which to be discussed in the next chapter about evaluation.

\vspace{1cm}

\State \textit{Note:} If an initial tag, whether it is a unigram or bigram, is not present in the constructed vocabulary $V$ during training, this process of measuring similarity cannot take place since the tag lacks a corresponding vector representation. This presented itself as another assessment score during the corpora comparison.

\newpage

\subsection{Evaluation}
\State When evaluating a word embedding neural model such as word2vec, it is important to categorize each assessment into intrinsic or\_and extrinsic. Intrinsic evaluators test the quality of a representation independent of NLP tasks by measuring it's ability to capture linguistic meanings and represent language patterns accurately. On the other hand, extrinsic evaluators use word embeddings as input to a practical specific downstream NLP task, estimating its performance on real world applications [\hyperlink{int_ext}{22}]. 

\State To properly evaluate such a model, especially in the case of unsupervised corpora such as ours, it is important to test it in both categories [\hyperlink{int_ext_due}{23}]. Therefore, we will proceed with multiple scoring functions that are part of both categories, starting with the model comparison in terms of training time.
\\ Categorizing total training time as an extrinsic evaluation related to the computational efficiency of the model, one should mention the general training complexity of the Continuous Skip-gram architecture by itself. An architecture that all models share for which Mikolov et al. [\hyperlink{word2vec}{7}] defined the training complexity proportional to: 

\begin{center}
    $Q = C \times (D + D \times log_2 (V))$ \ (7)
\end{center}
Where C is the maximum distance of words, D the vector dimensionality, and V the vocabulary size. Based on the formula, an increase in corpus size induces an increase in computational complexity.

\vspace{0.5cm}

\begin{center}
    \centering
    \begin{tabular}{|c|c|c|c|m{3cm}|} 
        \hline
        Corpus & Corpus Size & Vocab Size V & Epochs $e$ & Training Time  \\[2ex]
        \hline
        Implementome & 305474 & 6032 & 25 & 2min \\\hline
        100 Publications & 910324 & 15322 & 25 & 7min 45s \\\hline
        100 Books & 8924342 & 76219 & 25 & 116min 53s \\\hline
    \end{tabular}\\[0.4cm]
    \textit{Model Comparison - Skip-gram $+$ Negative Sampling Training Time}
\end{center}

\State As we can observe, training time increases progressively with the growth of the corpora. The substantial jump from 7 to 116 minutes reflects the computational demand that follows large corpora. It is also influenced by our large vector dimension size of $D = 500$ in addition to the hardware \footnote{\ The models were trained on a AMD Ryzen 7 4800H 2.90GHz CPU}.

\State Having said that, it's still worth highlighting the relatively fast time of being able to train word embeddings from scratch through small specialized corpora, on a single machine. It shows the comparative performance of word2vec and specifically, its Skip-gram with Negative Sampling architecture.

\subsubsection{The Odd One Out}
\State We will begin with an intrinsic evaluation method called the \textit{odd one out}. As the name suggests, it corresponds to finding the odd word out of groups of four randomly shuffled words. Each word group was hand-crafted by me through the word embeddings generated from a test set corpus, separated by our other corpora. \\ Specifically, a random word was selected from the vocabulary $V$ and the list of its most similar vectors was generated, akin to the \hyperlink{tagitbaby}{\textit{tag discovery}} task. From the list, two randomly selected words among the top 10 most similar words were added to the group. Finally, another word from lower down the list was chosen (the odd one), thereby creating a scenario to identify the odd one out.

\begin{center}
    \includegraphics[scale = 0.5]{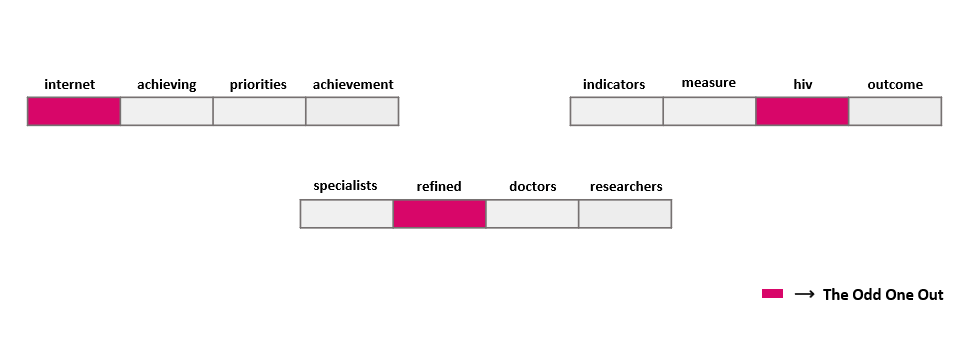}\\
    \textit{Word Groups Examples}
\end{center}

\State 1000 instances of these word groups were created with the objective to challenge the models's abilities to understand semantic and syntactic relationships by identifying dissimilar words. This is part of linguistic feature understanding thus categorizing this assessment function as intrinsic [\hyperlink{jurafsky_perp}{24}]. 

\vspace{0.5cm}

\begin{center}
    \centering
    \begin{tabular}{|c|c|} 
        \hline
        Corpus & OddOneOut Accuracy  \\[2ex]
        \hline
        Implementome & 0.628 \\\hline
        100 Publications & 0.658 \\\hline
        100 Books &  0.579 \\\hline
        \href{https://github.com/ncbi-nlp/BioSentVec#text-corpora}{\textcolor{unige}{\textit{\underline{BioWordVec}}}} & 0.487 \\\hline
    \end{tabular}\\[0.4cm]
    \textit{Model Comparison - The Odd One Out}    
\end{center}

\State Surprisingly, word2vec models trained with our small specialized corpora performed better for this task than the large pre-trained biomedical one. This observation can be explained from the fact that the testing set derived from the same corpus as the training set, both sharing similar contexts. As a result, a bias can be introduced, one that leads to a better performance and understanding for the specialized models.\\
Despite that, the average accuracy scores are still lower than initial expectations. The random selection manner in which the word groups were formed might have created a certain amount of confusing instances, consequently leading to a more challenging problem for all models. In future research, a hand-curated careful selection of word groups might be a more coherent supervised evaluation. \\
All in all, it was still an interesting evaluation method to create and study, as a small semantic understanding challenge for the models.

\subsubsection{Perplexity}
\State Another intrinsic evaluation score specialized for neural predictive models is perplexity, a modified variant of estimating the raw probability of a full test corpus. Knowing that language modelling is a probabilistic task in its nature, perplexity (sometimes called PP for short) stands as the inverse probability of the test set, normalized by the number of words [\hyperlink{jurafsky_perp}{24}].

\begin{center}
    $PP(W) = P(w_1, w_2, ...., w_M)^{\frac{1}{M}}$ \footnote{\ For a full test set $W = w_1, w_2, ..., w_M$} \ (8)
\end{center}

\State It is also presented as the held-out likelihood, deterministically transforming the log-likelihood to an information-theoretic quantity [\hyperlink{einstein_perp}{25}], deriving the perplexity score for a single word as: 

\begin{center}
    $Perplex(w) = 2^{\frac{l(w)}{M}}$\ (9), \\[0.3cm]
    $l(w) = - \sum_{m = 1}^M log_2 V = -M \ log_2 V$ \ (10) \ [\hyperlink{einstein_perp}{25}]
\end{center}

\State In our context, it measures the predictive capabilities of the word2vec models and serves as an estimation of how well they perform on unseen data and language structures.

\vspace{0.3cm}

\State \textit{Note:} Lower perplexities correspond to higher likelihoods, therefore a lower score -being less perplexed- reflects better predictive abilities of a model for a given language. In special cases where the considered word $w$ is not part of the vocabulary, the uniform probability $\frac{1}{V}$ is assigned to it, in order to avoid division by zero and continue with the calculation.

\vspace{0.3cm}

\begin{center}
    \centering
    \begin{tabular}{|c|c|c|} 
        \hline
        Corpus & Perplexity & $\frac{1}{V}$ assigned words  \\[2ex]
        \hline
        Implementome & 12.15 & 3.67\% \\\hline
        100 Publications & 9.95 & 2.37\% \\\hline
        100 Books &  16.4 & 1.151\% \\\hline
    \end{tabular}\\[0.4cm]
    \textit{Model Comparison - Perplexity} \footnote{\ As I used my corpora's localized test set to compute perplexity, it wasn't appropriate to also include the pre-trained biomedical model.}
\end{center}

\State Keeping in mind that perplexity scores can range in $[1, V]$, all of our models achieved a performant low score.  Given their domain-specific data and the high number of epochs $e$, the models have learnt to predict the context around words $w$ on unseen data. The subsequent increase in vocabulary size $V$ in the corpora is reflected by the lower number of words being assigned to the uniform probability $\frac{1}{V}$.\\
This evaluation method should be included when dealing with language models, as it nicely reflects the predictive abilities of them.

\subsubsection{Tag Discovery Comparison}
\State Having discussed the language understanding and predictive abilities of our models, we will now employ an extrinsic evaluation method to investigate the impact of corpus size for our research question. This evaluation method focuses on the practical application part of this task, tag discovery with regard to cosine similarity.

\State Therefore, we will compare the similarity query results of our filtered set of tags across each corpus to determine the corpus that provides the most relevant and semantically similar new tags. Hence, this is a subjective comparative evaluation where people are asked about their preferences among different word embeddings [\hyperlink{int_ext}{22}], rather than an absolute mathematical evaluation such as the methods described above. \\
If the exact tag as written on our \hyperlink{tag_set}{\textit{set}} is not present in the vocabulary constructed by the model, the query result will be empty since there is no corresponding vector for that tag. Thus, let us observe how the corpora differ on this small but crucial detail.

\begin{center}
    \centering
    \begin{tabular}{|c|c|c|c|} 
        \hline
        Corpus & Vocab Size V & Unfound Unigrams & Unfound Bigrams \\[2ex]
        \hline
        Implementome & 6032 & 24.19\% & 78.22\% \\\hline
        100 Publications & 15322 & 16.13\% & 69.13\% \\\hline
        100 Books &  76219 & 0.00\% & 40.49\% \\\hline
        \href{https://github.com/ncbi-nlp/BioSentVec#text-corpora}{\textcolor{unige}{\textit{\underline{BioWordVec}}}} &  1B & 1.69\% & 23.76\% \\\hline
    \end{tabular}\\[0.5cm]
    \textit{Model Comparison - Unfound Tags}\\[0.5cm]

    \includegraphics[scale = 0.35]{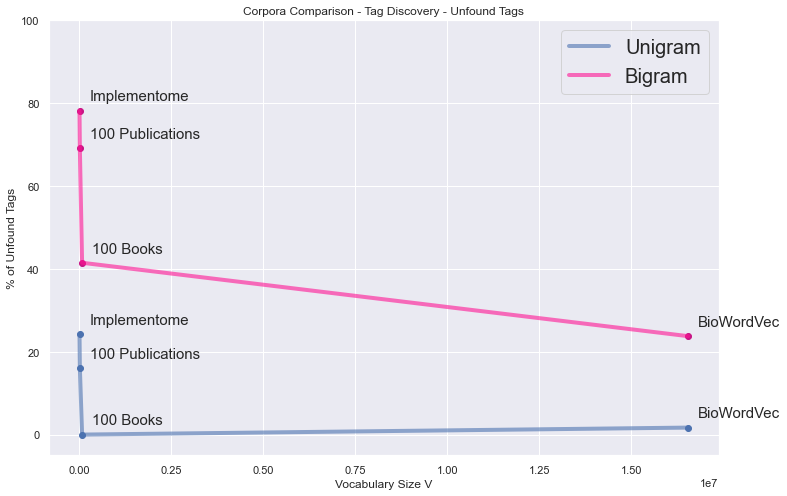}
\end{center}

\State As expected, an increase in vocabulary size derived from a larger corpus results in the localization of more tags. The performance of the specialized corpus, \textit{100 Books}, compared to the pre-trained model is an important topic of discussion regarding our question. Despite having a significantly smaller vocabulary (13000x less), the specialized corpus shows noteworthy performance.

\State Let us deduce if the same performance holds true in the comparative analysis during tag discovery too. Even though it stands as a subjective analysis, examples of query results that clearly indicated the meaningful semantic differences in corpora will be showed, for both unigram and bigram cases. A sub-sample of comparative results that were agreed on by global health experts in the team and that reflected the general performance for all tags.

\vspace{0.1cm}

\begin{center}
    \includegraphics[scale = 0.36]{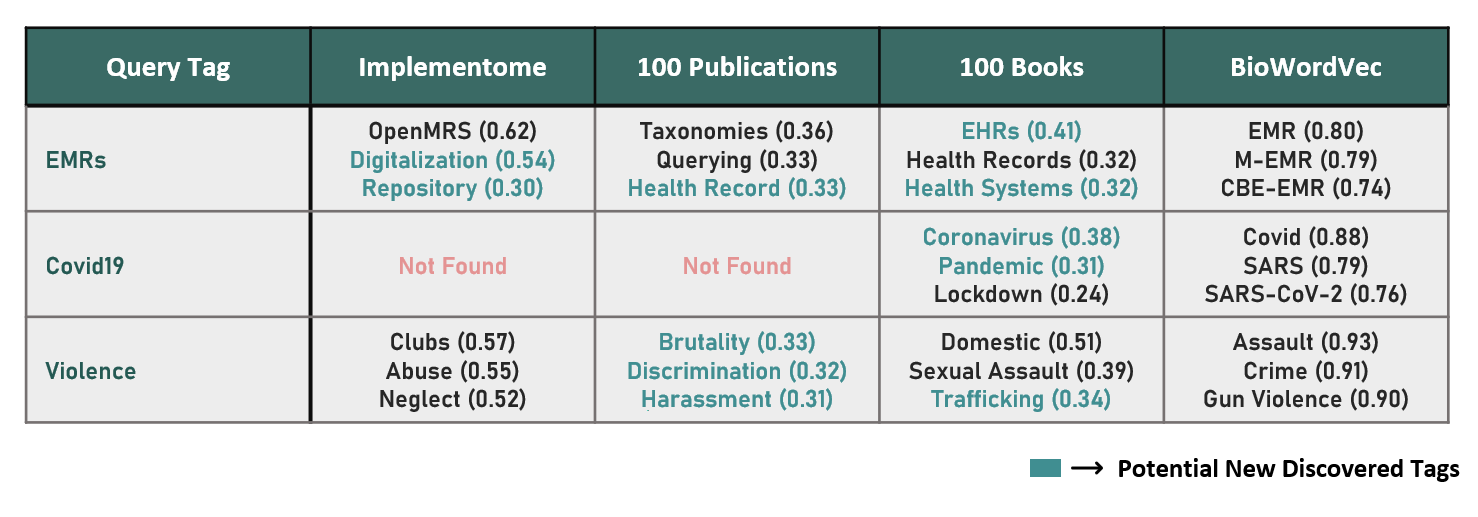}\\
    \textit{Tag Discovery Comparison - Unigrams}\\[0.5cm]

    \includegraphics[scale = 0.39]{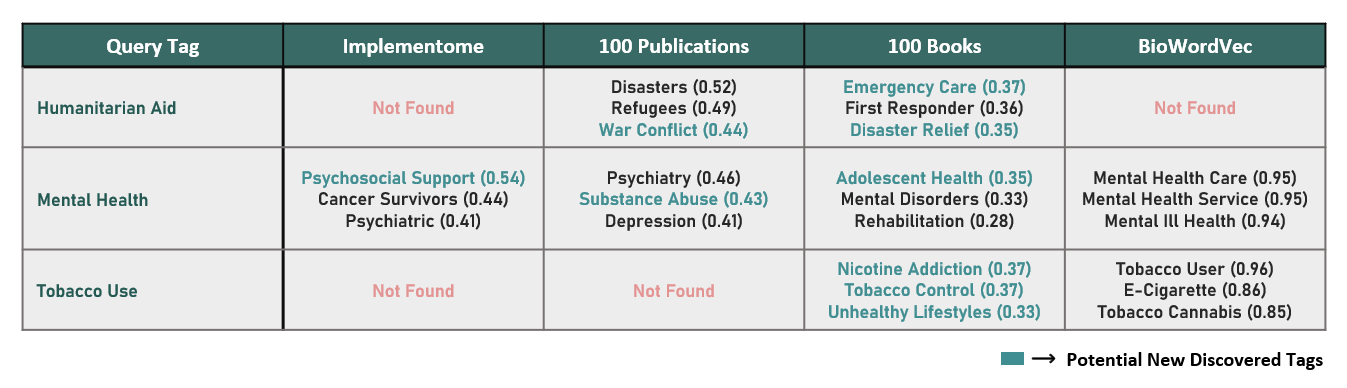}\\
    \textit{Tag Discovery Comparison - Bigrams}\\[1cm] 
\end{center}

\begin{center}
\underline{\textit{Remarks:}}    
\end{center}

\State It should be noted that the results were selected from the first 20 most similar words with regard to the queried tag, therefore the top 20 most similar vectors in terms of cosine similarity. It should be carefully observed how our models trained on specialized corpora, albeit not very generalizable to all tasks, have been adapted to our task of finding similar tags in terms of semantics and similar surrounding context.  

\vspace{0.2cm}

\State While the pre-trained \href{https://github.com/ncbi-nlp/BioSentVec#text-corpora}{\textcolor{unige}{\textit{\underline{BioWordVec}}}} model captures absolute true synonyms and overlapping $n$-grams (e.g. \textit{Violence} $\mapsto$ \textit{Gun Violence}) with a high $cosine \geq 0.85$ score, our models equipped with a small specialized corpora capture similar words that would appear in the same context (e.g. \textit{Tobacco Use} $\mapsto$ \textit{Nicotine Addiction}). This exact feature allowed us to discover new tags that may not initially seem semantically similar yet are mapped to similar vectors a cause of appearing in similar context. \\
The main reasons behind this difference between the specialized embeddings and the pre-trained ones are purely architectural. Skip-gram aiming to maximize predictions of similar contexts allows us to use the latter to find similar target words while also learning dissimilar ones due to negative sampling. Therefore, it can also be seen as somewhat of a fine-tuning of word2vec to our tag discovery task.

\State Regarding the differences between the models trained within my selection of corpora, we can observe the parallelism in performance and corpus size through the performance of the model trained on \textit{100 Books}. In further work, I would propagate further the size of it, while staying true to the specialized context of global digital health to avoid the generalizations seen in the results of pre-trained embeddings. 

\State Apart from the study, the results of this process allowed me to enrich the existing set of tags in order to create a hierarchical representation of them. One that would aid me in having an equally distributed taxonomy of labels during the \textit{topic classification} part of this project.

\subsubsection{Visualization by Dimensionality Reduction}
\State Despite their magic of mapping word analogies to analogies in a mathematical space, little is known about the structure and properties of these spaces [\hyperlink{viz_problem}{26}]. Knowing that we are dealing with high dimensional vectors ($D = 500$ in our case), we project it to a comprehensible lower dimension able to be visualized. Researchers commonly use t-distributed stochastic neighbour (t-SNE) or Principal Component Analysis (PCA) to project these vectors to 2D [\hyperlink{viz_problem}{26}].

\State We will use the former method as a tool to visualize high-dimensional data, implemented using \href{https://scikit-learn.org/stable/modules/generated/sklearn.manifold.TSNE.html}{\textcolor{unige}{\textit{\underline{scikit-learn}}}}. Two projections will be made, one for the unigram embeddings and another for the bigram ones. Only the output continuous vectors produced from \textit{100 Books} will be visualized, considering it as the best model.

\begin{center}
\begin{lstlisting}[language=Python, escapeinside={(*@}{@*)}]
from sklearn.manifold import TSNE
num_dimensions = 500

##### TSNE Dimensionality Reduction
tsne = (*@\textcolor{unige}{TSNE}@*)(n_components = num_dimensions, random_state = 0)
vectors = tsne.(*@\textcolor{unige}{fit\_transform}@*)(vectors)
\end{lstlisting}
\textit{t-SNE - sklearn}\\[1.5cm]
    \includegraphics[scale = 0.27]{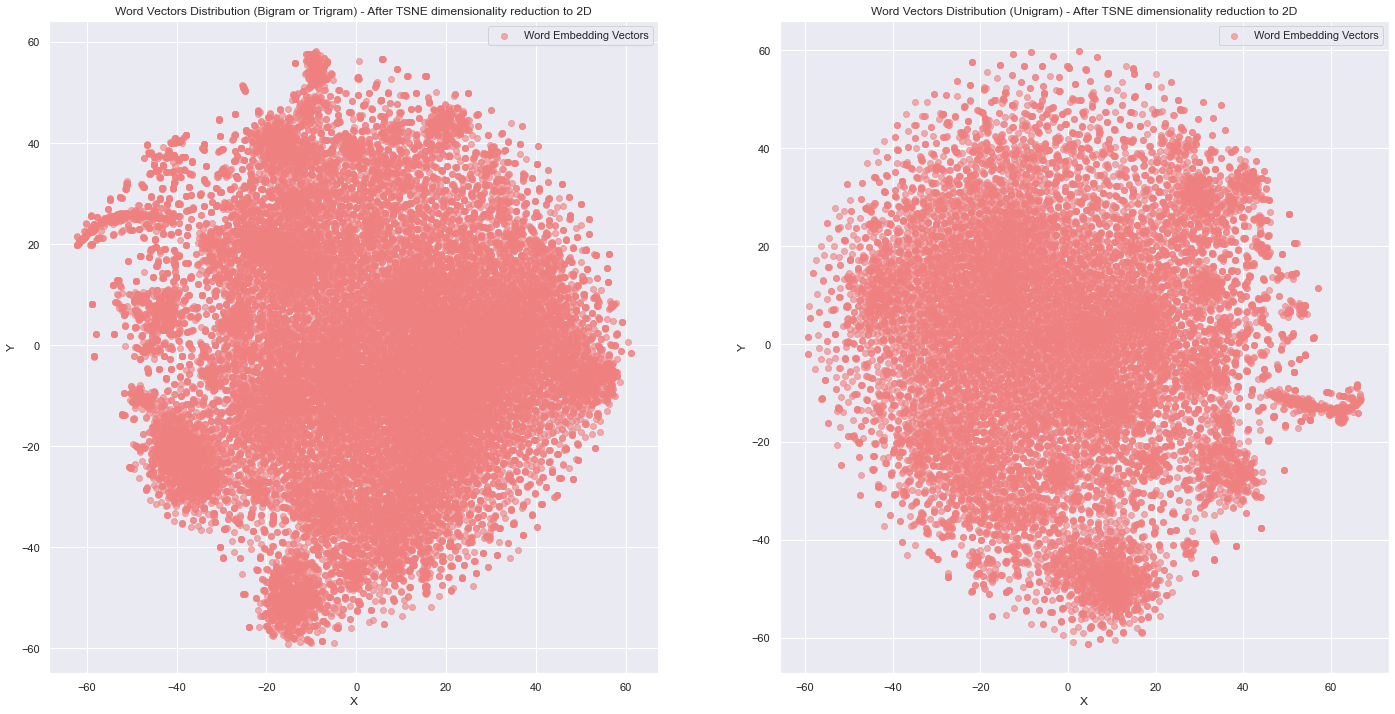}\\
    \textit{2D Visualization - Word Embeddings trained from \textit{100 Books}}
\end{center}

\State Even after dimensionality reduction, the complex and dense nature of our vectors makes it challenging to obtain clear separations that display patterns. The tremendous overlapping between vectors as a consequence of the very similar context that they share in our specific domain can also be observed from the graphs, revealing its non-linear nature. 
 
\vspace{0.2cm}

\State Each time you are faced with this problem in data mining or visualization, you turn your attention to clustering methods. \textit{k-means} and \textit{spectral clustering} were selected and implemented as clustering methods able to identify clusters in a non-linearly separable input space [\hyperlink{clustering}{27}]. The former clustering the data around a pre-defined number of centroids while the latter performing the clustering by exploiting the eigenvectors from a matrix derived from the data.

\begin{center}
    \includegraphics[scale = 0.25]{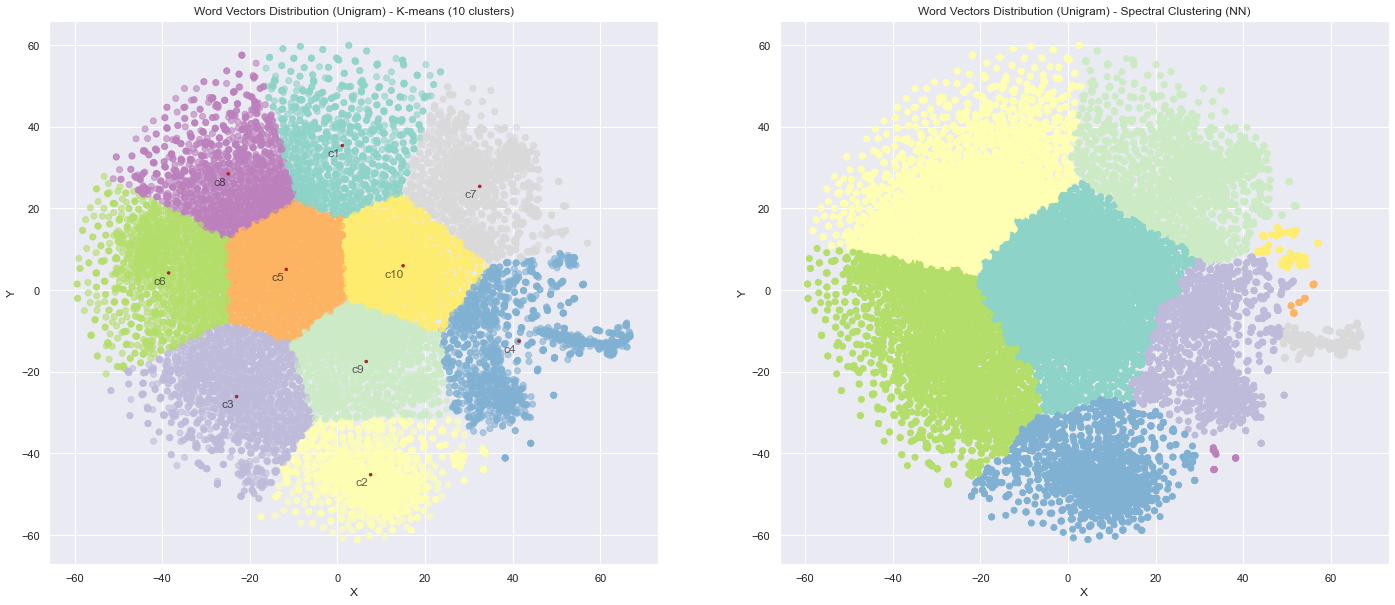}\\
    \textit{k-means and Spectral Clustering - Unigram}\\[0.5cm]
    \includegraphics[scale = 0.25]{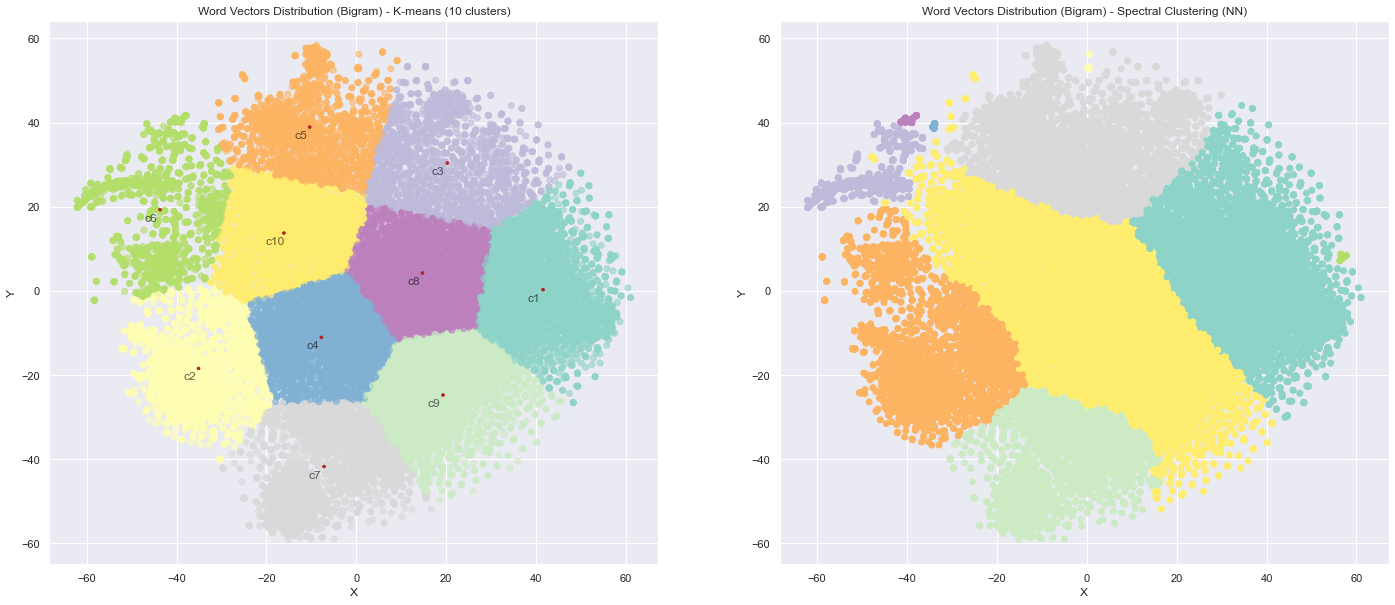}\\
    \textit{k-means and Spectral Clustering - Bigram}
\end{center}

\State Both clustering methods did not manage to provide a clear meaningful separation of the word embeddings. The confusion and lack of clear patterns in the low-dimensional data highlight one of the limitations or weaknesses of the word2vec model. Therefore, I opted to create a more human-curated visualization graph. One inspired by [\hyperlink{viz_me}{28}], and aimed at confirming word2vec's statement, where similar words are mapped to similar vectors.

\newpage

\begin{center}
\underline{\textit{Semantic Cluster Groups:}}    
\end{center}

\State Drawing inspiration from the tag discovery process, this form of clustering was created to visually represent the mathematical mapping of similarities described in that chapter while also ensuring global readability. \\
For both unigram and bigram clusters, the model's ability to capture semantic relationships is proven by the local proximity of similar tags. Nevertheless, the positioning is still unclear and subject to discussion.

\begin{center}
    \includegraphics[scale = 0.55]{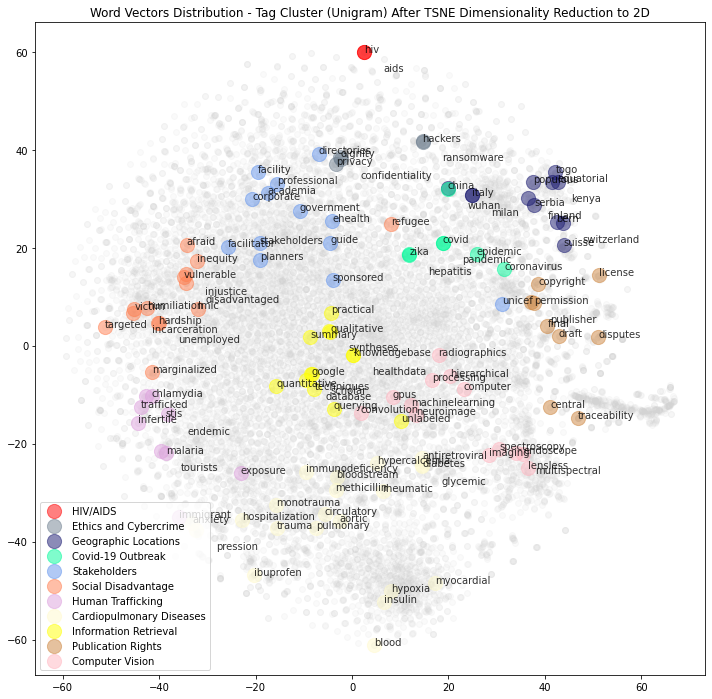}\\
    \textit{Semantic Clustering - Unigram}\\[0.5cm]
    \includegraphics[scale = 0.55]{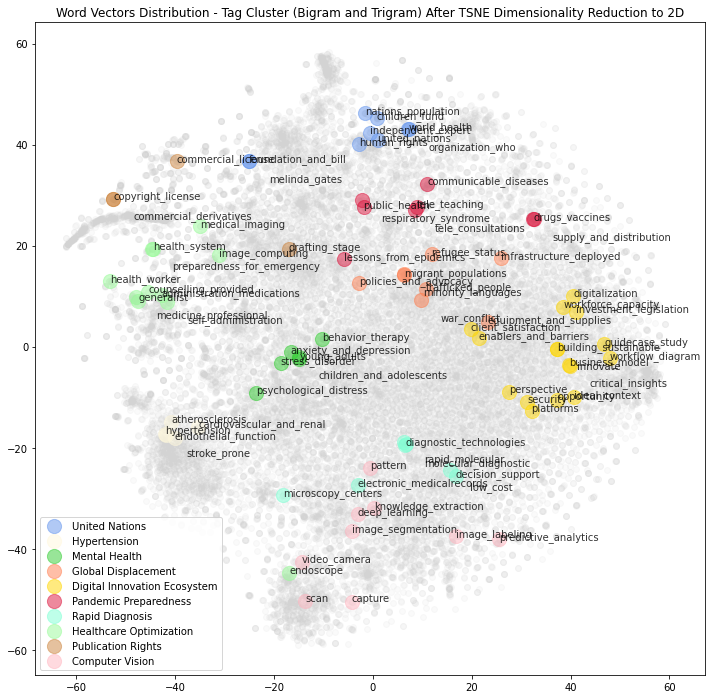}\\
    \textit{Semantic Clustering - Bigram}
\end{center}

\vspace{1cm}

\State After having concluded the theoretical study by comparing different levels of specialized corpora and a large pre-trained model, as well as applying NLP techniques for tag discovery, it is now time to shift our focus towards automatically identifying named entities in global health documents.

\newpage

\section{Named Entity Recognition}
\State Named entities are crucial in any knowledge management system, especially in those dedicated to global health that are built with the objective of sharing knowledge and connecting people, patients, professionals, shareholders, organizations, and even countries, globally. Therefore, the submission of accurate and high-quality named entities, free from any missing data, serves as a critical foundation for achieving seamless interconnectivity. This fact served as my main motive behind my choice to make named entity recognition the central focus of this project.

\State Standing as one of the most important sub-fields of global NLP research, \textit{Named Entity Recognition (NER)} is a classical problem in information extraction, able to recognize and extract mentions of named entities in text [\hyperlink{eisenstein_ner}{29}]. These can usually be people, locations, organizations, and geopolitical entities, as presented below through visualizing an element of one of the most historically renowned NER datasets, CoNLL-2003 [\hyperlink{artcapja}{30}]. 

\begin{center}
    \includegraphics[scale = 0.5]{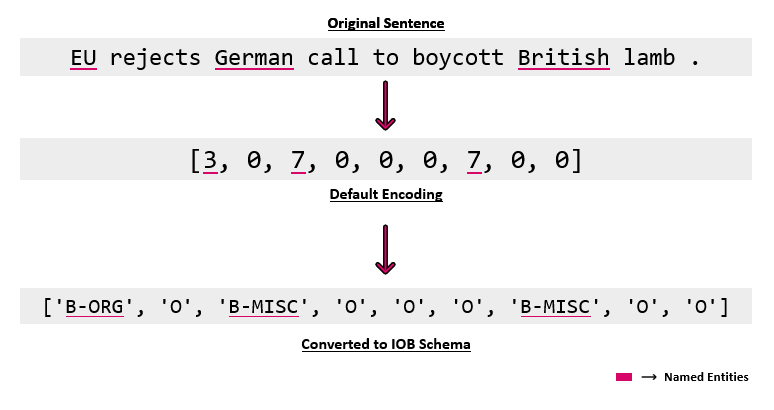}\\
    \textit{CoNLL-2003 - First Training Sentence} 
    \footnote{\ The IOB schema is a popular tagging format in NER, enabling accurate entity identification in text and its scoring functions.} \ [\hyperlink{artcapja}{30}]
\end{center}

\State Recently, the NER task has been expanded to include amounts of money, percentages, dates, times, etc. Most importantly, it has been revealed as a key task in biomedical NLP, given the custom labelling of proteins, DNA, RNA, cell lines, diseases, medications, organisms, mutations, as entity types [\hyperlink{eisenstein_ner}{29}]. In this project, we will witness this property through the deployment of a disease recognition model.

\subsection{Objectives}
\State There is an extrinsic evaluation objective at the core of this NER project, developing a full working pipeline able to localize the main entities in a real world global health publication. With that in mind, \href{https://spacy.io/}{\textcolor{unige}{\textit{\underline{spaCy}}}} \footnote{\ https://spacy.io/} was selected as the reliable NLP framework offering fast industrial-strength information extraction trained pipelines.

\vspace{0.2cm}

\begin{center}
    \underline{\textit{Ready to Deploy Pipeline:}}
\end{center}

\State By the end of this project, my aim was to have a fully developed and functional back-end pipeline specialized in global health, ready for deployment in industrial envrionments and applications. Its primary goal being to provide a named entity recognition (NER) option that can be applied to user uploaded or existing publications. Furthermore, users should be able to run the NER process on their documents, allowing for the real-time localization and identification of key entities, which should be presented to the user upon completion. This would free human annotation time from global health experts, one of the main industry need factors, discussed in my \hyperlink{motivy}{\textcolor{unige}{\textit{\underline{motivation}}}}.

\State Considering the nature of our domain, additional requirements emerged. Given that low connectivity speeds and software size limitations remain significant challenges in digital global health, especially in LMICs \footnote{\ Low and Middle-Income Countries} [\hyperlink{lmicsL}{31}], the pipeline should prioritize speed and retain a reasonable model size, to emphasize accessibility. 

\vspace{0.2cm}

\begin{center}
    \underline{\textit{Evaluation of spaCy:}}
\end{center}

\State Having defined the practical importance and objectives of the project, the theoretical goal presented itself, in order to ensure optimal performance for both objectives. This involved conducting a formal evaluation of the selected framework and its pre-trained NER models, specifically assessing their adaptability to our domain-specific task.  Concretely, a formal comparison of the model's state of the art architectures, their fine-tuning capability, and their ability to adapt to new custom biomedical labels. At the end of this evaluation, one should have a better theoretical understanding of named entity recognition models and frameworks.

\subsection{Model Comparison (spaCy)}
\State As an open-source library for natural language processing, \href{https://spacy.io/}{\textcolor{unige}{\textit{\underline{spaCy}}}} puts you in the disposal of multiple pre-trained models adapted to a number of tasks. Adapted with a simple and intuitive API, it makes it the first choice for a framework when the objective is a fast and accurate implementation. Through its streamlined processing pipeline, it can start performing NLP tasks by only a raw text sentence as input. 

\begin{center}
    \includegraphics[scale = 0.6]{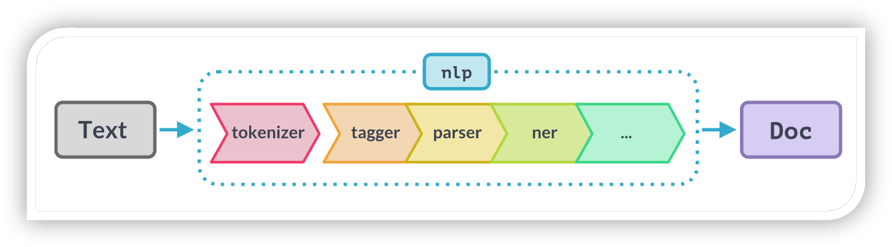}\\
    \href{https://spacy.io/usage/processing-pipelines}{\textcolor{unige}{\textit{\underline{spaCy - Language Processing Pipeline}}}}
\end{center}

\State As visualized above, all the various NLP tasks are constituted in the \textit{nlp} pipeline, converting text into a structured \textit{Doc} object that holds token information in addition to any results obtained during processing. The \textit{tokenizer} pipe performs tokenization, a necessary opreration given that segmented word tokens are expected as input to any further processing.

\vspace{0.1cm}

\State Our focus will rest towards the \textit{ner} component, given that it is tasked with named entity recognition of the text. It saves any information of found named entities in the \textit{.ents} attribute of the \textit{Doc} object. A useful attribute, in terms of debugging, that allows you to extract entities, their types, and their positions with regard to the sentence.

\begin{center}
    \includegraphics[scale = 0.52]{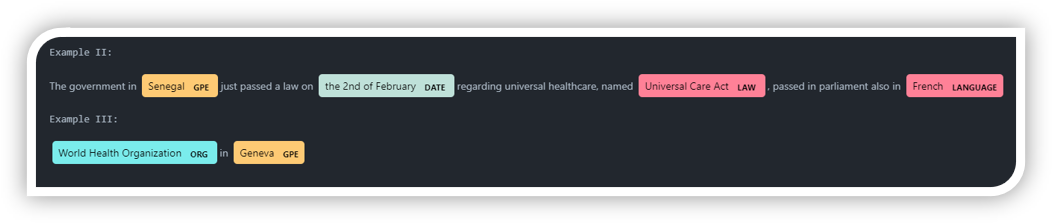}\\
    \textit{NER Component: Example Results}
\end{center}

\State The visualization of the results above is achieved through \href{https://demos.explosion.ai/displacy-ent}{\textcolor{unige}{\textit{\underline{displaCy}}}}, a great named entity visualizer library and one that we will use throughout these chapters. Having said that, we could notice the impressive out-of-the-box performance of the pre-trained model to localize multiple labels of named entities. \\
Out of all possible labels, the pipeline component was modified to focus on only extracting entities of type: \textit{PERSON, GPE, ORG, LAW} and \textit{PRODUCT}.

\begin{itemize}
    \item \textit{PERSON} $\Rightarrow$ The clear intention to identify individuals related to, or referenced in a publication. They would be considered main entities if multiple counts of the same individual occur.
    \item \textit{GPE \& ORG} $\Rightarrow$ Two crucial labels representing geopolitical entities and organizations. Given the nature of global health publications, geopolitical entities can be countries, regions, continents, intergovernmental bodies where a project is implemented while organizations can range from funders, NGOs, professional associations, academia, international players, etc. An accurate localization of these two entity types is the most crucial objective of our NER pipeline, knowing that they cannot be easily distinguished solely through publication metadata such as titles or keywords and you need to read it to locate them.
    \item \textit{LAW \& PRODUCT} $\Rightarrow$ While not being required entity types for our task, recognition of laws and products was incorporated to measure their accuracy and enrich the model overall. In terms of the domain, laws can be public health laws that are either referenced or studided in a publication while products represent specific names of medical devices, digital health publications, vaccines, or other products.
\end{itemize}

\State Having defined the general processing pipeline, let us move towards a more formal presentation of the pre-trained up for comparison models.

\subsubsection{The Two Architectures}
\State Our comparison will include four pre-trained spaCy models that are specialized in English. This choice was made knowing that the majority of global health research is published in english, however this task can be adapted to any language, by merely selecting another model. All models are able to recognize our selection of entity types given that they are trained on news data. Specifically, the 300-dimensional GloVe [\hyperlink{glove}{32}] vectors of the models are trained from Ontonotes 5.0 [\hyperlink{ontomom}{33}] as a large annotated corpus containing broadcast and newswires, Wordnet 3.0 [\hyperlink{wordie}{34}] as a popular large lexical database of english, and \href{https://commoncrawl.org/}{\textcolor{unige}{\textit{\underline{Common Crawl}}}} as an additional supplier of web crawl data. This enormous amount of train data was another reason to utilize the knowledge of these pre-trained models, rather than training one from scratch.

\State Out of the total four, three models are based on a Convolutional Neural Network (CNN) architecture while the other model is based on a transformer one. While all of them use the same word embedding method, the difference is found in their vocabulary size $V$ \footnote{\ $V =$ Number of unique word vectors}.

\begin{center}
    \centering
    \begin{tabular}{|c|c|c|c|m{2cm}|} 
        \hline
        Model & Base Architecture & Size & Vocab Size V & Preferred Device  \\[2ex]\hline
        \href{https://github.com/explosion/spacy-models/releases/tag/en_core_web_sm-3.5.0}{\textcolor{unige}{\textit{en\_core\_web\_sm}}} & CNN & 12 MB & 0 & CPU \\\hline
        \href{https://github.com/explosion/spacy-models/releases/tag/en_core_web_md-3.5.0}{\textcolor{unige}{\textit{en\_core\_web\_md}}} & CNN & 40 MB & 20k & CPU \\\hline
        \href{https://github.com/explosion/spacy-models/releases/tag/en_core_web_lg-3.5.0}{\textcolor{unige}{\textit{en\_core\_web\_lg}}} & CNN & 560 MB & 514k & CPU \\\hline
        \href{https://github.com/explosion/spacy-models/releases/tag/en_core_web_lg-3.5.0}{\textcolor{unige}{\textit{en\_core\_web\_trf}}} & Transformer & 438 MB & 0 & GPU \\\hline
    \end{tabular}\\[0.4cm]
    \textit{spaCy pre-trained default models comparison}
\end{center}

\State From observation, one can understand that $V$ directly affects the model size in the case of the CNN ones. A small vocabulary gives us the ability to obtain compact models of a very low size, allowing us to analyze isolated real world cases where disk space is a constraint. \\
On the other hand, vocabulary size is null for the transformer model given that it doesn't rely on traditional word embedding vectors. Instead, the architecture leverages contextualized embeddings that capture the meaning of words based on the context of other surrounding words, such as in BERT [\hyperlink{berti}{35}]. This method of learning embeddings is a heavily parallelized process, henceforth GPU is recommended as the device to optimally run them.

\vspace{0.1cm}

\State Having presented all models in terms of architecture, memory size, and vocabulary size, it is time to provide a more detailed introduction of their architectures. While the exact specific architecture with the full number of layers and elements inside them is disclosed by spaCy, we can nevertheless broach the differences between convolutional neural networks and transformers in terms of sequence processing.

\begin{center}
    \underline{\textit{CNNs for Sentence Classification}}
\end{center}

\State Through utilizing pre-trained word embedding vectors as 'universal' feature extractors, convolutional neural networks have shown impressive performance on NLP benchmarks, even with architectures composed of a single convolution layer [\hyperlink{cnn1}{36}]. Using convolutions, CNNs excel at capturing local features in data which in the case of NLP are important linguistic features such semantics, syntax, and context of a text. \\

\State Letting $x_{1:n} = \{x_1, x_2, ... , x_n\}$ represent a concatenation of words, a convolution operation applies the filter $w$ to a custom-length window of $h$ words in order to produce the feature $c_i$, a final condensed representation holding the most important features. 

\begin{center}
    $c_i = f(w \cdot x_{i:i+h-1} + b)$ \ (11)
\end{center}

\State This simple yet powerful mathematical operation would be expressed through an architectural visualization as:

\begin{center}
    \includegraphics[scale = 0.4]{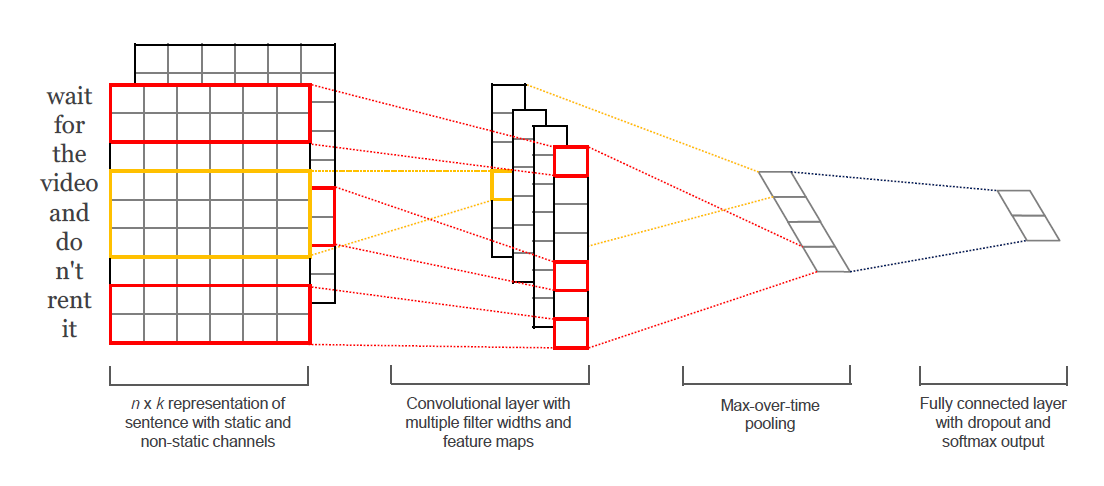}\\
    \textit{CNN for Sentence Classification - Model Architecture Example} [\hyperlink{cnn1}{36}]
\end{center}

\vspace{0.1cm}

\State In the case of our CNN-based spaCy models, convolutional neural networks apply their detection filters in shared convolution layers between the part-of-speech tagger, dependency parser and named entity recognizer pipeline [\hyperlink{spacy_book}{37}]. This parameter sharing within the model's language processing pipeline reduces the parameter count and enhances generalization, resulting in faster models.

\State In the context of our models, instead of analyzing each word, the convolutional layers process chunks of $n$ words. Therefore considering that we are dealing with 300-dimensional GloVe\footnote{\ Global Vectors for Word Representation [\hyperlink{glove}{32}]} vectors, they would be applied to $n \times 300$ matrix representing the vectors of the word chunks [\hyperlink{spacy_book}{37}]. 

\State This provides sufficient context needed for efficient sentence classification and in the case of a part-of-speech (PoS) tagger pipeline, it would be visualized as:

\vspace{0.3cm}

\begin{center}
    \includegraphics[scale = 1.1]{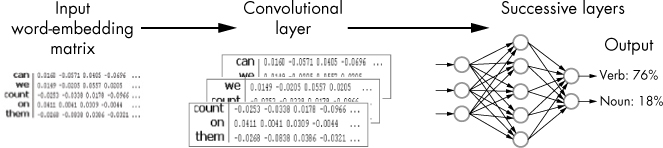}\\
    \textit{Convolutional Approach for PoS Tagging: Conceptual Look} [\hyperlink{spacy_book}{37}]
\end{center}

\vspace{0.3cm}

\State In the case of NER, the conceptual framework would be the same albeit with a different output, predicting entity labels instead of part-of-speech tags.

\State The simplicity and effectiveness of convolutional operations gives us the opportunity to have compact and efficient models at our disposal such as our selection.

\vspace{1cm}

\begin{center}
    \underline{\textit{Transformer-Based}}
\end{center}

\State As mentioned above, the \textit{en\_core\_web\_trf} model differs from the rest of the models by relying on learning contextualized embeddings based on the encoder part of the transformer architecture rather than utilizing word embeddings such as word2vec. Entirely dispensed of recurrence and convolutions, it only uses the attention mechanism to dynamically weigh the contextual importance of each word in a sentence [\hyperlink{attention}{38}].

\State The attention mechanism's superior performance in understanding relationships and dependencies between words revolutionized the NLP field, nevertheless accompanied by a higher computational cost in comparison to CNNs.

\begin{center}
    \includegraphics[scale = 0.6]{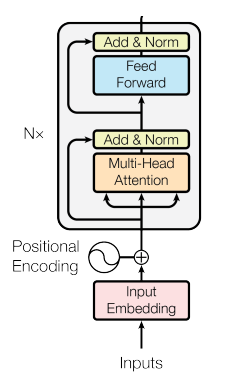}\\
    \textit{Original Transformer's Encoder Part - Attention Is All You Need} [\hyperlink{attention}{38}]
\end{center}

\State Specifically, spaCy's model is based on the base version of RoBERTa [\hyperlink{roberti}{39}], a variant of the BERT (Bidirectional Encoder Representations from Transformers) [\hyperlink{berti}{35}] model. This variant simply enhanced BERT's performance through a longer training process with optimized hyperparameters, achieving better results on various natural language processing tasks [\hyperlink{roberti}{39}]. \\ Therefore, as in BERT, our model is equipped with the ability to generate word representations that are influenced by the context of both preceding and succeeding words, thus bidirectional.

\vspace{0.1cm}

\State Having provided a brief introduction to the two architectures, let us highlight another property that all models incorporate into their NER pipeline, which is their dependency parsing capability when performing named entity recognition.

\vspace{0.3cm}

\begin{center}
    \underline{\textit{Dependency Parsing in NER}}
\end{center}

\State In most languages, syntactic dependencies exist between groups of two or more words in a sentence. This relation is usually held between a syntactically subordinate word, called the \textit{dependent}, and another word called the \textit{head} [\hyperlink{dependent_book}{40}]. For example, the noun \textit{news} is a dependent of the verb \textit{had} with the dependency type subject(SBJ) [\hyperlink{dependent_book}{40}].

\State In relation to NER, the incorporation of a dependency parsing tree leads to a better performance in the recognition of nested entities [\hyperlink{nested_ent}{41}].
A \textit{nested entity} is a named entity that contains references to other named entities (e.g. Canton of Geneva), as opposed to a single flat entity (e.g. Geneva).

\begin{center}
    \includegraphics[scale = 0.55]{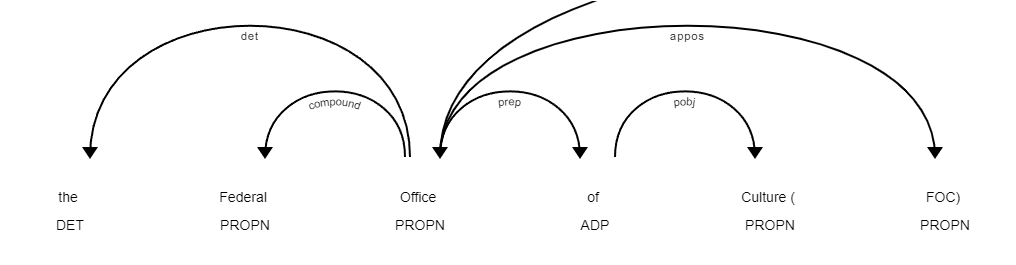}\\
    \textit{Dependency Parsing Example in a Nested Entity}
\end{center}

\State Given that the majority of early NER research was focused on the recognition of flat entities while most real-life entities tend to be of a nested nature, the inclusion of automatic dependency parsing systems improved NER results across all models [\hyperlink{nested_ent}{41}]. These systems predict a \textit{head} for each token, in order to automatically assign \textit{head $\mapsto$ dependent} relations, based on a general grammatical knowledge of the language. \\
Formally, given a Sentence $S = \{w_1, w_2, ..., w_i\}$, a dependency arc is the set $A = \{(h_i, r_i, d_i)\}$ with $h_i$ being the head word, $r_i$ the syntactic relationship, and $d_i$ the dependent word.

\State Intuitively, the fact that words in a nested entity tend to follow a pattern in terms of their usual syntactic relationships, stands at the core of the improvement in performance. In particular, the spaCy models incorporate a non-monotonic transition system for dependency parsing, one that permits parsers to fix their previous parsing mistakes on the fly, to result in larger accurate parsing trees [\hyperlink{transition_parser}{42}].  

\vspace{0.3cm}

\subsubsection{Scoring Function}

\State Sequence labeling evaluation is performed using \href{https://github.com/chakki-works/seqeval}{\textcolor{unige}{\textit{\underline{seqeval}}}} \footnote{\ https://github.com/chakki-works/seqeval} as the Python framework specialized for it. Respective lists of true values and predictions on a token level following the IOB schema will be passed as input to produce a classification score. \\
Inside-Outside-Beginning (IOB) is a widely popular tagging schema for NER that offers a simple yet effective to accurately compute the score of a model over entity spans. Specifically, it improves readability its the three-label structure of marking each token as the beginning with an entity with \textbf{B-}, the ensuing tokens within the entity with \textbf{I-} and any other word token that is not an entity with \textbf{O}.

\vspace{0.1cm}

\begin{center}
    \includegraphics[scale = 0.5]{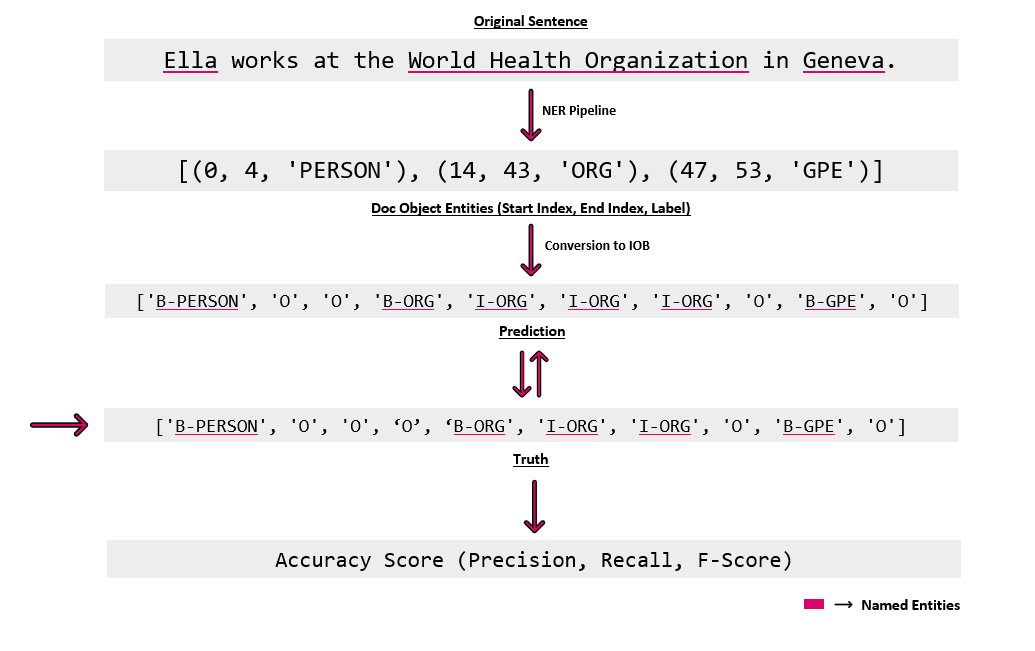}\\
    \textit{Scoring Schema}
\end{center}

\vspace{0.1cm}

\State The accuracy in the classification score will be a measure of \textit{\ precision, recall, and the $F$-Measure} [\hyperlink{scores}{43}] as the harmonic mean of the two. 

    \begin{center}
        \hypertarget{fff}{$F = \frac{2(r * p)}{r + p}$ \footnote{$r$ being recall and $p$ precision}} \ (12)
    \end{center}

\State \textit{Note: } Two types of scoring functions emerged from the displayed graph above: a strict one that requires exact index predictions of the entity spans to consider them accurate, and a more lenient one that counts correct matches of IOB labels per token. \\
Given the nature of our task and the fact that we are not dealing with entities of a delicate type such as biomedical ones, the lenient scoring function will be used for evaluation. (e.g. we still want to consider the prediction: \\ $pred =$ \textit{'the World Health Organization'} $\Leftrightarrow$ $true = $ \textit{'World Health Organization'}, as partially accurate, rather than giving it a null score)

\hypertarget{evalit}{\subsubsection{Results: Model Evaluation}}
\State A hand-picked selection of $1000$ sentences created our evaluation set, a selection made from the publications that constituted the \textit{100 Publications} corpus. This additional step was taken to compare our models on real unseen data sourced from publications of our domain, ultimately reflecting a better evaluation that aligns with our task. A more thorough view of this set will be provided in the \hyperlink{prodigy}{\textit{data annotation}} part of the next chapter.

\vspace{0.1cm}

\State Let us present the evaluation through a detailed classification score report for each model. The report includes per-label evaluation metrics in addition to averages across all labels, giving us more in-depth insights into the general performance.

\begin{center}
    \includegraphics[scale = 0.65]{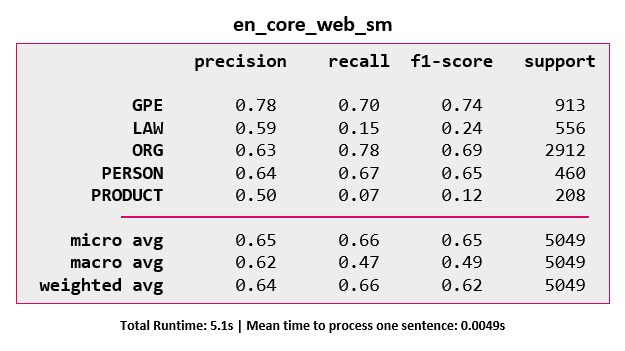}\\[1cm]
    \includegraphics[scale = 0.65]{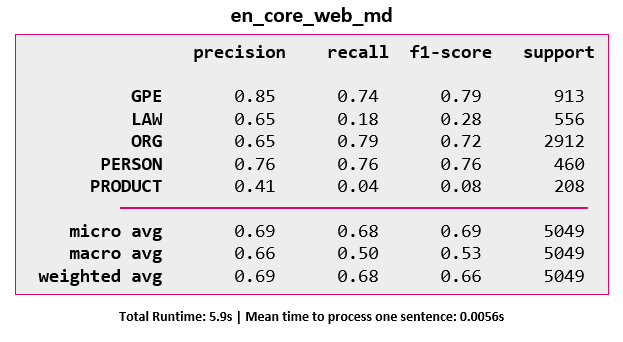}\\[1cm]
    \includegraphics[scale = 0.65]{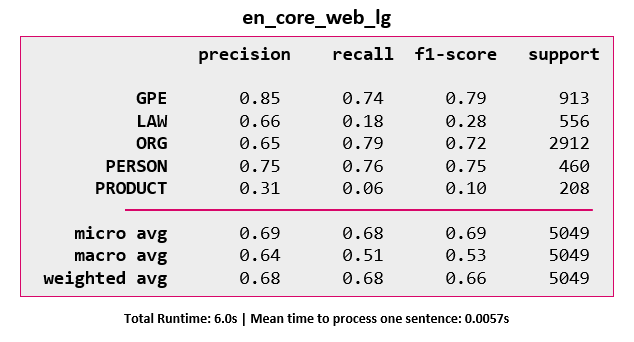}\\[1cm]
    \includegraphics[scale = 0.65]{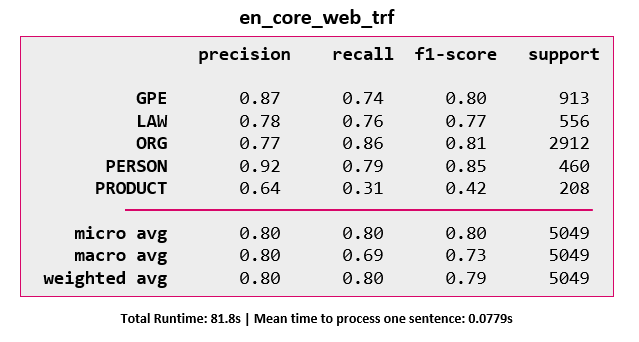}\\[1cm]
\end{center}

\State An increase in performance across all labels can be noticed after each selection of a larger model. Having said that, in order to provide a simpler comparison, the harmonized mean of micro average \hyperlink{fff}{F}-scores will be selected as the selected scoring value from now on.

\begin{center}
    $micro_F = \frac{2 \ (micro_r \ * \ micro_p)}{micro_r \ + \ micro_p}$ \ (13)
\end{center}

\State Being a metric that assigns equal weight to each instance of a prediction regardless of the label, micro averages also reflect our imbalanced label representation. However, the dataset was designed with the intention to reflect the real-world distribution of our labels, therefore $micro_F$ coheres.

\begin{center}
    \includegraphics[scale = 0.33]{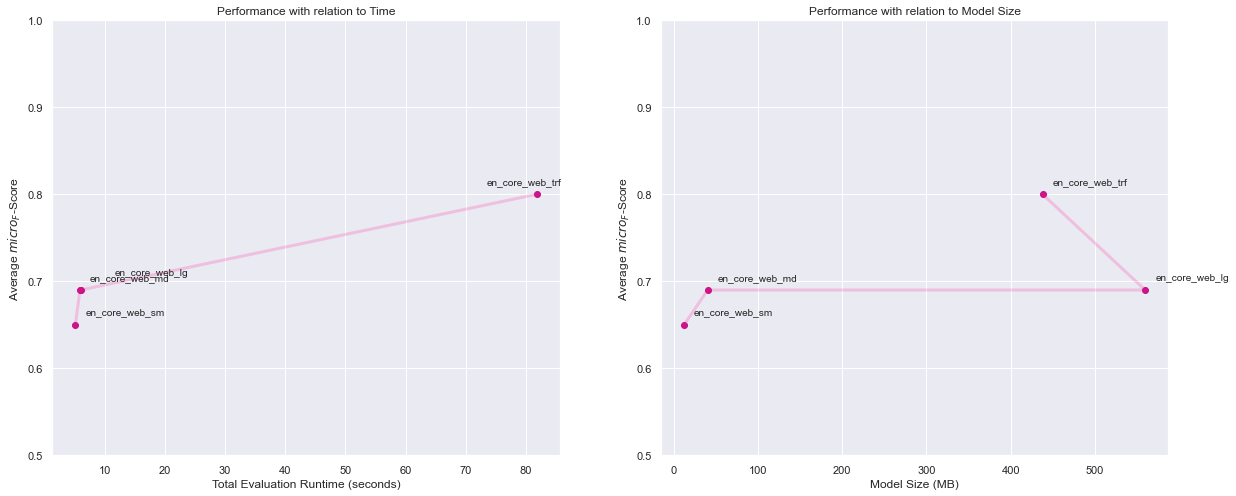}\\[0.5cm]
    \textit{Model Evaluation - Time and Size difference on performance}\\[0.5cm]
    \begin{tabular}{|c|c|c|c|m{2cm}|} 
        \hline
        Model & Size & Evaluation Runtime (s)\footnotemark & $\mu_{micro_F}$  \\[2ex]\hline
        \href{https://github.com/explosion/spacy-models/releases/tag/en_core_web_sm-3.5.0}{\textcolor{unige}{\textit{en\_core\_web\_sm}}} & 12 MB & 5.1007 & 0.65 \\\hline
        \href{https://github.com/explosion/spacy-models/releases/tag/en_core_web_md-3.5.0}{\textcolor{unige}{\textit{en\_core\_web\_md}}} & 40 MB & 5.8665 & 0.69 \\\hline
        \href{https://github.com/explosion/spacy-models/releases/tag/en_core_web_lg-3.5.0}{\textcolor{unige}{\textit{en\_core\_web\_lg}}} & 560 MB & 5.9935 & 0.69 \\\hline
        \href{https://github.com/explosion/spacy-models/releases/tag/en_core_web_lg-3.5.0}{\textcolor{unige}{\textit{en\_core\_web\_trf}}} & 438 MB & 81.7565 & 0.80 \\\hline
    \end{tabular}\\[0.4cm]
\end{center}

\footnotetext{\ Time to perform NER on the entire dataset of 1000 sentences.}

\vspace{0.2cm}

\State There are two principal remarks to make after observing the performance of these default pre-trained models. Firstly, the increase in model size is followed by an increase in performance, albeit not notable. All CNN models demonstrate an impressive speed by being able to process a single sentence in $\approx 5$ milliseconds ($ms$). Specifically, the medium CNN model \textit{en\_core\_web\_md} is distinguished, considering its performance despite the small size. \\
Regarding the transformer model, its contextual embeddings capabilities lead to a bump in performance, as per expectation. Having said that, their $11x$ slower processing time led to a decision to only proceed with the medium and large CNN models, hoping to increase their satisfactory default $~0.69$ performance score through fine-tuning.

\begin{center}
    \includegraphics[scale = 0.35]{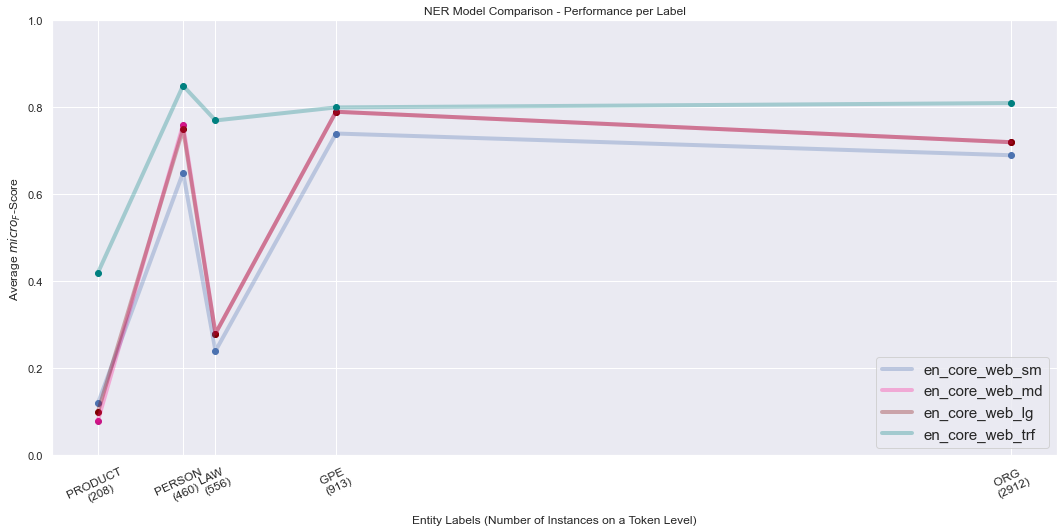}\\
    \textit{Performance Distribution per Entity Label}
\end{center}

\vspace{0.1cm}

\State We can observe the competent performance of all the models with regard to our most important entity labels: persons, organizations, and geopolitical entities. \\
Nevertheless, the average score is negatively affected by the poor performance in the recognition of laws and products, except for the transformer models which utilize their contextual understanding to achieve a better recognition. This is due to the low number of instances in addition to the complex nature of both entity types, given that laws may present themselves in different formats as well as the PRODUCT label not being clear if it's related to names of digital health marketable products. Having said that, it ultimately displays a need for fine-tuning.

\vspace{1cm}

\subsubsection{Results: The need for fine-tuning}
\State To have a better understanding of this need, we will dive deeper into instances of actual sentence predictions and the difference between entity predictions and true golden entities. A selection of sentences that encapsulate the shortcomings of the default models will be displayed to highlight specific areas in which we seek improvement. 

\begin{center}
    \includegraphics[scale = 0.5]{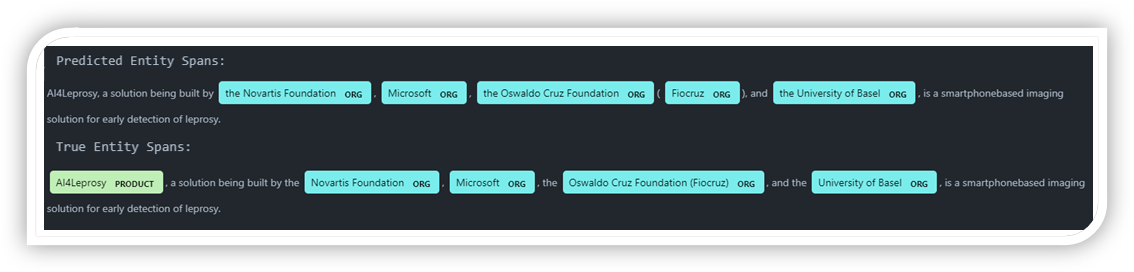}
\end{center}

\State A small noticed flaw observed during evaluation was the inclusion of determinants in the prediction of entity spans as well as incorrect predictions of nested organization entities when they are written in a '\textit{Organization Name (Acronym)}' format:

\vspace{0.1cm}

\begin{center}
\begin{itemize}
    \item $pred =$ \textit{'\textbf{the} University of Basel'} $\Leftrightarrow$ $true = $ \textit{'University of Basel'} \\[0.3cm]
    \item    $preds =$ [\textit{'the Oswaldo Cruz Foundation'}, \textit{'Fiocruz'}] $\Leftrightarrow$ $true = $ \textit{'Oswaldo Cruz Foundation (Fiocruz)'} \\[0.5cm]

\end{itemize}
   \includegraphics[scale = 0.4]{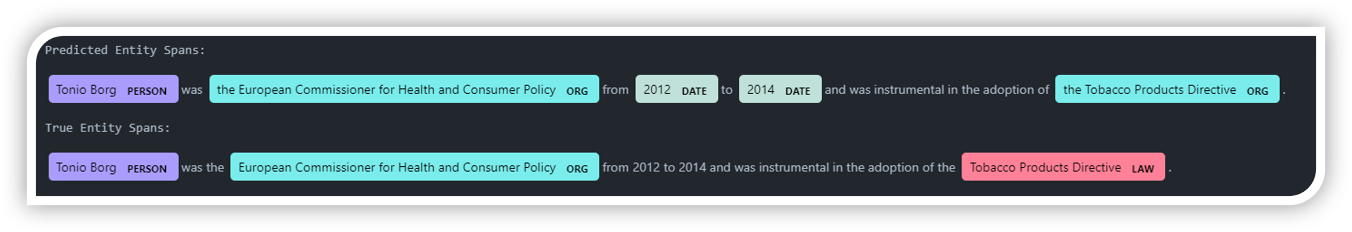}\\[0.2cm]
   \includegraphics[scale = 0.4]{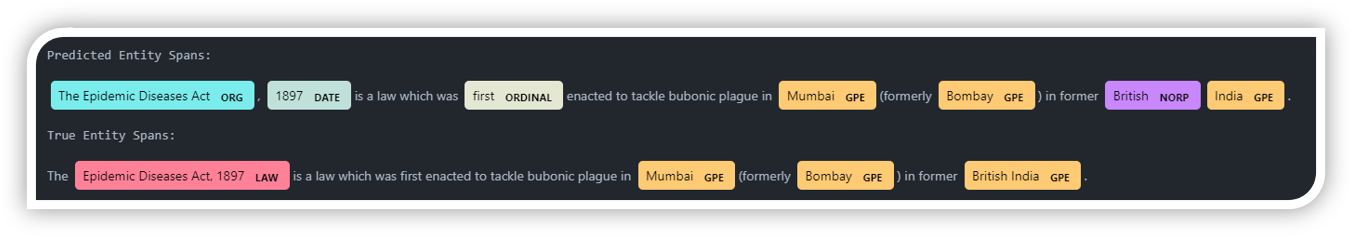}
\end{center}

\State As for the LAW entity label, the majority of false positives can be attributed to the model incorrectly predicting them as organizations instead. This demonstrates the default models' unfamiliarity with law formats.

\begin{center}
    \begin{itemize}
        \item $pred =$ \textit{'the Tobacco Products Directive'} (ORG) $\Leftrightarrow$ $true = $ \textit{'Tobacco Products Directive'} (LAW) \\[0.3cm]
        \item $pred =$ \textit{'the Epidemics Disease Act'} (ORG) $\Leftrightarrow$ $true = $ \textit{'Epidemics Disease Act 1897'} (LAW)
    \end{itemize}
\end{center}

\State For the most part, the PERSON labels' score was negatively affected by the inability of the models to predict the full nested entities as I intended on the evaluation set. Specifically, nested entities that included the full name of an individual, in addition to their professional title.

\begin{center}
    \includegraphics[scale = 0.4]{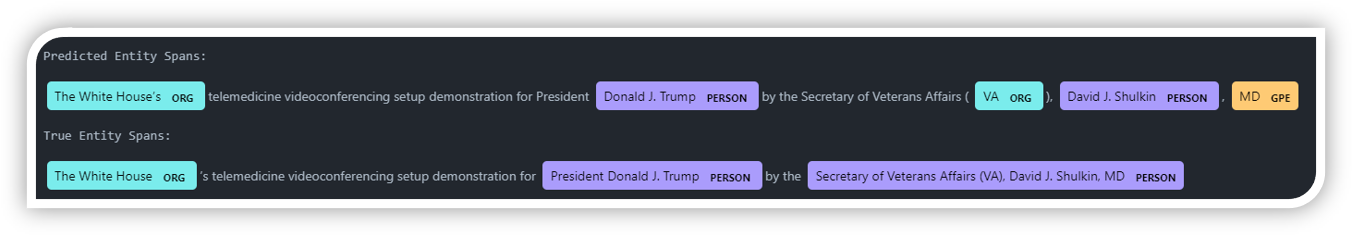}\\[0.2cm]
    \includegraphics[scale = 0.49]{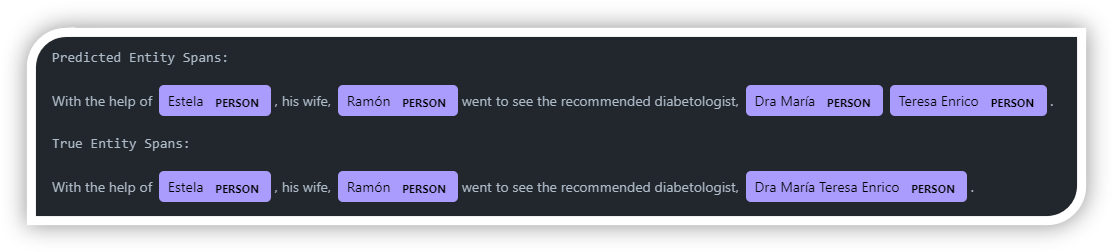}\\[0.2cm]
    \includegraphics[scale = 0.4]{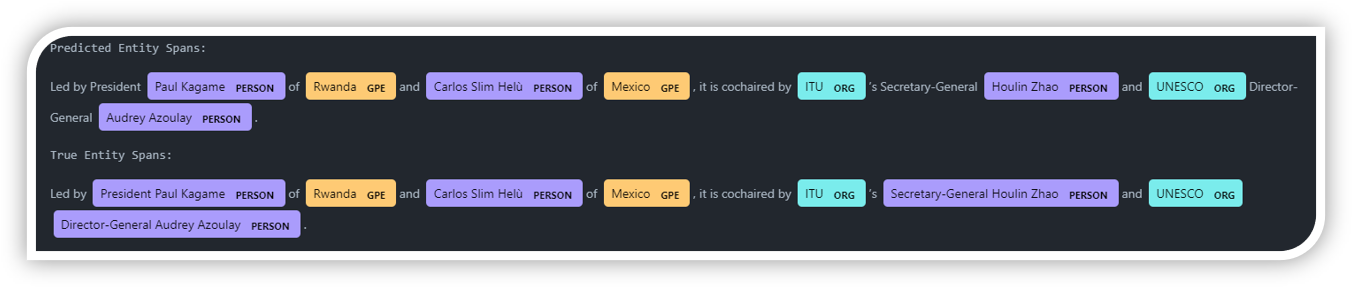}\\[0.5cm]

    \begin{itemize}
        \item $pred =$ \textit{'Donald J. Trump'} $\Leftrightarrow$ $true = $ \textit{'President Donald J. Trump'} \\[0.5cm]
        \item $preds =$ [\textit{'Dra. Maria'}, \textit{'Teresa Enrico'}] $\Leftrightarrow$ $true = $ \textit{'Dra. Maria Teresa Enrico'} \\[0.5cm]
        \item $pred =$ \textit{'Houlin Zhao'} $\Leftrightarrow$ $true = $ \textit{'Secretary-General Houlin Zhao'} \\[0.5cm]
    \end{itemize}
\end{center}

\State Recognizing instances of our domain's marketable products such as healthcare technologies, telehealth, medical devices, and more, to the PRODUCT label was the main difficulty of all models. This difficulty was the main contribution to a limitation in average score across all labels. 

\begin{center}
    \includegraphics[scale = 0.4]{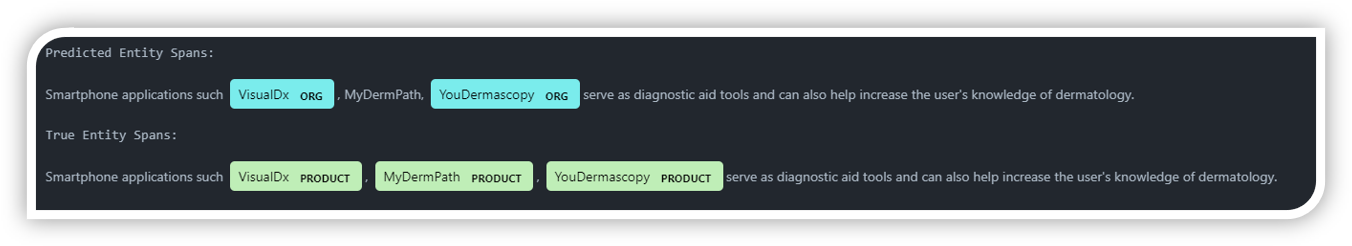}\\[0.2cm]
    \includegraphics[scale = 0.415]{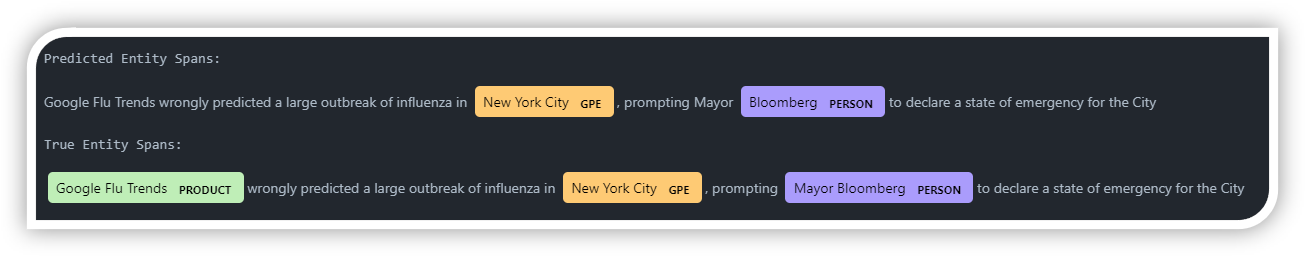}\\[0.2cm]
\end{center}

\State Other sentences to test the models' abilities were adversarial examples crafted to identify potential weaknesses. These sentences contain words that do not represent entities but appear in a similar context to entities, crafted with the intention of tricking the model. 

\begin{center}
    \includegraphics[scale = 0.5]{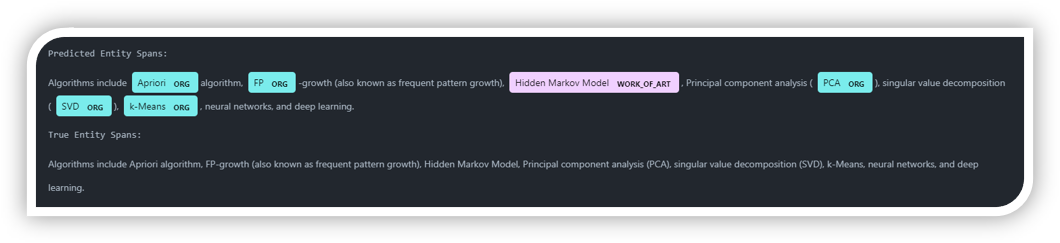}\\[0.2cm]
    \includegraphics[scale = 0.43]{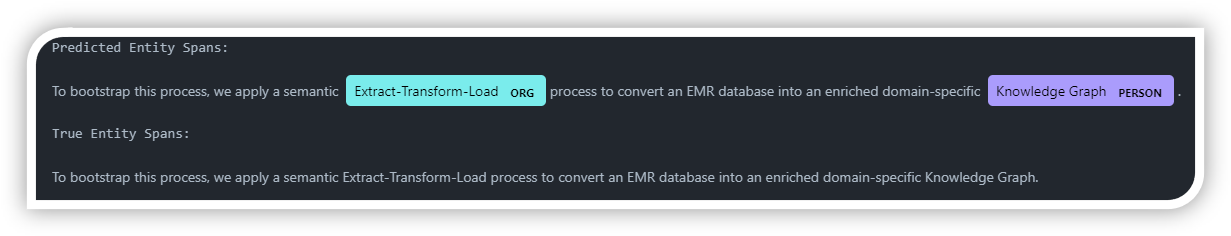}\\[0.2cm]
\end{center}

\State After being presented with these examples, one can understand the need for fine-tuning our models need, to better handle sentences derived from global health publications.

\newpage

\subsection{Fine-Tuning a model}
\State Fine-tuning is the process of taking the representations from our pretrained models, and further training them to be better adapted to the downstream task of named entity recognition to our labeled dataset. Typically, this training will make only minimal adjustments to the pretrained language model parameters [\hyperlink{finetuning}{44}].  

\State This concept stands as one of the capabilities of \textit{spaCy} models. They have the possibility to adjust their word vectors through supervised further-training on a given dataset. In our case, the presumed result will be a fine-tuned named entity recognition process, finely tuned to align more accurately with the entities annotated in our training dataset.

\vspace{0.3cm}

\State A mock example to demonstrate this property is a comparision between a default CNN model and a fine-tuned one fed with multiple sentences that included \textit{GDPR (LAW)} and \textit{United States (GPE)} entities, the results of which are displayed below:

\begin{center}
    \includegraphics[scale = 0.45]{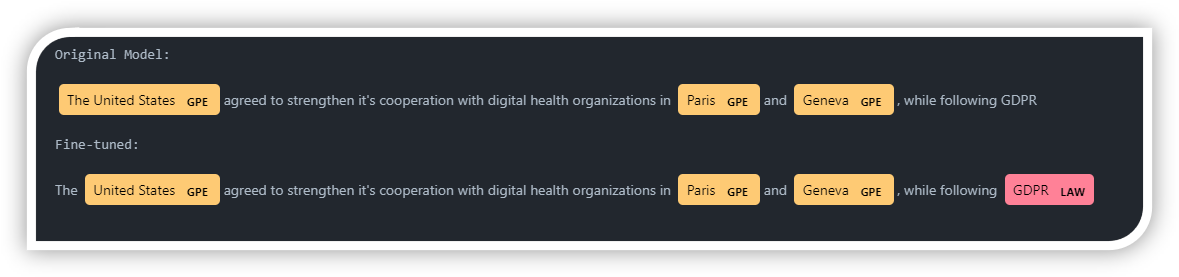}\\
    \textit{Fine-Tuning Results Example}    
\end{center}

\State As we can observe, the model learned to recognize GDPR as a law entity type, in addition to recognizing the geopolitical nested entity without the inclusion of the determinant.

\vspace{0.2cm}

\State One of the most important problems to keep in mind when performing fine-tuning is the '\textit{catastrophic forgetting}' phenomenon. It refers to the tendency that pretrained models have to forget previously learned representations when they are presented with new data during fine-tuning and is especially lethal when the fine-tuning dataset is relatively small, such as in our case. \\
A carefully curated annotated dataset in addition to an optimized selection of hyper-parameters is crucial to combat this phenomenon.

\hypertarget{prodigy}{\subsubsection{Data Annotation with Prodigy}}
\State The majority of traditional natural language processing tasks such as named entity recognition are not routine tasks, and therefore rely on supervised learning through an annotated corpus. To this day, human annotation still stands as one of the slowest and expensive tasks of supervised learning projects [\hyperlink{annot}{45}].

\State Hence why data annotation tools that facilitate the process have been indispensable in the industry and research. One such example is \href{https://prodi.gy/}{\textcolor{unige}{\textit{\underline{Prodigy}}}} \footnote{\ https://prodi.gy/}, a tool developed by the \textit{spaCy} team designed to streamline the data creation process in the field of natural language processing. Taking as input a set of raw sentences, it provides a user-friendly web interface where you can annotate each token to its corresponding named entity label by simply clicking on it. The result is a fully annotated dataset that is formatted according to the expectations of the models during training, making fine-tuning seamless.

\vspace{0.2cm}

\State Another key property of Prodigy related to active learning during annotation, is its ability to provide real-time predictions from the pretrained models. This helps the annotator debug the model's error in the case of misaligned predictions, in addition to accelerating the whole process in the case of correct predictions.

\begin{center}
    \includegraphics[scale = 0.6]{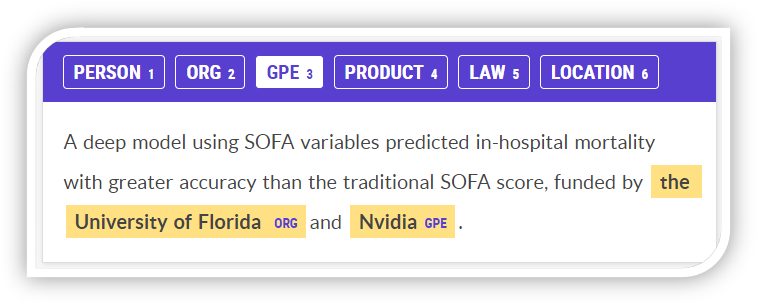}\\
    \textit{Prodigy - Active Learning}
\end{center}

\State Above is displayed an example where the model suggested a correct entity prediction while also providing a false positive (Nvidia as a Geopolitical Entity), needing to be fixed.

\begin{center}
    \includegraphics[scale = 0.5]{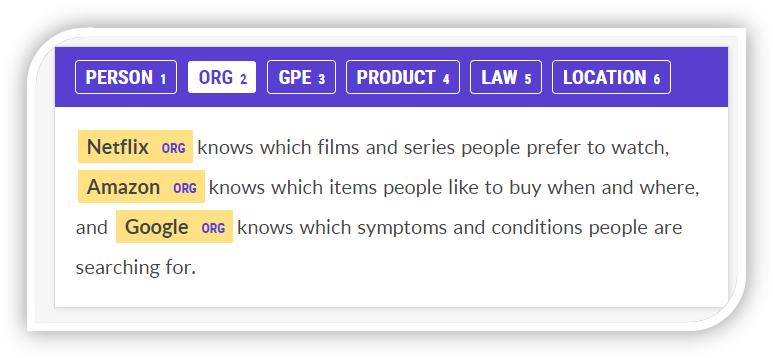}\\
    \includegraphics[scale = 0.5]{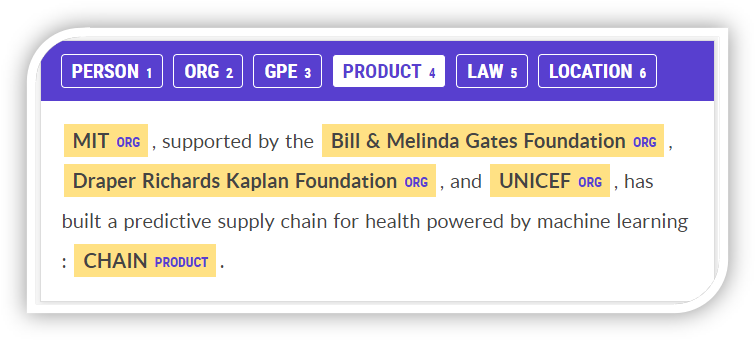}\\
\end{center}

\State The evaluation set on which the default models were compared was created in this manner. We can observe the dominance of organization entity types, and that reflects the reality given that these sentences were extracted from actual publications (\textit{100 Publications Corpus}).

\begin{center}
    \includegraphics[scale = 0.5]{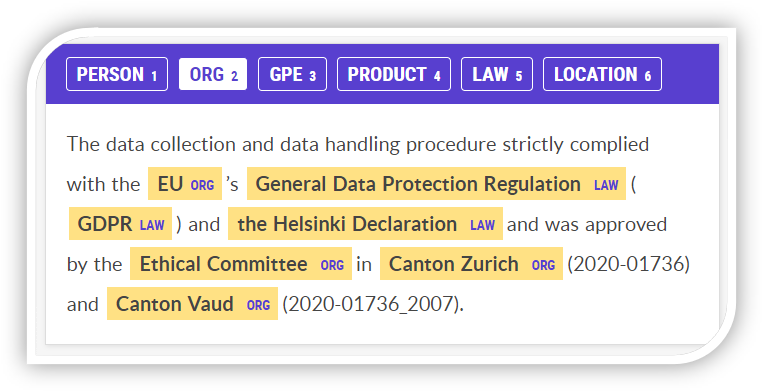}\\
\end{center}

\State The correct full prediction of lengthy nested entities that usually represent a law or organization entity type, is one of the challenges presented by this dataset.

\begin{center}
    \includegraphics[scale = 0.5]{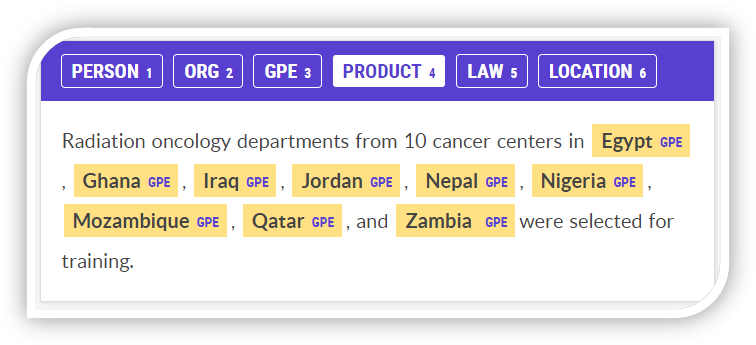}\\
    \includegraphics[scale = 0.5]{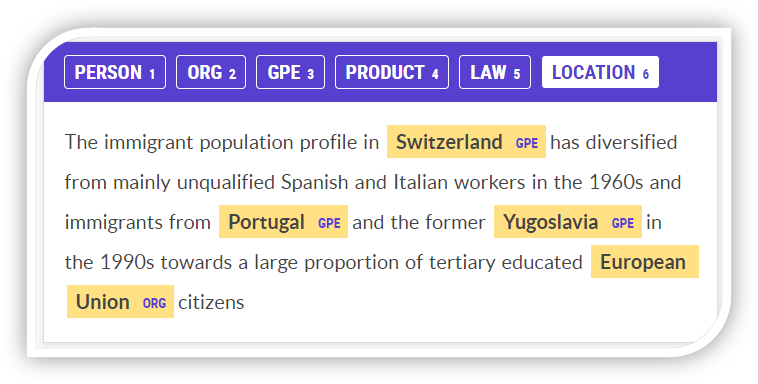}\\
\end{center}

\State One can notice that the context of the sentence is almost always related to the global digital health domain. \\
In the case of geopolitical entity (GPE) recognition, models tended to be fairly accurate knowing that names of countries are not changing anytime soon. Therefore, the annotation focus shifted towards fine-tuning the models to recognize more vast entities such as: continents, regions, LMICs, HICs, and more, to the GPE label.

\begin{center}
    \includegraphics[scale = 0.5]{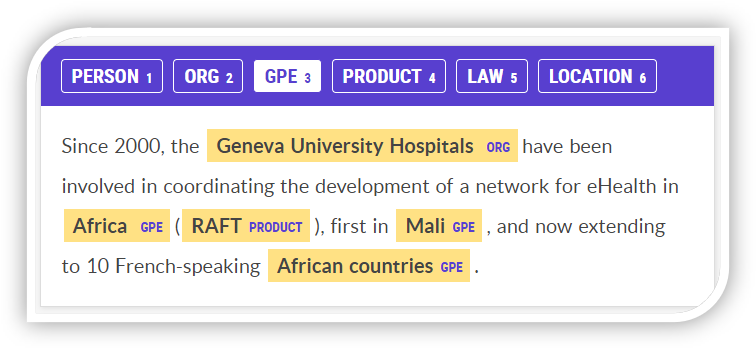}\\
\end{center}

\begin{center}
    \includegraphics[scale = 0.5]{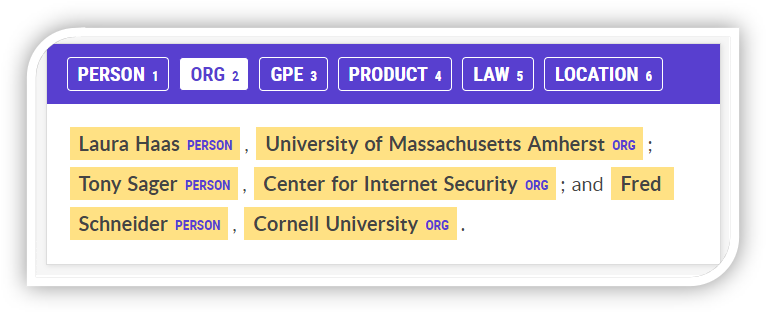}\\
    \includegraphics[scale = 0.5]{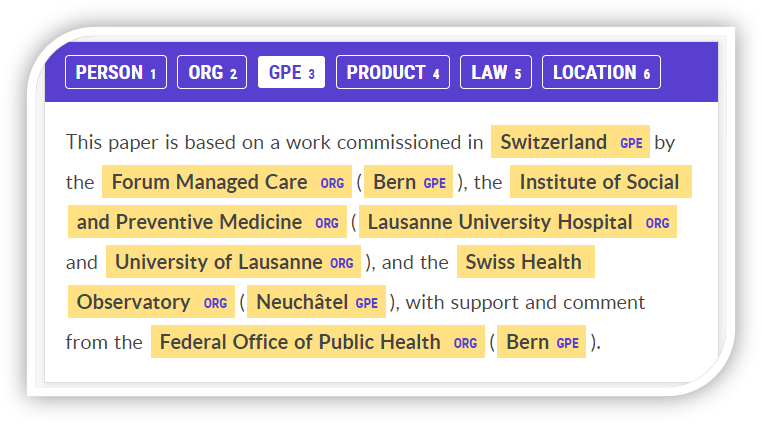}\\
\end{center}

\State Other challenging to predict instances were created through sentences that contain more named entities than other words, or even sentences that are composed solely of named entities. Inclusion of multiple entities of different labels in a single sentence is important in order to provide a balanced distribution while the inclusion of common easy-to-recognize entities combats the catastrophic forgetting problem.

\vspace{0.2cm}

\begin{center}
    \includegraphics[scale = 0.5]{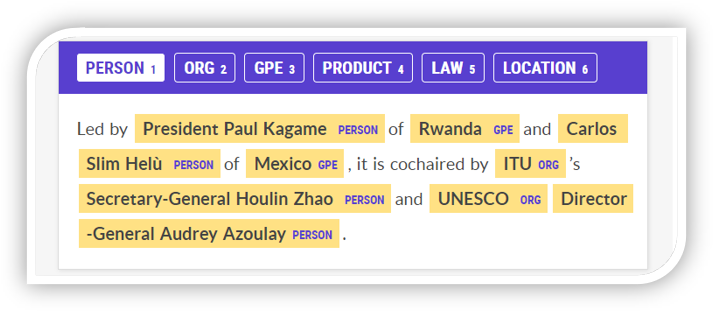}\\
\end{center}

\State Inclusion of professional titles to the PERSON nested entity was another priority of this fine-tuning.

\begin{center}
    \includegraphics[scale = 0.5]{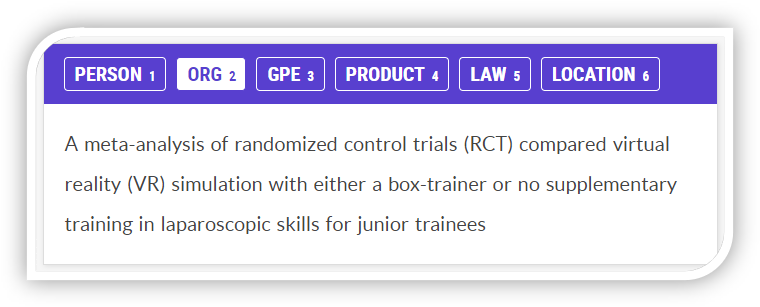}\\
    \includegraphics[scale = 0.5]{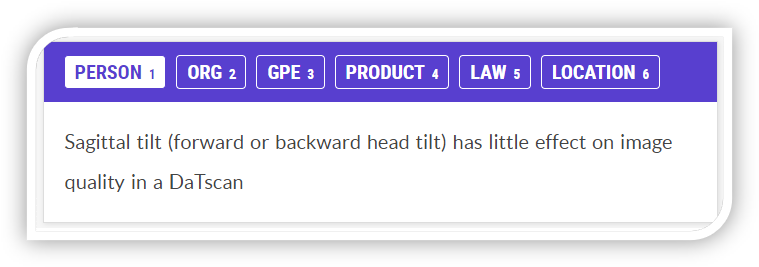}\\
\end{center}

\State Adversarial examples are passed as sentences that do not contain any entity. Apart from being significantly challenging to predict without having the proper context, these types of sentences also fine-tune the model. Specifically, training it to disregard any word included in them in relation to NER.

\vspace{1cm}

\State To conclude it, the annotations were produced with the goal of challenging NER models while also fine-tuning them to our specific task. While some ground truth labels are definitive, other annotations are subjective and depend on the intuition of annotators (e.g. tagging LMICs as GPE). Such cases were consulted with my supervisor and other global health experts, before arriving to the final annotation. \\
Annotation protocols should satisfy several criteria: \textit{expressive} enough to capture the phenomenon of interest, ability to be \textit{replicable} meaning that other annotators would produce a very similar annotation, and the ability to be \textit{scalable}, meaning they can be produced relatively fast [\hyperlink{critme}{46}]. 
I made every effort to create this dataset while staying faithful to these protocols.

\subsubsection{Training}
\State After the dataset was created, a 70\% Train / 30\% Test split was performed, in order to proceed with the fine-tuning. 

\State In spaCy, fine-tuning is carried through the command line, where the default model, train data, and a configuration file of training hyperparameters are passed as input. After the results of our \hyperlink{evalit}{\textit{model evaluation}} chapter, it was decided to move forward with only the medium and large versions of the CNN models: \textit{en\_core\_web\_md} and \textit{en\_core\_web\_lg}. The transformer model was abandoned for our document processing task due to the slower processing times (11x slower than CNNs), in addition to the high computational complexity it requires in order to be fine-tuned.

\State Therefore, the [\textit{tok2vec, ner}] pipelines of both models would be fine-tuned as according to the configuration file. The tokenizer, convolutional neural network, and dependency parsing architectures remained unchanged from the default ones, after experimenting.

\State Apart from that, our hyperparameters related to the training in itself were: number of epochs $e = 100$, batch size $b = 200$ to boost computational efficiency, dropout $p_D = 0.1$ and L2 Norm  = $0.01$ for regularization, and Adam V1 with a learning rate of $\eta = 0.01$ as one of the most efficient optimizers in stochastic gradient descent problems [\hyperlink{adam}{47}].

\begin{center}
    \includegraphics[scale = 0.51]{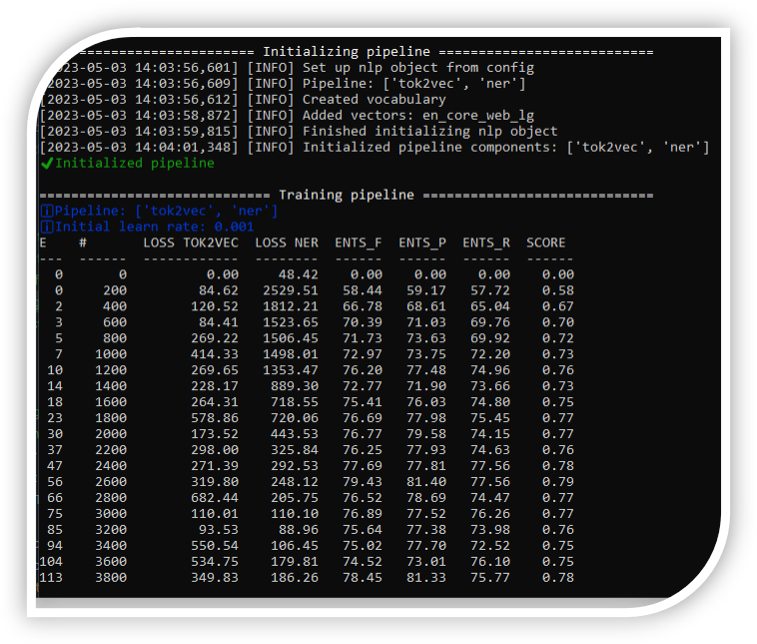}\\
    \textit{'spacy train' - Fine-Tuning Command Line}
\end{center}

\State After multiple runs with different configurations, the one displayed above resulting in $\approx 0.80$ average score was the most stable. The score is computed using it's strict version, which requires index-perfect entity prediction to count it as accurate. Therefore, the fine-tuning seemed promising knowing that the prior default model evaluation was computed through the more lenient scoring function.

\subsubsection{Conclusion}
\State Upon concluding the training, the more lenient scoring function was performed in order to obtain a classification score as to the prior evaluation. As during the interim score shown during training, the improvement in comparison to the default models was noticeable. Given that fine-tuning merely adjusts the original vectors without modifying their size, the processing speed per-sentence was still decent.

\begin{center}
    \includegraphics[scale = 0.44]{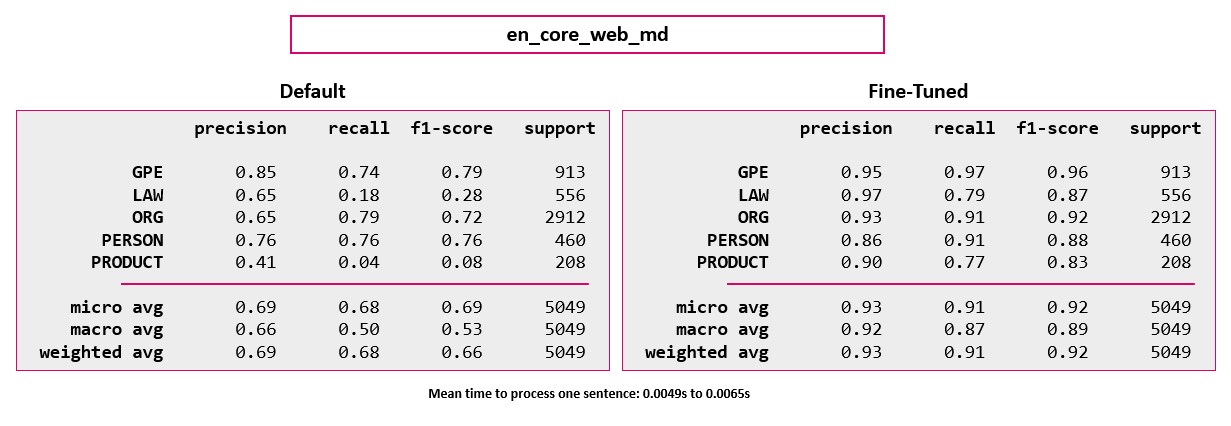}\\[0.3cm]
    \includegraphics[scale = 0.44]{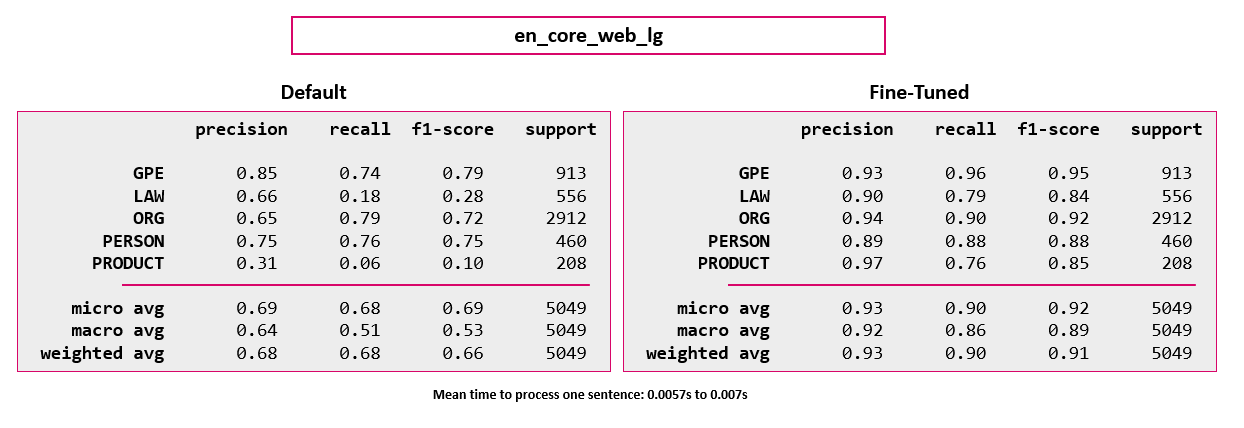}\\
\end{center}

\begin{center}
    \includegraphics[scale = 0.35]{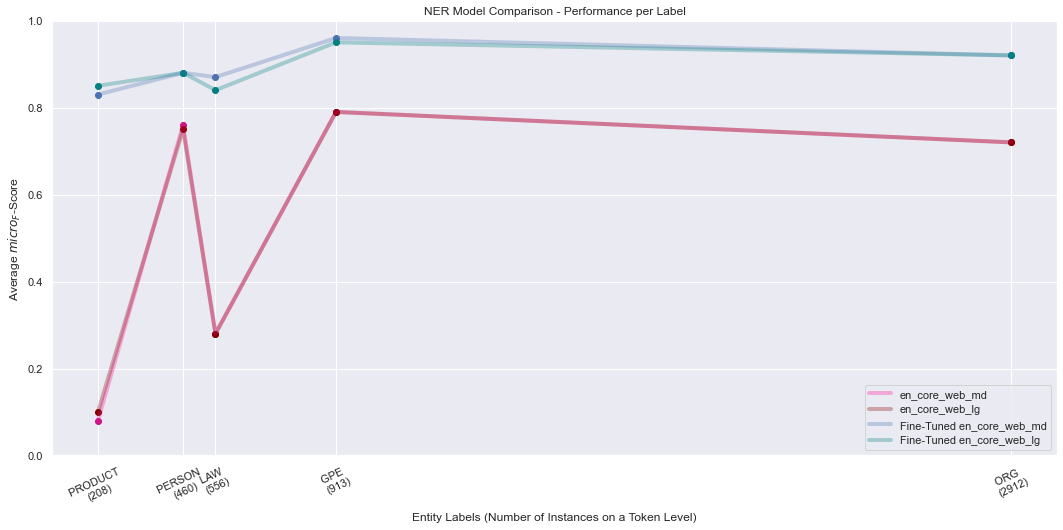}\\[0.3cm]    
    \textit{Performance Distribution per Entity Label - After Fine-Tuning}
\end{center}

\vspace{0.2cm}

\State The level of improvement in terms of performance for the fine-tuned model is easily distinguishable. For the majority of the labels, this is the result of the better adaptation to recognizing nested named entities as we intended during annotation. In addition to that, the recognition of new entities based on our labeling that the pre-trained models were previously unable to identify (e.g. LMICs, HICs, and Continents as GPE), enhanced the performance further.

\State Regarding the LAW label, fine-tuning with clear instances of laws that follow a certain format (in our case mostly American or Swiss Laws, written as "The Act of Something") resulted in a bump in performance regarding their recognition. Having said that, this is not expected to generalize well with the diverse range of laws worldwide, knowing that each legislative power institution writes laws in their own unique way.

\State Therefore, the cases where the difference in performance between the default and fine-tuned models is significant should still be viewed with caution. The limited size of the training set in addition to a high number of epochs might be a sign of overfitting, meaning that the model does not generalize well on unseen data.

\State After further benchmarking, it was found that recognition of LAWs still generalizes well while the PRODUCT label remains the least stable. This is due to the fact that the original model wasn't trained with global health products in mind, hence it was dropped as a label moving forward.

\subsection{Custom Label Training - Disease Recognition}
\State One of the advantages of using spaCy's pipelines for named entity recognition is their customizability, the fact that you are not limited to solely using the pretrained models. You are also able to train a blank NER model from scratch with your own training examples, adapting it to custom entity types of one's application requirements [\hyperlink{customit}{48}]. Essentially, as long as one has a trusted annotated dataset at disposal, one can train a NER model able to recognize any custom label entity out of it.

\State For the most part, one rarely trains a blank model from scratch as pretrained models are preferred for general named entity recognition tasks given their performance and reliability. Another constraint is the scarceness of available annotated NER data released for the public. However, when trusted data is available and the required entity type for the application is one of a unique nature, custom training is necessary.

\State Such was our case, where a DISEASE label was required in order to extract the main types of diseases mentioned in a global health publication.

\subsubsection{NCBI Disease Corpus}
\State Our chosen dataset is the public release of the \textit{NCBI Disease Corpus}, a corpus which contains a total of 6892 disease mentions, mapped to 790 unique disease concepts [\hyperlink{ncbi}{49}]. Out of these, $88\%$ are linked to a Medical Subject Heading (MeSh) identifier, while the rest to a Online Mendelian Inheritance in Man (OMIM) one [\hyperlink{ncbi}{49}].

\State The annotation process of this corpus enhances its credibility by involving 14 biomedical experts as annotators, paired randomly to avoid bias, in order to go through three phases of annotation by following certain guidelines [\hyperlink{ncbi}{49}]. Knowing the complex nature of disease names due to their variation and ambiguity, in addition to the numerous unique instances, choosing a community-trusted dataset such as this one emerged as the most favourable option.

\State In addition to that, another reason for choosing this corpus is that is already pre-split into a train, validation, and evaluation set, from its publication. This makes it ready to be integrated into the training process and eliminates any possibility that a random split from my side might not be optimal.


\subsubsection{Performance}
\State The dataset was downloaded from its instance stored in \href{https://huggingface.co/datasets/ncbi_disease}{\textcolor{unige}{\textit{\underline{HuggingFace}}}} \footnote{\ https://huggingface.co/datasets/ncbi\_disease}, a leading platform of natural language processing datasets and models. It contains 5433 train ($74\%$) instances, 924 validation ($13\%$), and 941 test ($13\%$) ones. Each instance is a tokenized sentence encoded as an array of tags,  where 0 indicates no disease mentioned, 1 signals the first token of a disease and 2 the subsequent disease tokens.

\State The next step to adapt the data into spaCy training pipeline was converting it from the default tag array format to Doc objects. This allows the training to start, one which will convert a blank model with an empty pipeline to a disease recognition one, by adapting its pretrained word vectors to our task. The training configuration file was kept identical to the one used to fine-tune the CNNs, given that its hyperparameters were established as most optimal.

\begin{center}
    \includegraphics[scale = 0.5]{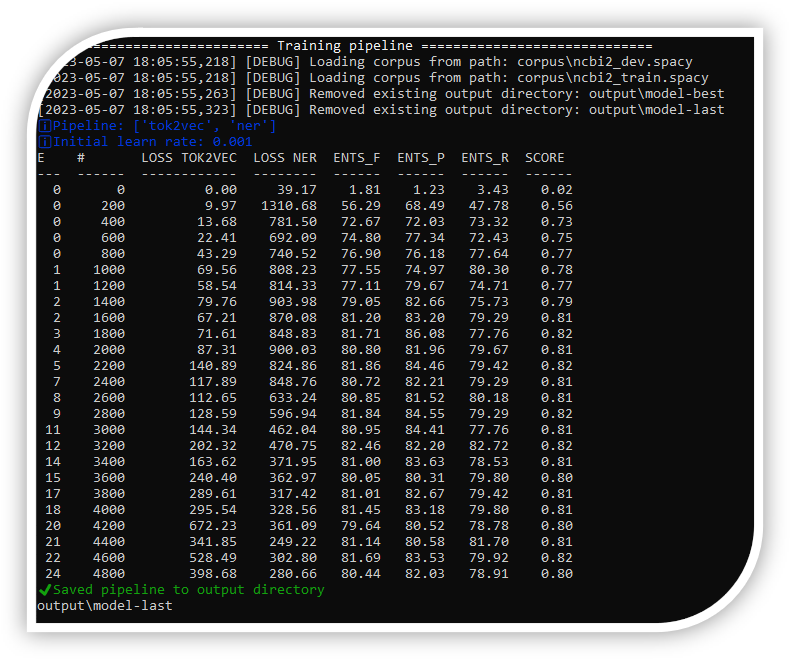}\\
    \textit{Disease Recognition - Model Training}\\[0.5cm]
    \includegraphics[scale = 0.5]{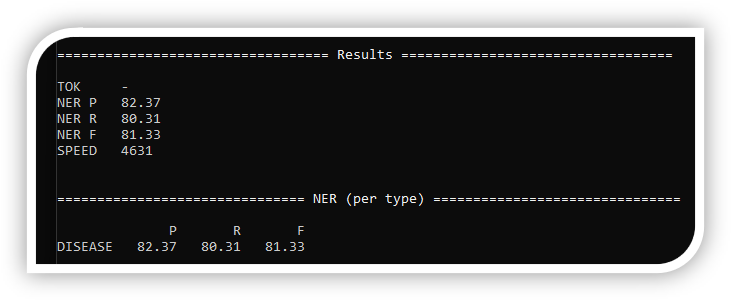}\\
    \textit{Disease Recognition - Test Set Evaluation} \footnote{\ P: Precision, R: Recall, F: F-Score}
\end{center}

\State After 25 epochs, the training resulted in a $\approx 81\%$ disease recognition CNN model, able to maintain the speed identified during the default model evaluation. Once again, this shows the customizability of spaCy's pipeline and named entity recognition as a task in itself. As long as you have a reliable annotated dataset, you can tailor NER to suit any custom labels for a specific domain.

\begin{center}
    \includegraphics[scale = 0.5]{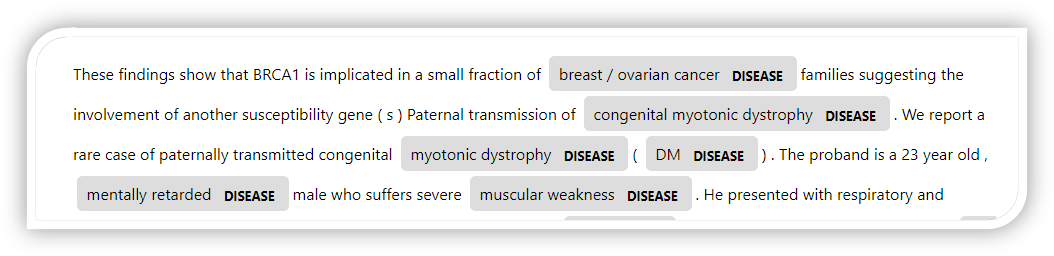}\\
    \textit{Disease Recognition - Test Set Prediction Examples I}\\[1cm]
    \includegraphics[scale = 0.51]{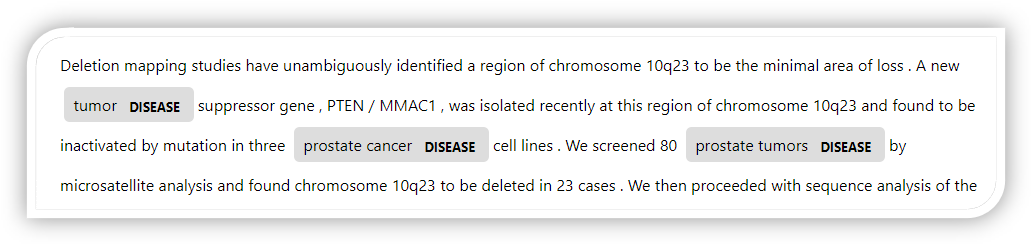}\\
    \textit{Disease Recognition - Test Set Prediction Examples II}
\end{center}

\hypertarget{finalmodel}{\subsubsection{The Complete Model}}
\State Having successfully established a stable disease recognition model, the next step was to integrate it with the rest of the fine-tuned model in order to create our final optimized named entity recognition pipeline.                                                                                                                                       
\begin{center}
\begin{lstlisting}[language=Python, escapeinside={(*@}{@*)}]
import spacy

### Load Both Optimized Models
fine_tuned_model = (*@\textcolor{unige}{spacy}@*).load("disk-path-to-model")
diseases_model = (*@\textcolor{unige}{spacy}@*).load("disk-path-to-model")
diseases_model.replace_listeners("tok2vec", "ner", ["model.tok2vec"])

### Add Disease Recognition to the fine-tuned model's pipeline
fine_tuned_model.(*@\textcolor{unige}{add\_pipe}@*)(
     "ner",
     name = "ner_diseases",
     source = diseases_model,
     after = "ner"
)
\end{lstlisting}
\textit{Merging the two models in Python}\\
\end{center}

\State The disease recognition pipeline was integrated after the default NER pipeline of the fine-tuned model, meaning that the DISEASE label is the last one predicted. It is a decision made to avoid any confusion, knowing that the two models are trained on datasets of different natures.

\State Having said that, this merge established the full model able to annotate all intended entity types in a sentence. Although not perfect in terms of accuracy, it maintained the speed of the CNN architecture, making predictions in milliseconds per sentence.

\begin{center}
    \includegraphics[scale = 0.5]{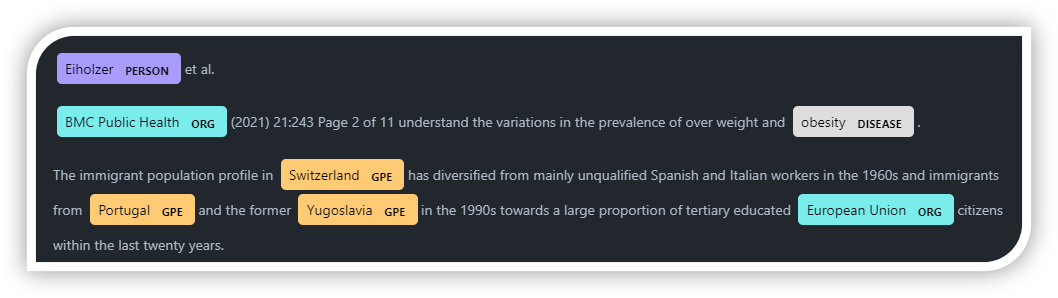}\\
    \textit{Full Model - Prediction Example I}\\[0.5cm]
    \includegraphics[scale = 0.5]{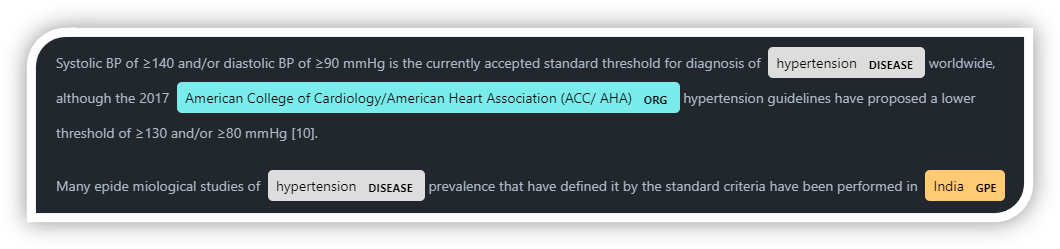}\\
    \hypertarget{fm2}{\textit{Full Model - Prediction Example II}}\\[0.5cm]    
    \includegraphics[scale = 0.47]{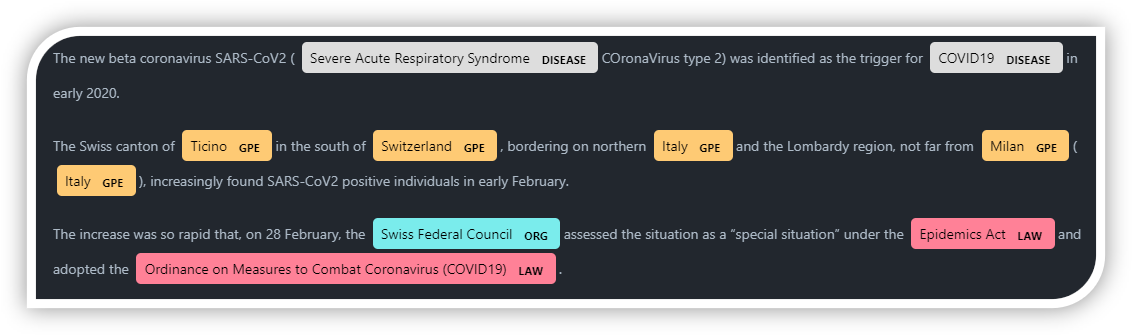}\\
    \textit{Full Model - Prediction Example III}   
\end{center}

\State Following the model evaluation, fine-tuning to our entity annotations, and establishing disease recognition, we are equipped with a ready-to-use model that could start annotating entire documents. Therefore, apart from the theoretical evaluation, we were one step closer to building a pipeline with regard to the industry need.

\newpage

\hypertarget{dpp}{\subsection{Document Processing Pipeline}}

\State Revisiting our original objective regarding the industry need, our aim is to proceed a from a PDF file containing a global health publication into a final list of main entities it encompasses. This was developed with the goal of automatically populating a knowledge base when only given a list of publications.

\vspace{0.2cm}

\begin{center}
    \includegraphics[scale = 0.55]{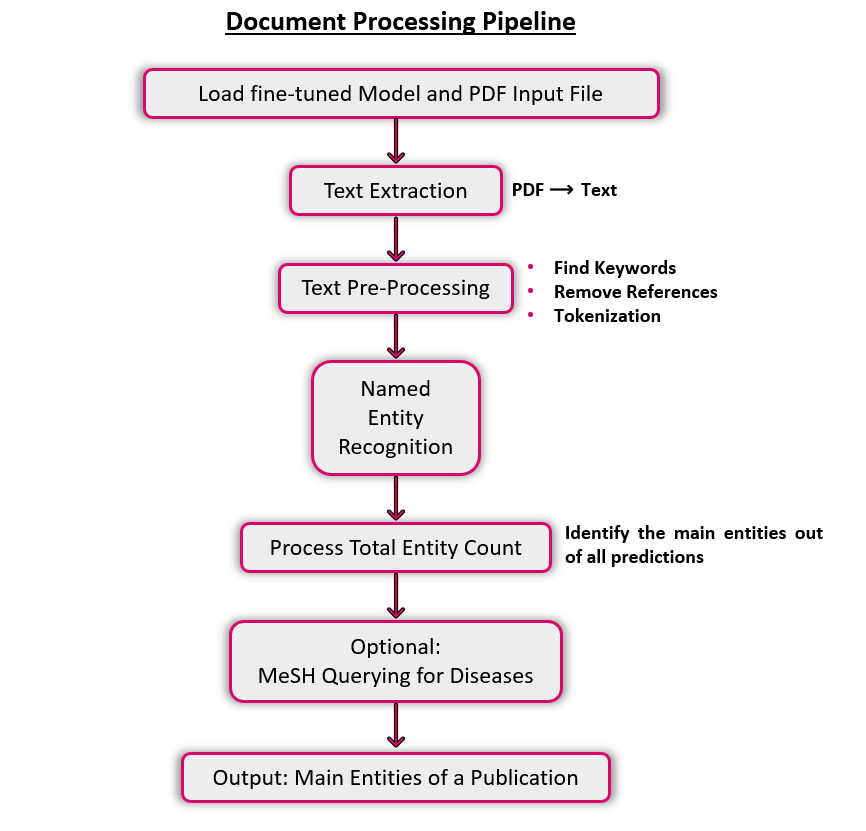}\\
\end{center}

\State The application requires only two disk paths prerequisites: the one to the PDF file of the publication and the one of the \hyperlink{finalmodel}{final} fine-tuned model. The entirety of the pipeline was developed as a Python script and was separated from the rest of the thesis code. This was a design choice made to think of it as a deployable application released on its own, or one that could be integrated into an already existing application that is in need of this task.

\subsubsection{Text Extraction}
\State For the first part that requires the entirety of the text to be extracted from the PDF, two text extraction tools adapted for Python were used and compared, \href{https://pypdf2.readthedocs.io/en/3.0.0/}{\textcolor{unige}{\textit{\underline{PyPDF2}}}} and \href{https://pypi.org/project/pdfminer/}{\textcolor{unige}{\textit{\underline{pdfminer}}}}. They extract text contents from the internal structure of a PDF, therefore were selected as a generally faster alternative than Optical Character Recognition (OCR) methods.

\State We were aiming for the most accurate extraction results, wanting the full extraction of the publication without any misses. This approach aimed to maximize the number of extracted entities while minimizing grammatical errors, thus avoiding creating any potential confusion for the model.

\State After benchmarking on one of our corpora, \href{https://pypi.org/project/pdfminer/}{\textcolor{unige}{\textit{\underline{pdfminer}}}} resulted in the more cleaner and accurate extraction of the information. Another positive remark was its capability to capture the text highlighted in complex elements such as tables, graphs, and other visually engaging content, therefore it was selected as the library.

\subsubsection{Text Pre-Processing}

\State The results of the text extraction process are presented in a string format, meaning the entirety of the text found in a publication is stored in a single string. On the other hand, our NER model expects a list of sentences in order to predict entities in each sentence, therefore a pre-processing step of sentence tokenization is required.

\State Before the tokenization process, Python's \textit{string.find(substring)} method that finds the first occurrence of a certain substring was exploited for two simple yet important tasks for our project, finding and extracting keywords and references in the text of a publication.

\State \textit{Keywords} of a publication contain a selection of words that best summarizes the topic of a document. They were chosen to be extracted to be compared with our final list of main extracted entities in terms of granularity.

\State \textit{References} that are usually listed on the final pages of a publication contain types of entities such as names of people, organizations, universities, etc. Therefore the goal of their extraction was to delete them from the string as they would serve no purpose to the named entity recognition task. On the contrary, passing them through the NER pipeline creates confusion given that the entities located in the references would be counted in the final results.

\begin{center}
    \includegraphics[scale = 0.5]{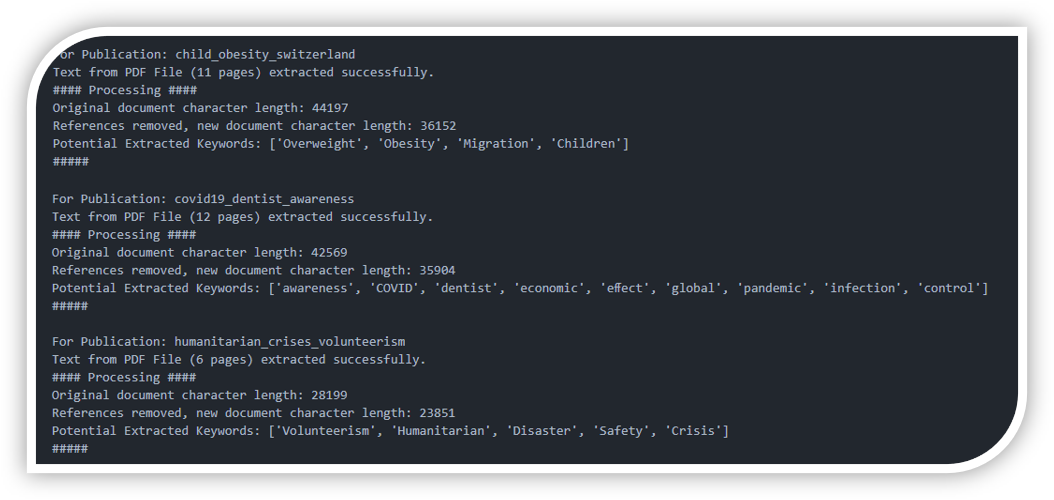}\\
    \textit{Text Pre-Processing - Extracting Keywords \& References}\\[1cm]
\end{center}

\State As observed through the sample of results displayed above, both methods are capable of successfully extracting both envisioned sections of a publication. Considering the efficiency of all string manipulation methods, the extraction process is conducted in $\approx 2ms$, hereby showing the motive to include this method in the document processing pipeline. Even in cases where neither references or keywords are extracted due to a nature of a publication of the failure of the finding method, the total runtime of the named entity recognition pipelined is hardly affected. 

\vspace{0.2cm}

\begin{center}
    \underline{\textit{Tokenizer}}
\end{center}

\State Tokenizing the document text string into a list of sentences stands as the final part of the pre-processing step, the format expected by the fine-tuned NER model. After a comparison of tokenizer offered by the two popular NLP libraries, \textit{spaCy} and \textit{nltk (Natural Language Toolkit)}, \href{https://www.nltk.org/api/nltk.tokenize.sent_tokenize.html}{\textcolor{unige}{\textit{\underline{nltk's sentence tokenizer}}}} was selected as a faster and more reliable solution. A solution in the form of a pre-trained Punkt english tokenizer that uses an unsupervised algorithm to build a model for abbreviation words, collocations, and words that start sentences, trained on a large collection of plaintext \footnote{\ \href{https://www.nltk.org/api/nltk.tokenize.punkt.html}{\textcolor{unige}{\textit{\underline{nltk - Punkt Sentence Tokenizer}}}}}.

\begin{center}
    \includegraphics[scale = 0.6]{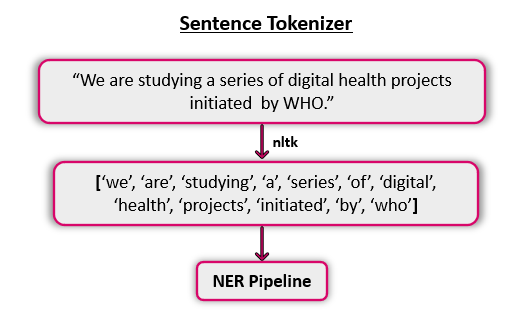}\\
\end{center}

\State A collection of lists, each containing a sentence of the document, is fed into the \hyperlink{fm2}{full} fine-tuned named entity recognition model, where predictions are made consecutively. Apart from the fact that the majority of NLP libraries expect a sentence-level input, this segmentation of the text into sentences rather than passing it as a single large string allows for a better contextual accuracy and facilitates error analysis as it is easier to pinpoint inaccurate predictions.

\vspace{1cm}

\subsubsection{NER: Filtering All Predictions}

\State This sub-chapter will provide a brief explanation of the screening that all entity predictions go through in order to obtain a final refined set of named entities that better portray a global health publication.

\State All predicted entities are stored in a dictionary that contains each unique  entity found in a publication as a key, ensuring a consistent set of references. The values associated with each unique key represent the number of occurrences said entity was found in the publication, acting as a total count. This data structure was chosen as the most versatile, allowing an efficient retrieval and manipulation of the data.

\State This selection of entity candidates was made with the simple verity in mind that the main named entities of a publication are those that occur the most through its text. Henceforth, the next step in the algorithm sorts the dictionaries by the highest count, to narrow down the number of named entities based on their occurrences in relation to the highest occurrence.

\begin{center}
    \includegraphics[scale = 0.6]{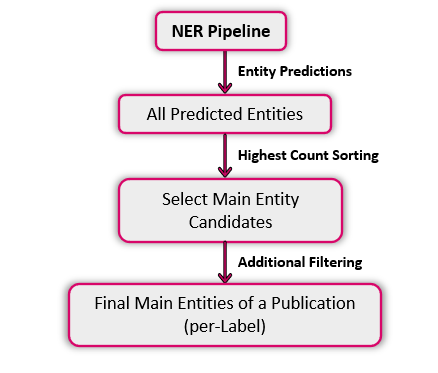}\\
\end{center}

\State With regard to each entity label, main entities are selected based on a threshold on their count in relation to the highest count, a threshold acting as a noise reduction mechanism. Knowing that entities that appear more frequently in a document will most likely be key organizations, countries, and persons that hold more significance in the text.\\
After benchmarking on a set of different documents, the main threshold chosen was $\frac{1}{3}$, meaning that succeeding entities with a count not less than $\frac{1}{3}$ of the highest count will be selected as main entity candidates. This approach aims to find a balance between simplicity and comprehensiveness, balancing the simplicity of choosing only the entity of the highest occurrence and the complexity of including every single rare correct predicted entity.

\begin{center}
    \includegraphics[scale = 0.65]{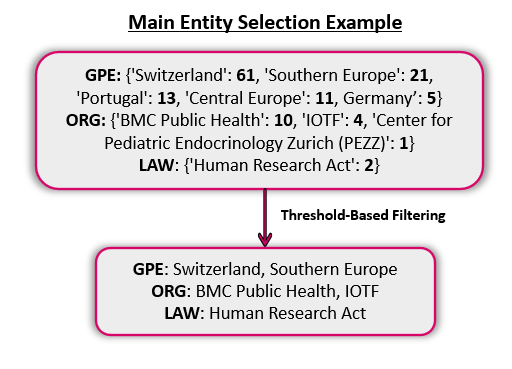}\\
\end{center}

\State The figure above displays a sample of the threshold-based filtering for the following publication: \textit{The increase in child obesity in Switzerland is mainly due to migration from Southern Europe – a cross-sectional study \footnote{\ \href{https://pubmed.ncbi.nlm.nih.gov/33514341/}{https://pubmed.ncbi.nlm.nih.gov/33514341/}}}, one of the publications that acted as a point of reference during benchmarking. Considering the content of the said publication, the entirety of extracted main entities represent central named entities to the content's narrative.

\State Once again, this strategy of extracting entities based on their frequency ensures a final entity collection that represents the publication's central themes. In addition to the interpretability, the computational efficiency of this approach needs to be mentioned as yet another benefit. Knowing that the extraction only leverages the distribution of entities rather than involving complex algorithms or extensive processing, computational efficiency is ensured.

\begin{center}
    \includegraphics[scale = 0.7]{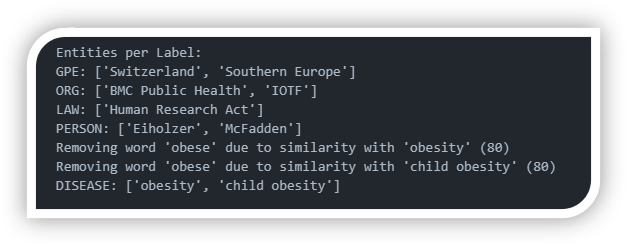}\\
    \textit{Example - Final Named Entities}
\end{center}

\State Apart from the threshold-based method, additional filtering methods that utilized the Levenshtein string similarity metric were applied, with the goal of providing a further refined final selection of extracted named entities. Specifically, methods that detect and eliminate similar words that hold the same meaning and a special method for disease entities that include sub-categories of a primary disease entity in the final list, irrespective of their frequency. (\textit{Child Obesity $\mapsto$ Obesity})

\newpage

\subsubsection{Medical Subject Headings (MeSH) Querying}
\State With all main named entities extracted, another method to link disease entities to their corresponding MeSH (Medical Subject Headings) tree was developed. MeSH is an extensive controlled vocabulary of medical information that provides a standardized terminology for medical terms such as our disease entities [\hyperlink{mesh}{50}]. Therefore, its hierarchical classification of diseases enriches the semantic context of found disease named entities within the realm of medical knowledge.

\begin{center}
    \includegraphics[scale = 0.6]{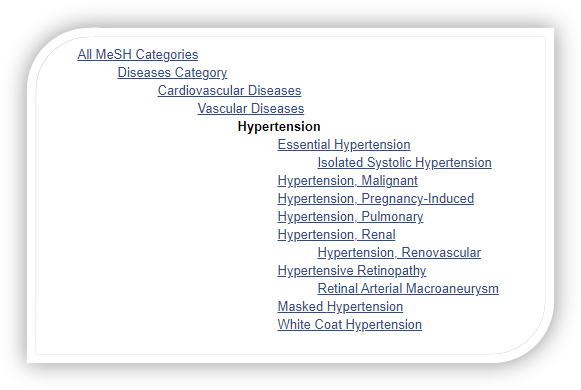}\\
    \textit{Corresponding MeSH Tree for 'Hypertension'}
\end{center}

\State Apart from enriched insights, including this classification into the final results is also beneficial in the realm of literature indexing. Knowing that the MeSH database is widely used in medical research, researchers can benefit from this feature when seeking publications in a knowledge base.

\vspace{0.5cm}

\State In terms of our development, the goal was to query and extract the tree displayed in the figure above for each found disease entity in order to provide additional information to potential users who are not familiar with certain medical terms. The \href{https://www.bionity.com/en/encyclopedia/Entrez.html}{\textcolor{unige}{\textit{\underline{Entrez}}}} Global Query Cross-Database Search System was utilized as a powerful engine that offers searching in the National Library of Medicine (NLM). Its incorporation into Python was achieved through \href{https://github.com/biopython/biopython}{\textcolor{unige}{\textit{\underline{Biopython}}}}, a library that offers freely available Python tools for computational molecular biology, including our live MeSH querying task.

\begin{center}
    \includegraphics[scale = 0.5]{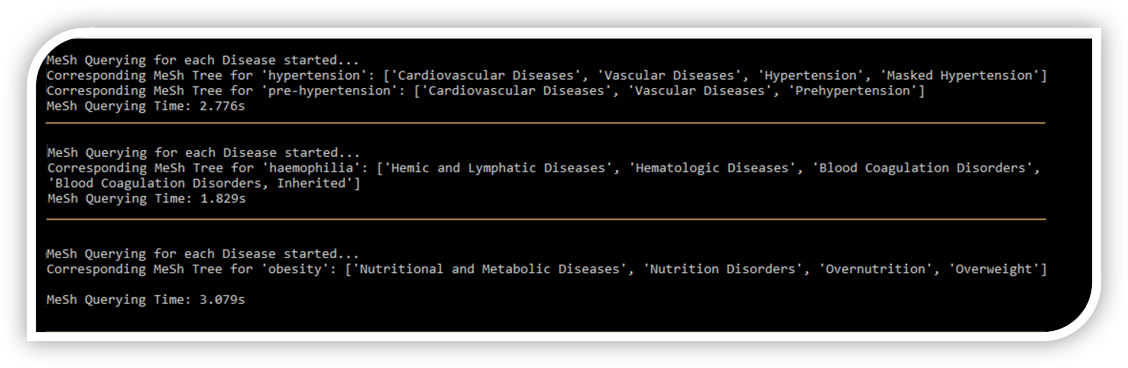}\\
    \textit{MeSH Querying - Example Results}
\end{center}

\State The sub-sample of the query results displayed above exhibits the ability of the library to extract the correct corresponding MeSH tree for each predicted disease entity in a publication. In our implementation, we exclusively extract branches within the hierarchical tree that are positioned in the 2 to 6 range. This ensures the obtention of contextual information related to the disease entity while avoiding the final branches of the tree that usually hold specialized variations of a disease, thus mitigating the risk of misclassification.

\State It should also be noted that the MeSH Querying algorithm is optional in relation to the entirety of the processing pipeline. Given the prompt execution time required for query completion, interested users have the freedom to employ it, without any requirement for its obligatory use.

\vspace{0.5cm}

\subsubsection{Final Benchmark}

\State After having established the full document processing pipeline, any global health publication can be passed in as a PDF input for the main named entities to be extracted. To ascertain the capacity of the developed pipeline, a benchmark was constituted. One that included a selection of publications from different health areas was created, in order to test the model's performance on an unrelated spectrum of global health literature and identify potential areas for refinements. \\
The performance of the model was evaluated in terms of ability to successfully process a publication and output a logical set of main named entities, the quality of extracted named entities with regard to their relatedness to the publication's central theme, and the total runtime.

\State At this point in time, the pipeline is run through the command line and a sample of the benchmarking results will be presented below via its visualization. Apart from a summarized view of the results displayed on each figure, additional remarks that elucidate the publication's context and the coherence of named entity extraction in relation to it will be shown.

\vspace{1cm}

\begin{center}
    \includegraphics[scale = 0.85]{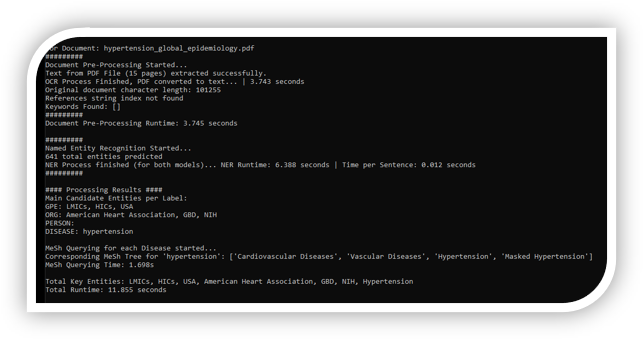}\\
    \textit{Remark:} \textit{The global epidemiology of hypertension \footnote{\ \href{https://pubmed.ncbi.nlm.nih.gov/32024986/}{https://pubmed.ncbi.nlm.nih.gov/32024986/}}}, an american publication that provides a statistical analysis on global hypertension in LMICs and HICs. Therefore the extractions of hypertension as a disease by itself, geopolitical named entities, and organizations linked to the study of global hyperentsion (e.g. GBD - Global Burden of Disease), is related to the publication. This is an example of an error-free robust named entity collection, yet a short one given the high count threshold.

    \includegraphics[scale = 0.85]{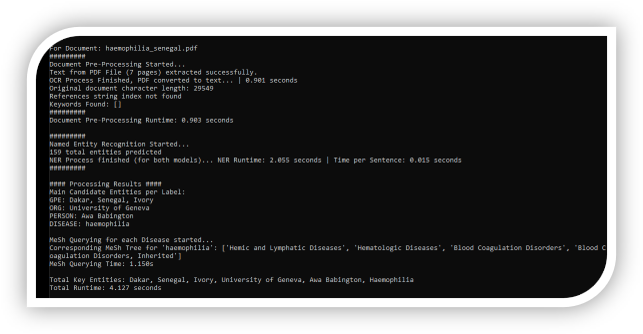}\\
    \textit{Remark:} \textit{Unravelling the knowledge, beliefs, behaviours and concerns of Persons with Haemophilia and their carriers in Senegal \footnote{\ \href{https://pubmed.ncbi.nlm.nih.gov/32666560/}{https://pubmed.ncbi.nlm.nih.gov/32666560/}}}, a qualitative study on haemophilia in Senegal done by researches at the University of Geneva. Apart from the entity results, the extracted MeSH sub-tree of haemophilia corresponds to the said disease and would attach further information to the publication in order to facilitate its indexing.\\[2cm]

    \includegraphics[scale = 0.8]{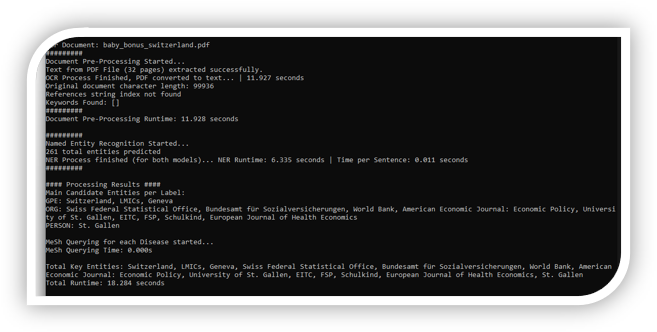}\\
    \textit{Remark:} \textit{Baby bonus in Switzerland: Effects on fertility, newborn health, and birth-scheduling \footnote{\ \href{https://pubmed.ncbi.nlm.nih.gov/34076325/}{https://pubmed.ncbi.nlm.nih.gov/34076325/}}}, a publication related more to health economics rather than a focus on a specific disease. This can be reflected on results too, given the numerous organizations. Apart from the mislabeling of a geolocation as a PERSON, the successful recognition of Bundesamt für Sozialversicherungen (Federal Statistical Office in German) should be noted. It is one entirely dependant on POS tags, knowing that the model is trained in English.\\[2cm] 

    \includegraphics[scale = 0.45]{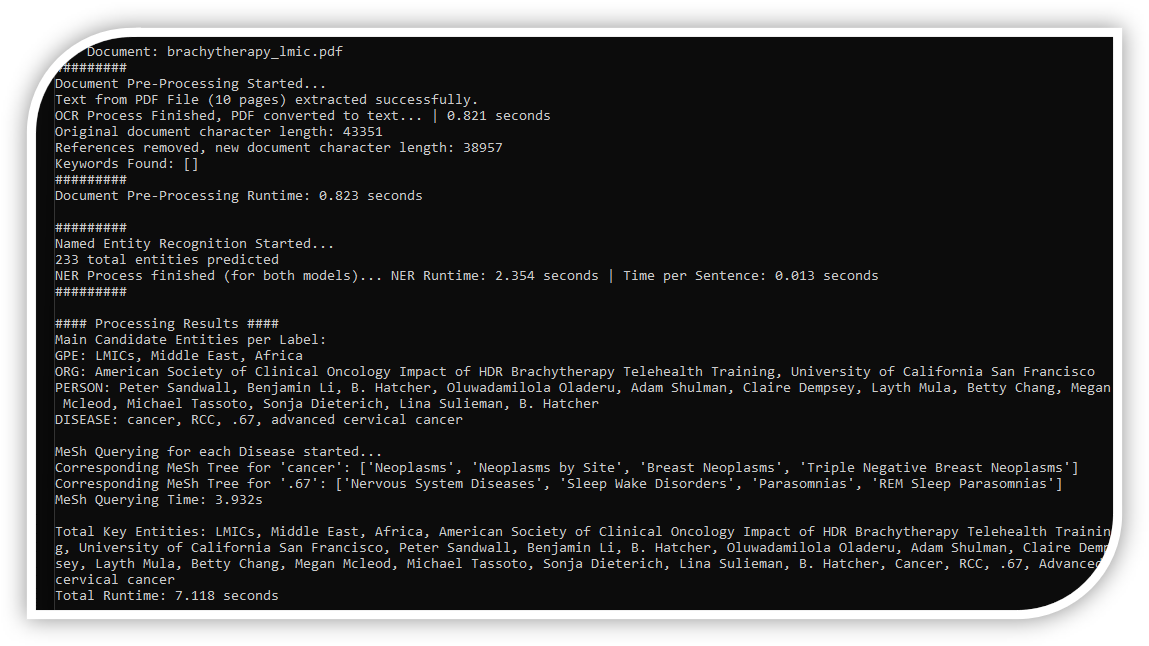}\\
    \textit{Remark:} \textit{Impact of High-Dose-Rate Brachytherapy Training via Telehealth in Low- and Middle-Income Countries \footnote{\ \href{https://www.ncbi.nlm.nih.gov/pmc/articles/PMC7713515/}{https://www.ncbi.nlm.nih.gov/pmc/articles/PMC7713515/}}}, a publication related to a telemedicine training course in brachyterapy in Middle East and Africa. The model's ability to extract numerous entities related to contributors in the publication should be noted in this case. Apart from that, the mislabeling of RCC (Rayos Contra Cancer, NGO) as a disease should be seen as a shortcoming of CNN models due to their lack of sentence context in comparison to transformer ones.
\end{center}

\vspace{1cm}

\State A pattern emerges from the results above, one that displays the model's ability to provide a final small robust selection of named entities that are related to each publication. Having said that, it would be difficult to define a discrete mathematical formula that determines the excellence of said selection, knowing the subjective nature of language and its context. Although if we would evaluate on a pragmatic criterion that considers as accurate a collection of error-free entities that contextually align with the publication's primary theme, our pipeline can be considered as \textbf{$\approx$ 91\%} accurate. \\
A lower threshold and a different set of hyperparameters provides a larger yet less robust final list of named entities. 

\State Another crucial factor that was considered before the development had started was the total runtime, where a reasonably fast final model was required. A 10-15 page publication requires an average of $\approx$ \textit{7s} to be fully processed on a local machine, a promising result considering that this is developed as potentially parallelizable pipeline able to process numerous documents at once. Having said that, let us take a closer look at how the different parts of the pipeline affect the total runtime distribution to provide a further analysis.

\vspace{0.5cm}

\begin{center}
    \includegraphics[scale = 0.6]{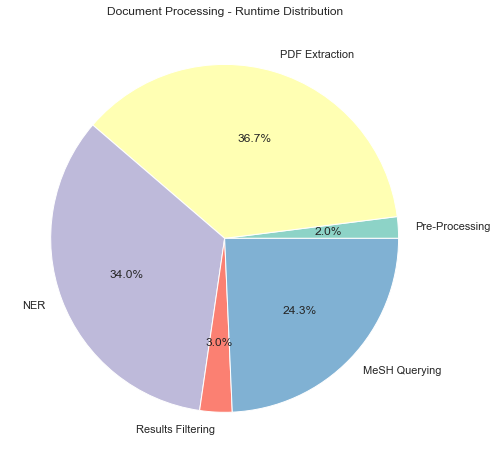}\\
    \textit{Document Processing Benchmark - Runtime Distribution}
\end{center}

\State One can notice that two external methods to the NER pipeline, PDF extraction to text and MeSH Querying, hold the majority of the total runtime. Notably, in addition to once again highlighting the speed of the NER model, this observation makes ground for further exploration on finding or developing new alternatives in text extraction and MeSH querying.

\newpage

\State One other potential evaluation method is the comparison of the final extracted main entities to the original keywords found in a publication, if there are any. This would directly compare our model to a handcrafted selection of keywords by the original authors.

\begin{center}
    \includegraphics[scale = 0.65]{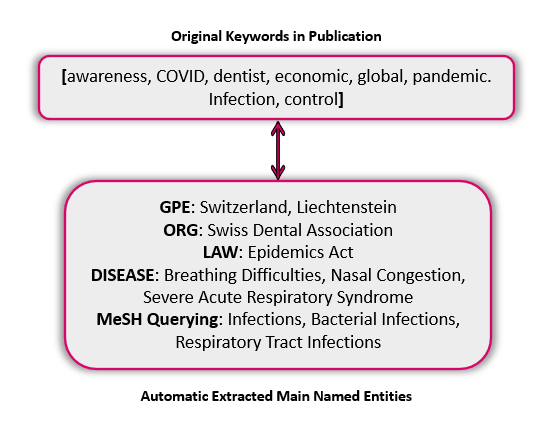}
\end{center}

\State The comparison example visualized above was generated from the results of the following publication: \textit{COVID-19 in Switzerland and Liechtenstein: A Cross-Sectional Survey among Dentists' Awareness, Protective Measures and Economic Effects \footnote{\ \href{https://pubmed.ncbi.nlm.nih.gov/33291659/}{https://pubmed.ncbi.nlm.nih.gov/33291659/}}}. It demonstrates the model's ability to provide named entities that function as more focused and contextual keywords, intricately connected to the publication's content, rather than its general topics. This granularity would provide a deeper document categorization that would facilitate indexing in a knowledge base.

\vspace{0.5cm}

\State Having said that, original keywords in a document are typically present to facilitate literature indexing and tend to align with broader general topics. Therefore, another NLP approach, distinct from NER yet of a lower significance to the project, was explored. An approach with the same objectives in mind that dove in the realm of automatic multi-label topic classification.

\newpage

\section{Multi-Label Topic Classification}
\State Topic classification is another natural language processing task that involves labelling a given textual content with the corresponding topic that it represents, essentially text categorization [\hyperlink{patternrec}{51}]. Text categorization stands as a macro-reading approach that grasps the overall theme of the content as opposed to named entity recognition that takes a micro-reading approach 
labelling each token in a sentence with its individual tag [\hyperlink{eisenstein_macro}{52}]. Therefore, it can be considered a more forgiving approach as it doesn't require the pinpoint accuracy of NER but rather an accurate theme that is related to the content.

\vspace{0.5cm}

\State The goal of this task is to assign a single or multiple categorical label to any given text input, a label based on the primary subject and theme of the input. Therefore, let us assume a given set $T$ of possible topics, and an input sequence $S$ represented as a sequence of words $S = \{w_1, w_2, w_3, ...., w_S\}$. The objective of the task to determine the most likely topic $t$ for a given sentence would be mathematically represented as: 

\begin{center}
    $Topic = argmax_{t \in T} P(t | S) = argmax_{t \in T} \frac{P(S | t) \cdot P(t)}{P(S)}$
\end{center}

\State The selected topic $t$ maximizes the conditional probability $P(t | S)$, consequently indicating that it is the most likely label given the sentence $S$. Usually and originally, this probability estimation leverages labelled sentence-topic data pairs, making this task another example of supervised learning.

\vspace{0.5cm}

\State Having said that, the majority of real world sentences and classification problems require each sample to be associated with multiple labels rather than a single one [\hyperlink{patternrec}{51}]. For example, the sentence $S = $ \textit{"Recent advancements in AI have revolutionized robotic surgery and paved the way for more precise procedures"} provides a more nuanced categorization if it is assigned to both labels $t_1$ = \textit{"Artificial Intelligence"} and $t_2$ = \textit{"Medical Surgery"}, rather than a singular one. \\

\State Therefore, given a set of possible labels $T$, another subset $T_S \ (T_S \subseteq T)$ that best categorizes the sentence $S$ must be determined.

\begin{center}
    $T_S = argmax_{T' \subseteq T} P(T' | S) = argmax_{T' \subseteq T} \frac{P(S | T') \cdot P(T')}{P(S)}$
\end{center}

\State Once again, the selection maximizes the conditional probability with relation to the sentence $S$, only now for a subset of possible topic labels $T'$ rather than a single label.

\State Apart from providing a more sophisticated and detailed categorization system, it is easily understood from both equations that multi-label topic classification provides for a more challenging task. It is a task that has to deal with trending challenges such as relationship between labels, presence of imbalanced labels, and high dimensionality of the output [\hyperlink{mll}{53}]. 

\State There exist a variety of neural and pre-neural approaches to this problem such as logistic regression, k-nearest neighbours, SVMs for multi-label classification, classifier chains, etc [\hyperlink{mll}{53}]. However, our limited amount of training data in addition to the requirement of obtaining a fast and accurate topic classification pipeline able to be merged with the NER one shifted the focus towards few-shot and zero shot classification methods. 

\subsection{Training a Classifier}

\State Knowing the nature of our task as a supervised learning one, we needed to provide pairs of topics and samples to any potential few-shot classification model, as zero shot models do not require any training. We opted out to the creation of a custom dataset that offered a set of labels $T$ and samples that are highly relatable to the nature of global health publications. This was a choice made to avoid high-dimensionality challenges that established state-of-the-art databases with thousands of labels such as EUR-Lex \footnote{\ \href{https://huggingface.co/datasets/eurlex}{HuggingFace - EUR-Lex Dataset}} provide.

\begin{center}
    \includegraphics[scale = 0.45]{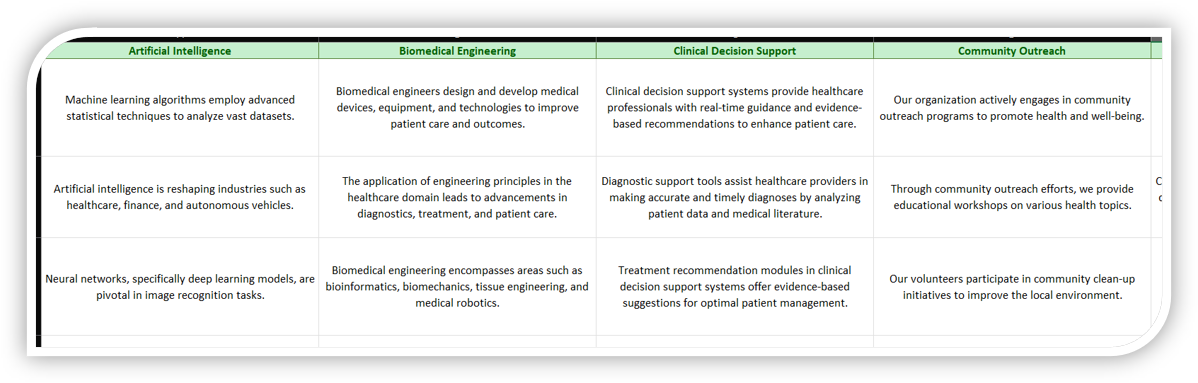}\\
    \textit{Labelled Data - Topic/Samples Examples}
\end{center}

\State The dataset was generated by including $50$ samples for each topic label, handcrafted samples that aimed to avoid repetition and ambiguity. Henceforth, this type of design was a choice made to exclusively evaluate the classification models on sentences that are related to our project and would normally be found in global digital health publications, rather than general ones.

\vspace{0.5cm}

\State Knowing that state-of-the-art transformer models are data-hungry to a degree that makes it very difficult or impossible to train one from scratch, few-shot learning models (FSL) stand as a great alternative that is generalizable to multi-label classification downstream tasks [\hyperlink{fsl}{54}]. It adheres to a lack of data through the design of models that require only handful of annotated examples in order to be fine-tuned to any set of topic labels, as long as it's annotated.

\vspace{0.5cm}

\State Our design of the labelled dataset is adapted to the prompt-learning method of FSL models, a method that provides examples in the forms of prompts to a model to guide its predictions during inference. In \textit{spaCy}, the library \textit{Classy Classification} \footnote{\ \href{https://spacy.io/universe/project/classyclassification}{https://spacy.io/universe/project/classyclassification}} was selected as one that offers this feature in \textit{spaCy} models. Specifically, the code snippet introduces a new \textit{text\_categorizer} pipe to our existing NER model, a pipe that utilizes a chosen pretrained sentence transformer FSL model. This design makes the entire pipeline able to classify topics too, alongside its main named entity recognition functionality, hence why it was chosen as the first approach.

\vspace{0.5cm}

\State Having selected the library, the optimal hyperparameters, and \textit{sentence-transformers/all-MiniLM-L6-v2} \footnote{\ \href{https://huggingface.co/sentence-transformers/all-MiniLM-L6-v2}{HuggingFace - sentence-transformers/all-MiniLM-L6-v2}} as our model, we arrived to the following results which will be visualized through a sub-sample:

\begin{center}
    \includegraphics[scale = 0.6]{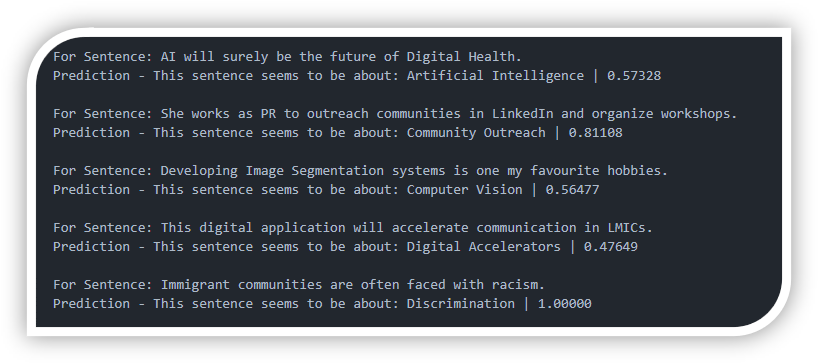}\\
    \textit{Few-Shot Classification - Examples of Results}
\end{center}

\State The results above evidenced the potential shortcomings of the few-shot classification model in the predictions of a test set designed for single label topic classification as an initial accuracy testing method. We arrived at the conclusion that the reliance solely on few-shot models falls short on making predictions withing the context of the requirement of this project and an alternative must be explored.

\State Even though the displayed sample of results showcases accurate predictions at first sight, it additionally reveals clear signs of overfitting in relation to the few-shot input. Specifically, overfitting is determined by the fact that the model's decisions are often driven by the limited word-by-word examples it was trained on. This leads to a substandard generalization to unseen data and can be observed through the imbalanced and overly confident determinations (e.g. a number of predictions with a probability of $1.0$, on account of correlating certain words in the sentence query and the train set). An outcome that prompts us to consider the presence of the "short-term memory loss" phenomena, where the model forgets contextual information learned at first training.

\State These observations might simply be a case of a substandard spaCy library (Classy Classification) in terms of providing an accurate robust text categorization pipeline based on few-shot models. Nevertheless, the increase in algorithmic complexity that multi-label classification introduces makes these observations even more concerning, amplifying the need to explore zero shot models as the other alternative for this task.

\subsection{Zero-Shot Classification}

\State Zero-shot classifiers stand as one of the most impressive evolutions in natural language processing. They are derived from unsupervised large language models (LLMs) that are trained on a huge amount of data that have accelerated the entire NLP research area through their great leap in performance for a variety of NLP tasks [\hyperlink{llms}{54}], including ours of multi-label classification. The ability of these models to be fine-tuned to any specific downstreamz task has made this pretrain-finetune paradigm the new approach to NLP tasks outside of research, replacing the more traditional knowledge based and fully supervised methods [\hyperlink{sidebyside}{55}]. In the context of our specific task, it means having a model able to make multi-label topic predictions on sentences without any prior training, a \textit{zero-shot} prediction.

\begin{center}
\begin{lstlisting}[language=Python, escapeinside={(*@}{@*)}]
from transformers import pipeline

### Load Chosen Zero Shot Model
### 0 for GPU Device, "pt" for PyTorch as the deep learning framework
zero_shot_classifier = (*@\textcolor{unige}{pipeline}@*)("zero-shot-classification", (*@\textcolor{unige}{model}@*) = "facebook/bart-large-mnli", (*@\textcolor{unige}{device}@*) = 0, (*@\textcolor{unige}{framework}@*) = "pt")

### Predict scores for a given list of sentences and required Topic Labels
all_scores = (*@\textcolor{unige}{zero\_shot\_classifier}@*)(
    (*@\textcolor{unige}{sequences}@*) = example_sentences,
    (*@\textcolor{unige}{candidate\_labels}@*) = all_labels,
    (*@\textcolor{unige}{multi\_label}@*) = True,
    src_lang="en",
)
\end{lstlisting}
\textit{Zero-Shot Classifiers in Python (HuggingFace)}\\
\end{center}

\State Therefore, a general purpose zero-shot model was seen as the most practical for our task, offering a unique solution to projects suffering from a scarcity of data. Apart from the scarcity challenge, the model's supposed ability to generalize over a variety of categories including topics that it has never seen before made it the natural choice for a domain that demands flexibility across a variety of different topics from multiple domains, such as global digital health.

\State In development, the Facebook trained BART model on the MultiNLI (MNLI) dataset was loaded in Python through HuggingFace. It excels as a zero-shot sequence classifier based on the NLI-based prediction method, proposed by the work of \textit{Yin et al} [\hyperlink{yiniyin}{57}].

\subsection{Evaluation}

\State A more formal test set for both the few-shot and zero-shot classifiers was designed, one that included 1000 handcrafted sentences. For each of one of the 50 topic labels, 20 sentences tagged with the corresponding true label were generated, in order to compute accuracy in a single-label classification scenario. The sentences were crafted with one main central topic in mind that they would have attached as a true label while retaining the potential for broader interpretations and further labelling with other topics.

\State Taking the following test sentence as an example: \textit{"Through the generous funding of the European Union's Horizon 2020 program under grant agreement No. 123456, our study investigates the role of artificial intelligence in personalized medicine."}, a sentence tagged with \textit{Publicly Funded Science} as its main topic label to evaluate on a single label scenario. Nevertheless, it stands as a sentence that can be tagged with the label 'Artificial Intelligence' and 'Digital Health' in a multi label scenario too. 

\State Apart from that, 10\% of the dataset was intentionally dedicated to adversary sentences meant to trick the model during classification. These are sentences that seem to align with a certain topic at first sight however they harbor altered words intended for misclassification, posing a deeper semantic challenge to the models.

\begin{center}
    \begin{tabular}{|m{4.5cm}|c|c|c|} 
        \hline
        Model & Single-Label Accuracy & Multi-Label Accuracy \\\hline
        Few-Shot Classifier (\textit{all-MiniLM-L6-v2})  & 59\% & 32\% \\\hline
        Zero-Shot Classifier (\textit{facebook/bart-large-mnli}) & 95.2\% & 88\% \\\hline
    \end{tabular}\\[0.4cm]
\end{center}

\State Even though the few-shot classifier manages to correctly label a simple majority of the sentences, these results prove once again the superiority of LLMs as generalizable zero-shot classifiers in the realm of topic classification too.

\State \textit{Note: } As for the multi label scenario, a certain degree of manual assessment becomes necessary given the subjective nature of contextual importance of labels. For example, given a sentence that mentions both "\textit{policy reform}" and "\textit{climate change}" with the the automated algorithm having them both tagged as true and a human annotator that thinks otherwise.

\begin{center}
    \includegraphics[scale = 0.55]{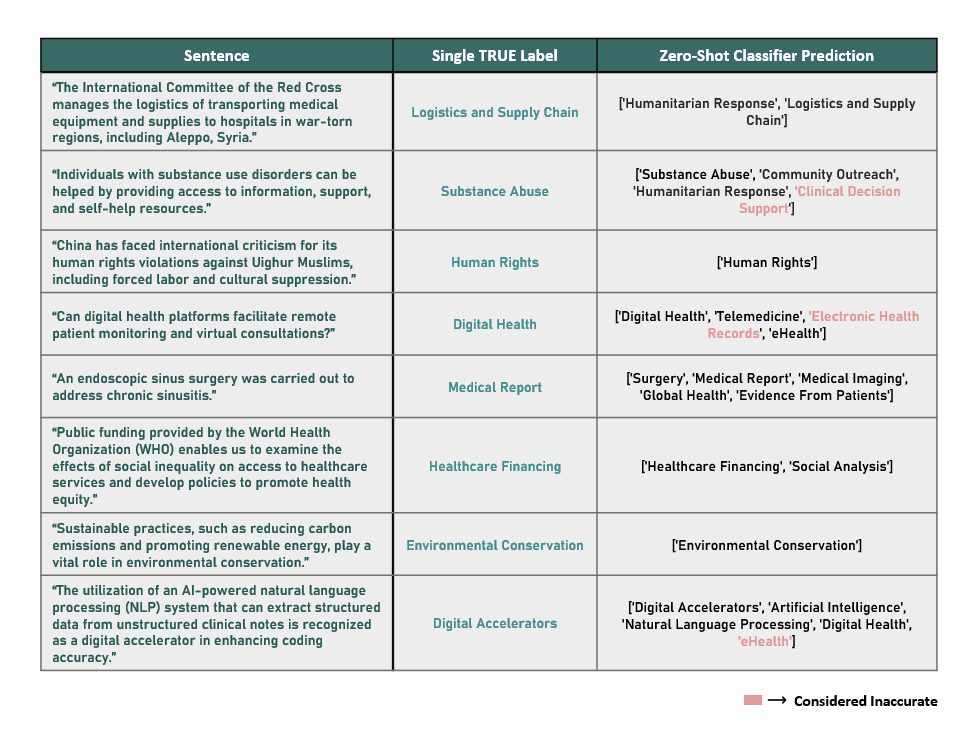}\\
    \textit{Zero-Shot Topic Classification - Sample of Results}
\end{center}

\State We could take a closer look at the results of the benchmark from the zero-shot classifier's point of view knowing that it was chosen as the favourite. Despite a few examples of incorrect predictions (again, \textit{subjectively thinking}), the performance in multi-labelling a sentence is clear. The selection threshold was a probability of $p \geq 0.7$ per topic, therefore these were predictions made with a high level of confidence by the model which exhibits a great contextual understanding from the chosen BERT classifier. \\
Furthermore, I intentionally selected for display above a sample of sentences that represent different domains ranging from human rights, to economy and health. A correct labelling of this selection stands as a sign of the generalization ability of the model no matter the origin of the topic label.

\State The zero-shot classifier was adapted to the document processing \hyperlink{dpp}{pipeline} of the named entity recognition model, in order to be tested with tagging a full publication instead of singular sentences. Frequency-based filtering of topic predictions was made, the same approach as in the case of NER.

\begin{center}
    \includegraphics[scale = 0.7]{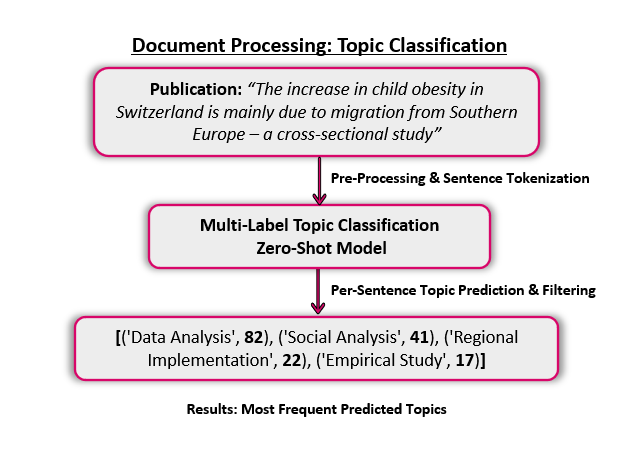}\\
    \textit{Topic Classification on a Full Document Level}
\end{center}

\State The final selected topics (most frequent ones) are compatible with the central theme of the publication, keeping in mind that the predictions are made from a pool of only 50 labels. This is a performance that stays true for most of the publications, showing the model's capability to automatically tag any given publication. 

\State \textbf{\textit{However}}, while the NER model requires 2 seconds to process the whole document, the selected zero-shot classifier requires the same 2 seconds to process a single sentence, for a total of 400s second for the full publication using a Nvidia GTX 1650 TI GPU. Therefore, despite the impressing accuracy, zero-shot classifiers come with a computational complexity that is not practical for every application. Such is the case of ours, where the combination of performance and speed were requirements knowing the nature of global health implementations.

\newpage

\section{Discussion \& Conclusion}
\State Before jumping to the conclusion, I would like to briefly discuss each of the three making parts of this thesis, separately. These should be seen as reflections in the terms of their effect on the main problem statement, research question, and industry need behind this research journey of the past year. Apart from that, each corresponding natural language processing topic should not be forgotten to be mentioned if it contributed to filling existing gaps in the general research in addition to meeting my initial expectations.

\vspace{0.5cm}

\begin{center}
    \underline{\textit{word2vec}}
\end{center}

\State Initially chosen as a word embedding method that can discover intricate similarities between words based on their vector representation in a computationally efficient manner, word2vec still stands as a fast way to embed words. For the majority of tags, the real semantic proximity between words directly corresponded to the cosine similarity between vectors in the mathematical space.

\State Comparing the same word2vec model against itself in different scenarios of being trained with a larger corpora each time gave us insights in the capturing of localized nuances. An insufficient corpora results in an insufficient model while a large corpora with a vocabulary size in the billions doesn't capture localized contexts and loses the main point of automatic tag discovery. If you want to discover underlying contextual insights of a certain domain in a certain region, train the word2vec model with publications of that domain and region exclusively.

\State Reflecting on the initial choice of using word2vec as a word embedding alternative, a further work would have included ELMo as another more novel word embedding method that potentially adresses the limitations of word2vec. Alternatively, considering the landscape of NLP research community and the development of contextual models such as BERT, traditional methods such as word2vec might stand as a relic of the past [\hyperlink{relic}{58}]. 

\newpage

\begin{center}
    \underline{\textit{Named Entity Recognition}}
\end{center}

\State As the main NLP study area of this project, named entity recognition (NER) proved itself as a fundamental task of entity information extraction in different domains and natures of text. My decision to choose NER for this project and for my further research opportunities is based on its remarkable customizability and adaptability to tasks. 

\State Adaptability to any task and domain that involves entities can be observed in the ability to train any NER model to recognize in language any type of given entity label as long as it has the required labelled data to train for the recognition. This was perceptible in our task from the variety of entity labels ranging from locations, geopolitical entities, global and national laws, to medical diseases. 

\State Frameworks such as spaCy support this venue of flexibility through their model's possibility to be trained from scratch or fine-tuned on domain-specific labels, achieving adequate entity extraction results in the process. Fine-tuning a pretrained off-the-shelf model to recognize domain-specific entities should be the norm of utilization of named entity recognition pipelines, as it is for the majority of other NLP downstream tasks.

\State Despite the explosion in popularity of transformer-based and large language models, the industry need speed requirement of this project showcased the practicality of another type of network, convolutional neural networks (CNNs). The rapid processing of documents without compromising accuracy aligned with the practical nature of the task where timely information retrieval was required, as imagined and would be required in a remote regional health implementation.

\State Apart from the performance and customizability of NER, some concerns and potential further works were raised during research. Despite the model's flexibility on entity labels, the lack of public labeled data remains a major difficulty. While there exist datasets for the main most popular entity types (organizations, locations, biomedical entities), other specific entity types (e.g. Global Health Laws) suffer from a lack of labelled data. Additionally, the document processing pipeline suffered from PDF text extractors in both information loss and total runtime given that incorrect text extractions result in a data loss which confuses any type of model, including the NER one. Therefore, increasing the accuracy text extraction methods should be seen as another potential future research area.

\begin{center}
    \underline{\textit{Topic Classification}}
\end{center}

\State In any case where we need to label a certain piece of information content for retrieval purposes, automatic topic classification should be explored as a task. An accurate classification that is able to categorize content into predefined topics facilitates information search, enhances knowledge, and provides a better user experience. Knowing that content (especially textual) has contextual nuances and cannot always be labelled with only a single topic, multi-label classification should be a prerequisite.

\State Despite the numerous traditional statistical methods and models to approach a classification task, the more modern performant large language models are recommended to be chosen in cases where the task isn't the focal point of the research such as this one. The comparison of two modern NLP evolutions such as the few-shot and general zero-shot classifiers proved the superiority of the latter in single and multi-label classification tasks.

\State The only risen concern behind the utilization of resource intensive models such as LLMs is related to their computational and time complexity, substandard in comparison to the NER model. Consequently, further work to tackle this problem should be explored in the form of more lightweight architectures or other model compression techniques. 

\newpage

\begin{center}
    \underline{\textit{Conclusion}}
\end{center}

\State Through this natural language processing's odyssey, we have unveiled the importance and adaptability of NLP tasks to any problem related to language and text, be it for research or industrial applications. What started as a task to annotate global digital health publications for a knowledge base, turned into a theoretical research of three pivotal NLP areas.

\State Word embeddings such as word2vec remain a prominent way to grasp how text is transformed into a mathematical representations, but recent developments based on contextual embeddings and transformers might have left them in the past. \\
Named Entity Recognition (NER) stands as one of the most important and customizable task of NLP, allowing the recognition of any thought entity type as long as it has been trained with the proper labelled data to do so. Convolutional Neural Networks (CNNs) still stand as a balance between speed and accuracy, achieving named entity extraction of entire research publications in under 2s. \\
Topic Classification should be part of any system that contains information retrieval of textual content, where multi-label classification provides a more efficient and contextually complete information. Modern large language models adaptable to the task with minimal (few-shot) or no (zero-shot) fine-tuning should be prioritized because of their accuracy, keeping in mind the computational complexity behind them.

\State Further work and research should be put towards recent developments in NLP such as generative models, zero-shot classifiers, and multi-modal language models. In a rapidly advancing field such as the one of AI, these might be the models of the future and turn obsolete anything released prior to them.


\newpage

\begin{thebibliography}{9}

\hypertarget{eisenstein_intro}{\bibitem{eisenstein_intro}}
\href{https://cseweb.ucsd.edu/~nnakashole/teaching/eisenstein-nov18.pdf}{Eisenstein J., (2018), Natural Language Processing, Chapter 1.1: Natural language processing and its neighbors}

\hypertarget{nlp_applications}{\bibitem{nlp_applications}}
\href{https://www.mecs-press.org/ijitcs/ijitcs-v7-n8/IJITCS-V7-N8-7.pdf}{Iroju G. O., Olaleke O. J., (2015), Natural Language Processing, A Systematic Review of Natural Language
Processing in Healthcare}

\hypertarget{liberate}{\bibitem{liberate}}
\href{https://www.researchgate.net/publication/360937780_Diseases_Classification_using_advance_Machine_learning_Techniques}{Payal M. et al, (2022), Diseases Classification using advance Machine learning Techniques}

\hypertarget{word_embeddings}{\bibitem{word_embeddings}}
\href{http://josecamachocollados.com/book_embNLP_draft.pdf}{Camacho-Collados J., Pilehvar T. M., (2020), Embeddings in Natural Language Processing: Theory and Advances in Vector Representations of Meaning, Chapter 3: Word Embeddings}

\hypertarget{pred_embeddings}{\bibitem{pred_embeddings}}
\href{http://josecamachocollados.com/book_embNLP_draft.pdf}{Camacho-Collados J., Pilehvar T. M., (2020), Embeddings in Natural Language Processing: Theory and Advances in Vector Representations of Meaning, Chapter 3.2: Predictive Models}

\hypertarget{dae_embeddings}{\bibitem{dae_embeddings}}
\href{https://cseweb.ucsd.edu/~nnakashole/teaching/eisenstein-nov18.pdf}{Eisenstein J., (2018), Natural Language Processing, Chapter 14.5: Neural word embeddings}

\hypertarget{word2vec}{\bibitem{word2vec}}
\href{https://arxiv.org/abs/1301.3781}{Mikolov T. et al, (2013), Efficient Estimation of Word Representations in Vector Space}

\hypertarget{word2vec_jura}{\bibitem{word2vec_jura}}
\href{https://web.stanford.edu/~jurafsky/slp3/}{Jurafsky D., Martin H. J., (2021), Speech and Language Processing, Chapter 6.8: Word2vec}

\hypertarget{neo4j_imp}{\bibitem{neo4j_imp}}
\href{https://www.sciencedirect.com/science/article/pii/S0306457322002667}{Aldwairi M. et al, (2023), Graph-based data management system for efficient information storage, retrieval and processing}

\hypertarget{who_dhis}{\bibitem{who_dhis}}
\href{https://www.who.int/publications/i/item/WHO-RHR-18.06}{World Health Organization, (2018), Classification of digital health interventions v1.0}

\hypertarget{classify}{\bibitem{classify}}
\href{https://www.who.int/publications/i/item/WHO-RHR-18.06}{Bailey D. K.,  (1994), Typologies and Taxonomies: An Introduction to Classification Techniques (Quantitative Applications in the Social Sciences)}

\hypertarget{represent}{\bibitem{represent}}
\href{https://www.taylorfrancis.com/chapters/edit/10.4324/9780203856949-6/building-spoken-corpus-adolphs-svenja-knight-dawn}{Svenja A., (2010), Building a spoken corpus - What are the basics?}

\hypertarget{neppy}{\bibitem{neppy}}
\href{https://pubmed.ncbi.nlm.nih.gov/35280611/}{Gupta P. P. et al, (2021), Study of the impact of a telemedicine service in improving pre-hospital care and referrals to a tertiary care university hospital in Nepal}

\hypertarget{booky}{\bibitem{booky}}
\href{https://pubmed.ncbi.nlm.nih.gov/35280611/}{Vermund H. S. et al, (2021), Sexually Transmitted Infections: Adopting a Sexual Health Paradigm}

\hypertarget{wow}{\bibitem{wow}}
\href{https://pubmed.ncbi.nlm.nih.gov/35280611/}{Zhang Y. et al, (2019), BioWordVec, improving biomedical word embeddings with subword information and MeSH}

\hypertarget{skippy}{\bibitem{skippy}}
\href{https://pubmed.ncbi.nlm.nih.gov/35280611/}{Krishna P. P., Sharada A., (2019), Word Embeddings - Skip Gram Model}

\hypertarget{norm_me_sire}{\bibitem{norm_me_sire}}
\href{https://web.stanford.edu/~jurafsky/slp3/}{Jurafsky D., Martin H. J., (2021), Speech and Language Processing, Chapter 2.4: Text Normalization}

\hypertarget{mikolov2}{\bibitem{mikolov2}}
\href{https://arxiv.org/abs/1301.3781}{Mikolov T. et al, (2013), Distributed Representations of Words and Phrases and their Compositionality}

\hypertarget{npmi}{\bibitem{npmi}}
\href{https://arxiv.org/abs/1301.3781}{Bouma G., (2009), Normalized (Pointwise) Mutual Information
in Collocation Extraction}

\hypertarget{hyperpamparam}{\bibitem{hyperpamparam}}
\href{https://arxiv.org/abs/1301.3781}{Rong X., (2016), word2vec Parameter Learning Explained}

\hypertarget{melvins}{\bibitem{melvins}}
\href{https://arxiv.org/pdf/1402.3722.pdf}{Goldberg Y., Levy O., (2014), word2vec Explained: deriving Mikolov et al.'s negative-sampling word-embedding method}

\hypertarget{int_ext}{\bibitem{int_ext}}
\href{https://www.cambridge.org/core/journals/apsipa-transactions-on-signal-and-information-processing/article/evaluating-word-embedding-models-methods-and-experimental-results/EDF43F837150B94E71DBB36B28B85E79}{Wang B. et al, (2019), Evaluating word embedding models: methods and experimental results, Cambridge University Press}

\hypertarget{int_ext_due}{\bibitem{int_ext_due}}
\href{https://aclanthology.org/D15-1036.pdf}{Schnabel T., Labutov I., Joachims T. M. D., 2015, Evaluation methods for unsupervised word embeddings}

\hypertarget{jurafsky_perp}{\bibitem{jurafsky_perp}}
\href{https://web.stanford.edu/~jurafsky/slp3/}{Jurafsky D., Martin H. J., (2021), Speech and Language Processing, Chapter 3.2.1: Perplexity}

\hypertarget{einstein_perp}{\bibitem{einstein_perp}}
\href{https://cseweb.ucsd.edu/~nnakashole/teaching/eisenstein-nov18.pdf}{Eisenstein J., (2018), Natural Language Processing, Chapter 6.4.2: Perplexity}

\hypertarget{viz_problem}{\bibitem{viz_problem}}
\href{https://ieeexplore.ieee.org/document/8019864}{Liu Sh. et al, (2018), Visual Exploration of Semantic Relationships in Neural Word Embeddings}

\hypertarget{clustering}{\bibitem{clustering}}
\href{https://dl.acm.org/doi/pdf/10.1145/1014052.1014118}{Dhillon S. I., Guan Y., Kulis B., (2004), Kernel k-means, Spectral Clustering and Normalized Cuts}

\hypertarget{viz_me}{\bibitem{viz_me}}
\href{https://aclanthology.org/N16-1082.pdf}{Jurafsky D. et al, (2016), Visualizing and Understanding Neural Models in NLP}

\hypertarget{eisenstein_ner}{\bibitem{eisenstein_ner}}
\href{https://cseweb.ucsd.edu/~nnakashole/teaching/eisenstein-nov18.pdf}{Eisenstein J., (2018), Natural Language Processing, Chapter 8.3: Named Entity Recognition}

\hypertarget{artcapja}{\bibitem{artcapja}}
\href{https://aclanthology.org/W03-0419.pdf}{Tjong Kim Sang E., De Meulder F., (2003), Introduction to the CoNLL-2003 Shared Task: Language-Independent Named Entity Recognition}

\hypertarget{lmicsL}{\bibitem{lmicsL}}
\href{https://globalizationandhealth.biomedcentral.com/articles/10.1186/s12992-018-0424-z}{Labrique B. A. et al, (2018), Best practices in scaling digital health in low and middle income countries}

\hypertarget{glove}{\bibitem{glove}}
\href{https://aclanthology.org/D14-1162/}{Pennington J., Socher R., Manning D. Ch., (2014), GloVe: Global Vectors for Word Representation}

\hypertarget{ontomom}{\bibitem{ontomom}}
\href{https://catalog.ldc.upenn.edu/LDC2013T19}{Weischedel R. et al, (2013), OntoNotes Release 5.0, Philadelphia: Linguistic Data Consortium}

\hypertarget{wordie}{\bibitem{wordie}}
\href{https://wordnet.princeton.edu/}{Princeton University, (2010), Wordnet 3.0}

\hypertarget{berti}{\bibitem{berti}}
\href{https://arxiv.org/abs/1810.04805}{Devlin J. et al, (2019), BERT: Pre-training of Deep Bidirectional Transformers for Language Understanding}

\hypertarget{cnn1}{\bibitem{cnn1}}
\href{https://arxiv.org/abs/1408.5882}{Kim Yoon, (2014), Convolutional Neural Networks for Sentence Classification}

\hypertarget{spacy_book}{\bibitem{spacy_book}}
\href{https://books.google.ch/books?id=w_ZqywEACAAJ&printsec=copyright&redir_esc=y#v=onepage&q&f=false}{Vasiliev Yuli, (2020), Natural Language Processing with Python and spaCy: A Practical Introduction}

\hypertarget{attention}{\bibitem{attention}}
\href{https://arxiv.org/abs/1706.03762}{Vaswani A. et al, (2017), Attention Is All You Need}

\hypertarget{roberti}{\bibitem{roberti}}
\href{https://arxiv.org/abs/1907.11692}{Liu Y. et al, (2019), RoBERTa: A Robustly Optimized BERT Pretraining Approach}

\hypertarget{dependent_book}{\bibitem{dependent_book}}
\href{https://compass.onlinelibrary.wiley.com/doi/epdf/10.1111/j.1749-818X.2010.00187.x}{Nivre Joakim, (2010), Language and Linguistics Compass, Volume 4, Issue 3}

\hypertarget{nested_ent}{\bibitem{nested_ent}}
\href{https://compass.onlinelibrary.wiley.com/doi/epdf/10.1111/j.1749-818X.2010.00187.x}{Yu J., Bohnet B., Poesio M., (2020), Named Entity Recognition as Dependency Parsing}

\hypertarget{transition_parser}{\bibitem{transition_parser}}
\href{https://compass.onlinelibrary.wiley.com/doi/epdf/10.1111/j.1749-818X.2010.00187.x}{Honnibal M., Johnson M., (2015), An Improved Non-monotonic Transition System for Dependency Parsing}

\hypertarget{scores}{\bibitem{scores}}
\href{https://cseweb.ucsd.edu/~nnakashole/teaching/eisenstein-nov18.pdf}{Eisenstein J., (2018), Natural Language Processing, Chapter 4.4.1: Precision, Recall, and F-Measure}

\hypertarget{finetuning}{\bibitem{finetuning}}
\href{https://web.stanford.edu/~jurafsky/slp3/}{Jurafsky D., Martin H. J., (2021), Speech and Language Processing, Chapter 11.3: Transfer Learning through Fine-Tuning}

\hypertarget{annot}{\bibitem{annot}}
\href{https://web.stanford.edu/~jurafsky/slp3/}{Halevy A., Norvig P., Pereira F., (2009), The Unreasonable Effectiveness of Data}

\hypertarget{critme}{\bibitem{critme}}
\href{https://cseweb.ucsd.edu/~nnakashole/teaching/eisenstein-nov18.pdf}{Eisenstein J., (2018), Natural Language Processing, Chapter 4.5: Building Datasets}

\hypertarget{adam}{\bibitem{adam}}
\href{https://arxiv.org/abs/1412.6980}{Kingma P. D., Ba L. J., (2015), Adam: A method for Stochastic Optimization}

\hypertarget{spacy_book}{\bibitem{spacy_book}}
\href{https://books.google.ch/books?id=w_ZqywEACAAJ&printsec=copyright&redir_esc=y#v=onepage&q&f=false}{Vasiliev Yuli, (2020), Natural Language Processing with Python and spaCy: A Practical Introduction, Chapter 10: Training Models}

\hypertarget{ncbi}{\bibitem{ncbi}}
\href{https://www.researchgate.net/publication/259606335_NCBI_Disease_Corpus_A_Resource_for_Disease_Name_Recognition_and_Concept_Normalization}{Dogan Islamaj Rezarta et al., (2014), NCBI Disease Corpus: A Resource for Disease Name Recognition and Concept Normalization}

\hypertarget{mesh}{\bibitem{mesh}}
\href{https://www.nlm.nih.gov/hmd/collections/digital/MeSH/mesh.html}{National Library of Medicine, (1960), Medical Subject Headings: Main headings, Subheadings, and Cross references used in the Index Medicus and the National Library of Medicine Catalog}

\hypertarget{patternrec}{\bibitem{patternrec}}
\href{https://www.academia.edu/12473445/Learning_multi_label_scene_classification}{Brown M. C. et al, (2004), Learning multi-label scene classification, Pattern Recognition}

\hypertarget{eisenstein_macro}{\bibitem{eisenstein_macro}}
\href{https://cseweb.ucsd.edu/~nnakashole/teaching/eisenstein-nov18.pdf}{Eisenstein J., (2018), Natural Language Processing, Chapter 17: Information Extraction}

\hypertarget{mll}{\bibitem{mll}}
\href{https://www.researchgate.net/publication/267154292_Multilabel_Learning_A_Review_of_the_State_of_The_Art_and_Ongoing_Research#fullTextFileContent}{Gibaja E., Ventura S., (2014), Multilabel Learning: A Review of the State of The Art and Ongoing Research}

\hypertarget{fsl}{\bibitem{fsl}}
\href{https://arxiv.org/pdf/2302.10447.pdf}{Liao W. et al, (2021), Mask-guided BERT for Few Shot Text
Classification}

\hypertarget{llms}{\bibitem{llms}}
\href{https://arxiv.org/pdf/2101.02661.pdf}{Gera A. et al, (2022), Zero-Shot Text Classification with Self-Training}

\hypertarget{sidebyside}{\bibitem{sidebyside}}
\href{https://arxiv.org/pdf/2210.17541.pdf}{Sainz O., Rigau G., (2021), Ask2Transformers: Zero-Shot Domain labelling with Pre-trained Language Models}

\hypertarget{yiniyin}{\bibitem{yiniyin}}
\href{https://arxiv.org/pdf/1909.00161.pdf}{Yin W. et al, (2019), Benchmarking Zero-shot Text Classification: Datasets, Evaluation and Entailment Approach}

\hypertarget{relic}{\bibitem{relic}}
\href{https://arxiv.org/pdf/2110.01804.pdf}{Sezerer E., Tekir S., (2020), A Survey On Neural Word Embeddings}




\end{thebibliography}
\end{document}